\documentclass[10pt,twocolumn,letterpaper]{article}
\usepackage[pagenumbers]{cvpr}
\usepackage{mathtools}
\usepackage{amsthm}
\usepackage{tikz}
\usepackage{pgfplots}
\pgfplotsset{compat=1.18}
\usetikzlibrary{arrows.meta,positioning,fit,calc}
\usepackage{algorithm}
\usepackage{algpseudocode}
\usepackage{microtype}
\usepackage{placeins}
\usepackage{needspace}
\usepackage{colortbl}
\definecolor{oursrow}{HTML}{F9E9EE}
\newcommand{\method}{ConsensusAlign}
\newcommand{\R}{\mathbb{R}}
\newcommand{\V}{\mathcal{V}}
\newcommand{\Lset}{\mathcal{L}}
\newcommand{\E}{\mathcal{E}}
\newcommand{\polar}{\operatorname{polar}}

\DeclareMathOperator*{\argmax}{arg\,max}

\DeclareMathOperator{\Hung}{Hung}
\newtheorem{proposition}{Proposition}
\definecolor{visionblue}{HTML}{0072B2}
\definecolor{textorange}{HTML}{D55E00}
\definecolor{graphgreen}{HTML}{009E73}
\tikzset{enc/.style={draw,rounded corners=2pt,minimum height=6mm,align=center,font=\fontsize{8}{9}\selectfont},flow/.style={-{Latex[length=1.6mm]},semithick}}

\newcommand{\CAReviewWithinPairAgreementAUROC}{0.9444}
\newcommand{\CAReviewWithinPairObjectiveAUROC}{1.0000}

\newcommand{\CAPooledMean}{0.0099}

\newcommand{\CAPooledAVG}{19.6}

\newcommand{\CAPooledIToT}{22.0}

\newcommand{\CACurrentWP}{0.1457}

\newcommand{\CADownSuccessCells}{21}

\newcommand{\CADownPipelineMinutes}{36.0}
\newcommand{\CADownPairwiseMinutes}{35.9}
\newcommand{\CADownSuccessWPAVG}{15.4}
\newcommand{\CADownSuccessOursAVG}{21.4}
\newcommand{\CADownSuccessWPIT}{12.2}
\newcommand{\CADownSuccessOursIT}{26.3}

\newcommand{\CAAttachSecondsLow}{0.2}
\newcommand{\CAAttachSecondsHigh}{1.0}

\newcommand{\CAPipeDistributeSeconds}{2.5}

\newcommand{\CADiagMaps}{161}
\newcommand{\CADiagFailures}{124}
\newcommand{\CADiagAUROCAgree}{0.974}
\newcommand{\CADiagAUROCObj}{0.873}

\newcommand{\CAROCOFOSOurs}{0.51}
\newcommand{\CAROCOAgreeOurs}{0.0004}

\newcommand{\CAROCOFOSPair}{0.47}

\newcommand{\CAROCOFOSSup}{0.15}

\newcommand{\CAROCOPairsPerSeed}{18}
\newcommand{\CAROCOSeeds}{3}

\newcommand{\CAFullWeighted}{0.0126}

\newcommand{\CAFullUnweightedFails}{4}

\newcommand{\CAVLMCollAVG}{24.9}

\newcommand{\CAVLMCollIT}{27.7}
\newcommand{\CAVLMCollTI}{23.9}
\newcommand{\CAVLMTTCAVG}{21.9}

\newcommand{\CAVLMTTCIT}{15.0}
\newcommand{\CAVLMTTCTI}{30.8}

\newcommand{\CACeilGridMean}{0.0036}

\newcommand{\CACeilAVG}{24.7}

\newcommand{\CACeilIToT}{36.0}
\newcommand{\CACeilTToI}{29.2}

\newcommand{\CAMechCKAPooled}{0.73}

\newcommand{\CAMechCKANativeMax}{0.68}
\newcommand{\CAMechCKANativeMin}{0.48}

\newcommand{\CAMechFailN}{48}
\newcommand{\CAMechFailHigher}{44}

\newcommand{\CAMechChanceN}{17}
\newcommand{\CAMechChanceHigher}{2}

\newcommand{\CAMechCells}{103}

\newcommand{\CAHPSotalignCpuGpuMaxDiff}{0.0001}
\newcommand{\CAHPSotalignCpuGpuCells}{9}
\newcommand{\CAHPSotalignCpuGpuMaxDiffOne}{0.069}

\newcommand{\CAHPIbotOrthoThousand}{0.0193}

\newcommand{\CAHPIbotWPThousand}{0.0227}

\newcommand{\CAHPIbotStructureThousand}{0.0218}

\newcommand{\CAHPIbotSotalignThousand}{0.0433}
\newcommand{\CAHPIbotOursZero}{0.0198}

\newcommand{\CAHPIbotOursTen}{0.0147}

\newcommand{\CAHPDinoOrthoThousand}{0.0052}

\newcommand{\CAHPDinoWPHundred}{0.0074}
\newcommand{\CAHPDinoWPThousand}{0.0048}

\newcommand{\CAHPDinoStructureThousand}{0.0050}

\newcommand{\CAHPDinoSotalignThousand}{0.0089}
\newcommand{\CAHPDinoOursZero}{0.0097}

\newcommand{\CAHPDinoOursTen}{0.0051}

\newcommand{\CANonlinPolarMean}{0.0098}

\newcommand{\CANonlinRidgeMean}{0.0079}

\newcommand{\CANonlinPooledMean}{0.0060}
\newcommand{\CANonlinPooledSD}{0.0031}

\newcommand{\CANonlinCells}{105}
\newcommand{\CANonlinRidgeBetterCells}{96}
\newcommand{\CANonlinRidgeGapRecovered}{49}

\newcommand{\CANonlinRidgeMlpMaxAbsDiff}{0.0002}

\newcommand{\CAHeadMean}{0.0068}

\newcommand{\CAHeadLinearMean}{0.0099}
\newcommand{\CAHeadRatio}{0.68}
\newcommand{\CAHeadBetterCells}{96}
\newcommand{\CAHeadFails}{0/105}
\newcommand{\CAHeadWorst}{0.0130}

\newcommand{\CAHeadAVGDelta}{+0.1}
\newcommand{\CAHeadIN}{13.2}

\newcommand{\CAHeadItoTGain}{8.7}
\newcommand{\CAHeadTtoIGain}{2.3}
\newcommand{\CAHeadItoTPairs}{12}
\newcommand{\CAHeadTtoIPairs}{11}
\newcommand{\CAHeadAVGOutside}{2}
\newcommand{\CAHeadSeedRatioBest}{0.37--0.41}
\newcommand{\CAHeadSeedRatioRest}{0.75--0.79}
\newcommand{\CAHeadNotBetterGap}{0.0012}
\newcommand{\CAHeadDistillSeconds}{144}
\newcommand{\CAHeadSinkhornItsLow}{366}
\newcommand{\CAHeadSinkhornItsHigh}{558}

\newcommand{\CAHeadPilotWorse}{0}

\newcommand{\CAMolPairs}{20}
\newcommand{\CAMolSeeds}{5}
\newcommand{\CAMolPoolN}{100,000}
\newcommand{\CAMolHalfN}{50,000}
\newcommand{\CAMolTestN}{3,300}
\newcommand{\CAMolTrainN}{26,407}
\newcommand{\CAMolCeilLow}{0.034}
\newcommand{\CAMolCeilHigh}{0.096}
\newcommand{\CAMolCeilMean}{0.056}
\newcommand{\CAMolCeilBelowFive}{9}
\newcommand{\CAMolCeilRone}{6.2}

\newcommand{\CAMolOursSeedZero}{0.484}
\newcommand{\CAMolOursLow}{0.484}
\newcommand{\CAMolOursHigh}{0.573}
\newcommand{\CAMolOursMean}{0.524}
\newcommand{\CAMolCellLow}{0.440}
\newcommand{\CAMolCellHigh}{0.605}
\newcommand{\CAMolPairMean}{0.510}
\newcommand{\CAMolOursRone}{0.02}
\newcommand{\CAMolPTwo}{15}
\newcommand{\CAMolPThreeMax}{0}
\newcommand{\CAMolPFourOrdered}{2}
\newcommand{\CAMolPFourRandom}{6}
\newcommand{\CAMolPFourViol}{12}
\newcommand{\CAMolCertCands}{40}
\newcommand{\CAMolAgreeMaxOurs}{0.0016}

\newcommand{\CAMolAgreeMaxSup}{0.0008}
\newcommand{\CAMolAgreeMaxRows}{8}
\newcommand{\CAMolObjOurs}{0.455}
\newcommand{\CAMolObjPair}{0.521}
\newcommand{\CAMolObjSup}{0.272}

\newcommand{\CAMolFitMinutesOurs}{104}
\newcommand{\CAMolFitMinutesPair}{97}

\newcommand{\CAMolDiagObjUp}{20}
\newcommand{\CAMolDiagSupObj}{0.271}
\newcommand{\CAMolDiagRefObj}{0.471}
\newcommand{\CAMolDiagSupFOS}{0.056}
\newcommand{\CAMolDiagRefFOS}{0.188}
\newcommand{\CAMolPooledBridgeObj}{0.505}
\newcommand{\CAMolPooledSupObj}{0.254}

\newcommand{\CAMolPooledSupFOS}{0.020}

\newcommand{\CAMolPooledRefFOS}{0.107}
\newcommand{\CAMolCocoBridgeObj}{0.605}
\newcommand{\CAMolCocoSupObj}{0.606}
\newcommand{\CAMolCocoBridgeFOS}{0.003}
\newcommand{\CAMolCocoSupFOS}{0.001}

\newcommand{\CAMolCocoRefFOS}{0.003}
\newcommand{\CAMolWeightLow}{0.80}
\newcommand{\CAMolWeightHigh}{1.00}

\newcommand{\CAMolWeightedRefFOS}{0.107}

\newcommand{\CAUnionPairs}{21}
\newcommand{\CAUnionN}{42,213}
\newcommand{\CAUnionHalfN}{21,106}

\newcommand{\CAUnionCeilMean}{0.005}

\newcommand{\CAUnionOursSeedZero}{0.028}
\newcommand{\CAUnionOursLow}{0.007}
\newcommand{\CAUnionOursHigh}{0.038}
\newcommand{\CAUnionOursCellHigh}{0.054}
\newcommand{\CAUnionOursFailsMax}{2}
\newcommand{\CAUnionOursFailsSeedZero}{0}
\newcommand{\CAUnionPairMean}{0.314}
\newcommand{\CAUnionPairFails}{19}
\newcommand{\CAUnionOursBelowPair}{21}
\newcommand{\CAUnionTransOursSeedZero}{0.020}
\newcommand{\CAUnionTransOursLow}{0.010}
\newcommand{\CAUnionTransOursHigh}{0.023}

\newcommand{\CAUnionTransPairMean}{0.302}
\newcommand{\CAUnionTransCeilMean}{0.003}
\newcommand{\CAUnionFitMinutesOurs}{53}
\newcommand{\CAUnionFitMinutesPair}{66}
\newcommand{\CAUnionSeeds}{5}
\newcommand{\CAUnionOursMean}{0.024}
\newcommand{\CAUnionAgreeOursLow}{0.005}
\newcommand{\CAUnionAgreeOursHigh}{0.054}

\newcommand{\CAXlingPairs}{25}
\newcommand{\CAXlingSeeds}{5}
\newcommand{\CAXlingPoolN}{60,000}
\newcommand{\CAXlingHalfN}{30,000}
\newcommand{\CAXlingHalfK}{30k}
\newcommand{\CAXlingTestN}{3,000}
\newcommand{\CAXlingTrainN}{7,000}
\newcommand{\CAXlingCeilLow}{0.0009}
\newcommand{\CAXlingCeilHigh}{0.0024}
\newcommand{\CAXlingCeilMean}{0.0015}

\newcommand{\CAXlingCeilRone}{92.9}
\newcommand{\CAXlingCeilRoneZtoE}{93.2}
\newcommand{\CAXlingOursSeedZero}{0.0018}
\newcommand{\CAXlingOursLow}{0.0017}
\newcommand{\CAXlingOursHigh}{0.0019}
\newcommand{\CAXlingOursMean}{0.0018}
\newcommand{\CAXlingOursCellHigh}{0.0030}
\newcommand{\CAXlingOursRone}{89.5}
\newcommand{\CAXlingOursRoneZtoE}{89.9}

\newcommand{\CAXlingPairMean}{0.277}
\newcommand{\CAXlingPairFails}{15}
\newcommand{\CAXlingPairPassed}{10}
\newcommand{\CAXlingPairRone}{35.7}
\newcommand{\CAXlingPairFailLow}{0.229}
\newcommand{\CAXlingPairFailHigh}{0.555}
\newcommand{\CAXlingPairPassHigh}{0.0027}
\newcommand{\CAXlingPTwo}{24}
\newcommand{\CAXlingPThreeMin}{25}
\newcommand{\CAXlingPFourOrdered}{17}

\newcommand{\CAXlingPFourViol}{8}
\newcommand{\CAXlingViolGap}{0.012}
\newcommand{\CAXlingAgreeFailMax}{0.0028}
\newcommand{\CAXlingAgreeFailRows}{14}
\newcommand{\CAXlingAgreePassMin}{0.077}
\newcommand{\CAXlingAgreeOursLow}{0.093}
\newcommand{\CAXlingAgreeOursHigh}{0.184}
\newcommand{\CAXlingAgreeSupLow}{0.086}
\newcommand{\CAXlingAgreeSupHigh}{0.183}
\newcommand{\CAXlingObjOurs}{0.313}
\newcommand{\CAXlingObjPair}{0.304}
\newcommand{\CAXlingObjSup}{0.324}
\newcommand{\CAXlingFitMinutesOurs}{51}
\newcommand{\CAXlingFitMinutesPair}{45}

\newcommand{\CACKAOutcome}{FAIL}
\newcommand{\CACKASubsets}{39}
\newcommand{\CACKASeeds}{3}

\newcommand{\CACKAMonoSeeds}{0}

\newcommand{\CACKAAboveSeeds}{0}
\newcommand{\CACKARho}{0.88}
\newcommand{\CACKARhoWithinMin}{0.88}
\newcommand{\CACKARhoWithinMax}{0.95}

\newcommand{\CACKAGapMeanAvg}{+0.020}
\newcommand{\CACKAGapBestAvg}{-0.132}
\newcommand{\CACKAAboveMeanCount}{20}
\newcommand{\CACKAMultiSubsets}{30}
\newcommand{\CACKAFullPooled}{0.600}
\newcommand{\CACKAFullBest}{0.718}

\newcommand{\CACKAFullBestName}{DINOv2-B}
\newcommand{\CACKAFullWorstName}{MAE-L (mean)}
\newcommand{\CACKAFullWorst}{0.087}

\newcommand{\CACKAWAboveSeeds}{3}
\newcommand{\CACKAWRho}{0.91}
\newcommand{\CACKAWFullPooled}{0.723}
\newcommand{\CACKAWFullBest}{0.716}

\newcommand{\CACKAWFullGrid}{0.0107}
\newcommand{\CACKAFullGrid}{0.2030}

\newcommand{\CACKAWSizeFour}{0.699}
\newcommand{\CACKAUSizeFour}{0.575}
\newcommand{\CACKAWSizeSeven}{0.720}
\newcommand{\CACKAUSizeSeven}{0.569}
\newcommand{\CACKAOneSeeds}{5}

\newcommand{\CACKAOnePooled}{0.733}
\newcommand{\CACKAOneBestVision}{0.718}
\newcommand{\CACKAOneBestText}{0.698}
\newcommand{\CACKAOneBestNative}{0.681}
\newcommand{\CACKAOneBestVisionName}{DINOv2-B}
\newcommand{\CACKAOneBestTextName}{Qwen3-Emb}
\newcommand{\CACKAOnePooledAboveVision}{5}
\newcommand{\CACKAOnePooledAboveText}{5}

\newcommand{\CAIdSubsets}{39}
\newcommand{\CAIdSeeds}{3}
\newcommand{\CAIdMaxDev}{8\times10^{-14}}
\newcommand{\CAIdTrunc}{0.0002}
\newcommand{\CAIdGammaMin}{1.000}
\newcommand{\CAIdGammaMax}{1.423}
\newcommand{\CAIdGammaTwo}{1.14}
\newcommand{\CAIdGammaFour}{1.23}
\newcommand{\CAIdGammaSeven}{1.36}
\newcommand{\CAIdGammaEleven}{1.41}
\newcommand{\CAIdBelowMean}{10}
\newcommand{\CAIdBelowMeanExplained}{10}
\newcommand{\CAIdMulti}{30}
\newcommand{\CAIdWAlphaDown}{21}
\newcommand{\CAIdWUp}{20}

\newcommand{\CASolverOutcome}{FAIL}

\newcommand{\CASolverSeeds}{5}

\newcommand{\CASolverWorseSentence}{Against our prediction, VecMap and MUSE are worse on the pooled spaces than on the native pairs}
\newcommand{\CASolverPassWord}{No}
\newcommand{\CASolverSolversWord}{seven}
\newcommand{\CASolverTwoFoldWord}{one}
\newcommand{\CASolverWithRefWord}{six}
\newcommand{\CASolverPooledMin}{0.107}
\newcommand{\CASolverPooledMax}{0.475}
\newcommand{\CASolverFailsTotal}{724}
\newcommand{\CASolverCellsTotal}{735}
\newcommand{\CASolverBestName}{mini-vec2vec}
\newcommand{\CASolverBestPooled}{0.107}
\newcommand{\CASolverBestNative}{0.230}

\newcommand{\CASolverBestFails}{94}
\newcommand{\CASolverBestCells}{105}

\newcommand{\CASolverWPNative}{0.146}
\newcommand{\CASolverWPNativeFails}{68}
\newcommand{\CASolverWPNativeCells}{105}
\newcommand{\CASolverWPPooled}{0.0099}

\newcommand{\CABudgetNativeSeeds}{3}
\newcommand{\CABudgetPFourGrid}{0.0110}

\newcommand{\CABudgetPOneGrid}{0.0217}

\newcommand{\CABudgetPOneFails}{3}
\newcommand{\CABudgetPOneCells}{105}
\newcommand{\CABudgetPOneOk}{102}

\newcommand{\CABudgetPShortGrid}{0.0311}

\newcommand{\CABudgetPShortFails}{24}

\newcommand{\CABudgetNOneFails}{58}
\newcommand{\CABudgetNOneCells}{63}

\newcommand{\CABudgetNOneFailPct}{92}
\newcommand{\CABudgetDefaultPooled}{0.0099}

\newcommand{\CABudgetDefaultNativeFailPct}{65}

\newcommand{\CABudgetRestartsFourFailPct}{62}
\newcommand{\CABudgetPredOne}{does not hold on the mean}
\newcommand{\CABudgetPredShort}{holds}
\newcommand{\CABudgetPredNative}{holds}

\definecolor{cvprblue}{rgb}{0.21,0.49,0.74}
\usepackage[breaklinks,colorlinks,allcolors=cvprblue]{hyperref}
\usepackage[capitalize,noabbrev]{cleveref}
\usepackage{float}
\usepackage{fontawesome5}
\hypersetup{pdftitle={From Pairwise to Collective: Encoder Collections Make Unpaired Vision-Language Alignment Reliable},pdfauthor={Xuanhua Yin; Shunqi Mao; Chuanzhi Xu; Boyu Wei; Wei Guo; Weidong Cai}}
\def\confName{CVPR}
\def\confYear{2027}
\title{From Pairwise to Collective: Encoder Collections\\Make Unpaired Vision-Language Alignment Reliable}
\author{Xuanhua Yin\footnotemark[1]\quad Shunqi Mao\quad Chuanzhi Xu\quad Boyu Wei\quad Wei Guo\quad Weidong Cai\\
{\normalsize School of Computer Science, The University of Sydney}\\
{\normalsize Sydney, NSW 2006, Australia}\\[3pt]
{\normalsize\href{https://xuanhuayin.github.io/ConsensusAlign/}{\faGlobe\enspace Project}\qquad\href{https://github.com/xuanhuayin/ConsensusAlign}{\faGithub\enspace Code}}}
\makeatletter
\patchcmd{\@maketitle}{\vspace*{12pt}}{\vspace*{1pt}}{}{}
\patchcmd{\@maketitle}{\vspace*{24pt}}{\vspace*{4pt}}{}{}
\patchcmd{\@maketitle}{\vskip .375in}{\vskip .125in}{}{}
\patchcmd{\abstract}{\vspace*{12pt}}{\vspace*{4pt}}{}{}
\makeatother
\begin{document}
\twocolumn[{%
\renewcommand\twocolumn[1][]{#1}%
\maketitle
\begin{center}
\centering
\captionsetup{type=figure}
\includegraphics[width=\linewidth]{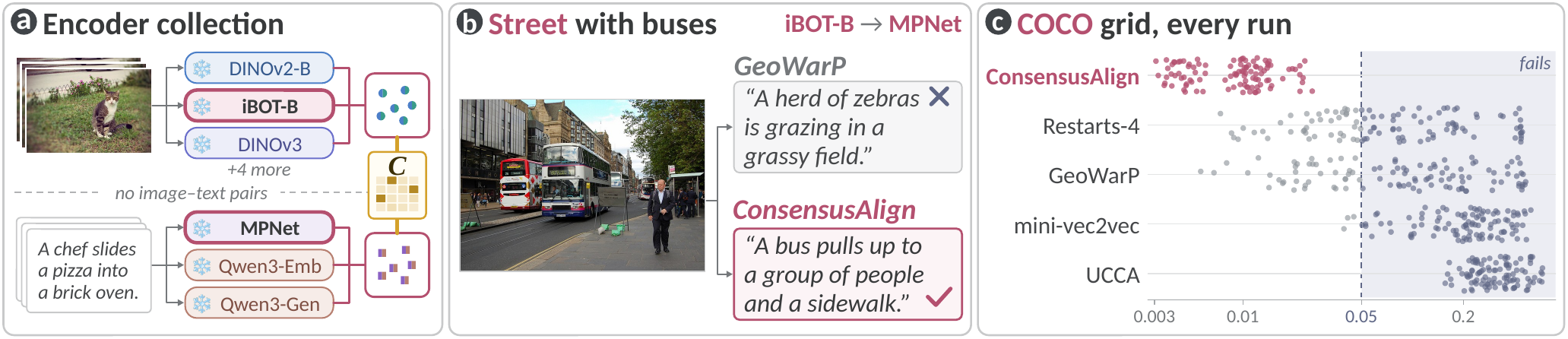}

\vspace{-10pt}
\captionof{figure}{\textbf{From pairwise to collective alignment.} (a) One bridge serves all encoder pairs. (b) COCO rank-1 retrieval for iBOT-B/MPNet; both captions belong to other images. (c) One dot per encoder pair and seed on the $7\times3$ grid; FOSCTTM $>0.05$ marks failure.}
\label{fig:teaser}
\end{center}

}]
\begingroup
\renewcommand{\thefootnote}{\fnsymbol{footnote}}
\footnotetext[1]{Email: \href{mailto:xuanhua.yin@sydney.edu.au}{\texttt{xuanhua.yin@sydney.edu.au}}.}
\endgroup
\begin{abstract}
Strong vision and text encoders can be trained independently, but their representations are not directly comparable. Connecting them usually requires paired image-text data, which may be unavailable. Unpaired alignment instead fits a map between a vision and a text encoder from feature-space geometry alone, enabling text-based retrieval and zero-shot classification without paired data. Treating each pair independently leads to fragile searches, repeated fitting, and uncertain map quality. A collection of encoders supplies additional supervision: all vision encoders embed the same images and all text encoders the same texts. Correspondences within each modality are therefore known exactly, while the cross-modal correspondence remains hidden. \method\ pools each modality into a shared space, fits one unpaired cross-modal bridge, and transfers it in closed form to every encoder pair. New encoders join through same-modality fitting without repeating the cross-modal search. On 21 vision-text encoder pairs and five seeds, pooled fitting aligns every pair, whereas pairwise fits fail in 68 of 105 pair-seed cells. Mean alignment error (FOSCTTM) falls from $\CACurrentWP$ to $\CAPooledMean$, and zero-shot accuracy nearly doubles on ten datasets. Gains persist across the tested within-modality pooling constructions with a WP bridge. New corpora reuse the bridge through native maps fitted to unlabeled target features from every source encoder. Agreement with separately fitted reference maps detects failures on COCO.
\end{abstract}

\section{Introduction}
\label{sec:intro}
Self-supervised vision encoders such as DINOv3, Franca and I-JEPA learn strong image representations without captions~\cite{simeoni2026dinov3,venkataramanan2026franca,assran2023ijepa,caron2021dino,he2022mae}, while sentence encoders provide text representations~\cite{reimers2019sentencebert}. Independent training leaves their coordinate systems unaligned. Text-based retrieval and zero-shot classification require a map from these vision features into a text encoder's space. This map provides a language interface. Such interfaces are usually built from paired image-text data. CLIP learns a joint image-text space from hundreds of millions of pairs~\cite{radford2021clip}, while ASIF connects frozen encoders through paired anchors~\cite{norelli2023asif}. Images and texts are often available as separate, unmatched collections.

Unpaired alignment fits this map without image-text pairs. GeoWarP~\cite{schnaus2026rosetta} aligns one vision encoder with one text encoder by Wasserstein Procrustes (WP), which alternates matching a batch of images to a batch of texts with fitting an orthogonal map to the matches. UCCA and vec2vec use other objectives~\cite{benari2026ucca,jha2025vec2vec}. These methods treat alignment as a search between two feature clouds, using distributional evidence from the chosen pair alone. The search is fragile: on our COCO grid of seven vision configurations and three text encoders, pairwise WP fails in 65\% of the pair-seed cells (FOSCTTM $>0.05$). The WP objective can be high even for incorrect image-text matches. Aligning every pair independently also requires $MN$ cross-modal fits for $M$ vision and $N$ text encoders.

A collection provides additional information through shared inputs within each modality. We run all vision encoders on the same images and all text encoders on the same texts. The known item correspondences let us pool the encoders' representations into one space per modality before cross-modal fitting. The resulting cross-modal search uses the combined geometry of the collection. In the base COCO collection, the pooled image and text spaces have more similar geometry than any native encoder pair. This observation motivates aligning the two pooled spaces (\cref{fig:teaser}).

We propose \method, which constructs these pooled spaces by concatenation and principal component analysis (PCA), then fits one unpaired WP bridge between them. A closed-form orthogonal Procrustes fit uses the known item correspondences to map each encoder into its own pooled space. Composing these maps with the bridge produces a native map for every vision-text pair, so inference uses only the two target encoders and one matrix. For heterogeneous collections, within-modality agreement weights reduce the contribution of outlying encoders. An optional diagnostic compares candidate matches with those from separately fitted pairwise reference maps.

We evaluated seven vision configurations and three text encoders on COCO, fitting on disjoint image and text halves so that no fitting image has its caption in the text pool. Across five seeds, each pooled fit aligns all 21 encoder pairs. Mean FOSCTTM falls from $\CACurrentWP$ to $\CAPooledMean$, and fitting from stored features takes roughly the time of one pairwise fit. Average zero-shot classification accuracy nearly doubles on ten datasets. The bridge supports new encoders and corpora through same-modality fitting without another cross-modal search. Our contributions are:
\begin{itemize}
    \item We identify the exact within-modality correspondences in a collection of frozen unimodal encoders as a supervision source for unpaired vision-language alignment.
    \item We propose \method, which pools each modality, fits one cross-modal bridge and constructs native maps for every encoder pair in closed form. One fit replaces one search per pair.
    \item Across 21 encoder pairs and five seeds, pooled fitting reduces alignment failures from 68/105 with pairwise WP to 0/105. Average zero-shot accuracy on ten datasets rises from 10.9\% to 19.6\%, and the bridge transfers to new encoders and corpora without another cross-modal search.
\end{itemize}

\section{Related Work}
\label{sec:related}
\begin{figure*}[t]
\centering
\includegraphics[width=\textwidth]{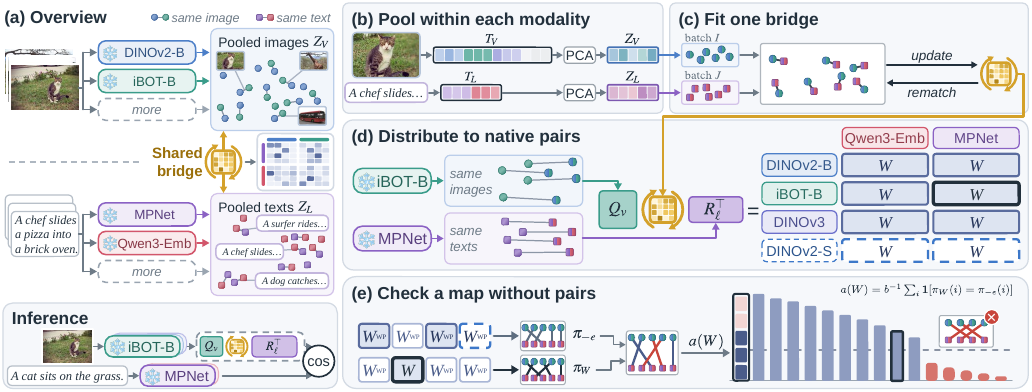}
\caption{\textbf{The \method\ pipeline.} Circles are images, squares are texts, colors mark encoders, and gold marks the bridge $C$. (a--b) Shared inputs permit modality pooling by \eqref{eq:pooling}, with unit weights or agreement weights \eqref{eq:member-weight}. (c) WP fits one bridge by alternating assignment and map updates \eqref{eq:pooled-wp}. (d) Closed-form factors give $W_{v\ell}=Q_vCR_\ell^\top$ \eqref{eq:spokes}; a new encoder needs only its own factor. (e) Optional diagnosis adds pairwise WP reference fits and compares their assignments with the candidate's (\cref{sec:quality}). Graphics are schematic. }
\label{fig:overview}
\end{figure*}

\paragraph{Multimodal Spaces and Language Interfaces.}
The standard language interface of a vision encoder is a joint image-text space. CLIP, ALIGN, LiT and SigLIP learn such spaces from hundreds of millions of pairs~\cite{radford2021clip,jia2021align,zhai2022lit,zhai2023siglip,tschannen2025siglip2}, and ImageBind, LanguageBind and related systems bind further modalities to them or fuse their features~\cite{girdhar2023imagebind,zhu2024languagebind,yin2026mind,yin2026brainvista}. Text-to-Concept and GPUA map a unimodal encoder into a pretrained vision-language model (VLM) and use its text tower as the interface~\cite{moayeri2023textconcept,yu2026gpua}. In both cases the interface rests on paired supervision. We build this interface from unimodal encoder collections and include a VLM as an optional member.

\paragraph{Aligning Two Frozen Encoders with Few or No Pairs.}
A cheaper route keeps two frozen encoders and fits a map between them. ASIF, Local CKA, STRUCTURE, SUE and SOTAlign do so from a few paired anchors or a paired teacher~\cite{norelli2023asif,maniparambil2024localcka,groger2025structure,yacobi2025sue,roschmann2026sotalign}. Without any pair, the map must come from the geometry of the two spaces alone. Related searches in bilingual lexicon induction use adversarial initialization, self-learning, Gromov-Wasserstein distances and Wasserstein Procrustes with Sinkhorn matching~\cite{conneau2018wordtranslation,artetxe2018vecmap,alvarez2018gromov,grave2019wasserstein,Cuturi2013Sinkhorn,Peyre2019ComputationalOT}, and can fail even when a linear map exists~\cite{hartmann2018unsupervised}. Recent work carries this approach across modalities: vec2vec and mini-vec2vec translate between text embedding spaces~\cite{jha2025vec2vec,dar2025minivec2vec}, Blind Match and RankAlign match class prototypes to class names~\cite{schnaus2025blindmatch,zhao2026rankalign}, MACK and MASK learn unpaired image-text matching through word-level bridges~\cite{huang2022mack,yin2026mask}, and GeoWarP and UCCA align a vision encoder with a text encoder~\cite{schnaus2026rosetta,benari2026ucca}. Encoder-level pairwise alignment fits one map from two feature distributions. We apply WP to two pooled spaces built from the encoder collection, so that one search serves every encoder pair.

\paragraph{Representation Similarity and Multi-Space Alignment.}
The Platonic Representation Hypothesis concerns shared geometry across independently trained models~\cite{huh2024platonic,balestriero2026platonic}; SVCCA, PWCCA and centered kernel alignment (CKA) measure representation similarity on corresponding inputs~\cite{raghu2017svcca,morcos2018insights,kornblith2019cka}. Relative representations, latent translation and atlas construction connect spaces through shared anchors~\cite{moschella2023relative,maiorca2023latent,crisostomi2024charts}. Multi-space methods include unsupervised multilingual hyperalignment and barycenter alignment, as well as generalized Procrustes constructions~\cite{alaux2019unsupervised,lian2020barycenter,achara2026multiway}. Combining several embeddings of the same words also produces NLP meta-embeddings~\cite{yin2016metaembeddings,bollegala2022metaembeddings}. Our setting supplies exact correspondences within each modality, where the encoders process the same inputs. We use these correspondences to pool each side and reserve the unpaired search for one cross-modal bridge. Concatenation-PCA, GPA and known-ID hyperalignment are alternative constructions under this resource.

\section{Method}
\label{sec:method}
\method\ uses known within-modality correspondences to construct shared image and text spaces, fits one WP bridge between them, and distributes it to native encoder pairs (\cref{fig:overview}). For heterogeneous collections, member agreement sets each encoder's weight. The fitted bridge can be reused for new encoders and corpora. A separate, optional diagnostic uses reference maps to assess alignment quality without image-text pairs.

\subsection{Problem Formulation}
\label{sec:problem}
Let $\V$ and $\Lset$ contain $M$ frozen vision encoders and $N$ frozen text encoders. Every vision encoder processes the same $n_x$ images, and every text encoder the same $n_y$ texts. Their features are $X_v\in\R^{n_x\times d_v}$ and $Y_\ell\in\R^{n_y\times d_\ell}$. We center each matrix using its column mean on the fitting data, the \emph{fitting mean}, and normalize its rows. Row $i$ of every $X_v$ describes the same image, and row $j$ of every $Y_\ell$ the same text. The images and texts come from disjoint corpus halves, so no fitting image has its caption among the fitting texts.

We seek a map $W_{v\ell}\in\R^{d_v\times d_\ell}$ for each requested pair, so that new images from $v$ can be compared with new texts from $\ell$.
Known row identities supervise each pool; cross-modal matches are inferred during bridge fitting.

\subsection{Shared Spaces and a Single Bridge}
\label{sec:pooling}
\paragraph{Constructing Shared Spaces.}
Native vectors can have different dimensions and coordinate axes. Horizontal concatenation combines their information along the known item rows before choosing a shared basis. We reduce the resulting representation to $d=768$ dimensions by PCA. On the vision side:
\begin{equation}
 T_V=\tfrac{1}{\sqrt M}\,[\,w_vX_v\,]_{v\in\V},\qquad
 Z_V=HT_VP_V,
\label{eq:pooling}
\end{equation}
where brackets denote horizontal concatenation, $H=I_{n_x}-\frac{1}{n_x}\mathbf1\mathbf1^\top$ centers columns, and $P_V$ contains the leading $d$ right singular vectors of $HT_V$, without whitening. The text side gives $Z_L$ in the same way. The base collection uses $w_v=w_\ell=1$, and \cref{sec:memberweights} gives weights for heterogeneous collections. The extra centering removes the column mean left by row normalization.

\paragraph{Aligning the Shared Spaces.}
\label{sec:bridge}
The two pools have separate coordinate systems and no observed image-text matches. We fit an orthogonal bridge $C\in O(d)$ between the two pools with Wasserstein Procrustes (WP)~\cite{grave2019wasserstein,schnaus2026rosetta}. For the centered, row-normalized pooled features $\widetilde Z_V,\widetilde Z_L$, WP optimizes:
\begin{equation}
 \max_{C\in O(d),\,\pi\in\Pi_b}
 \frac{1}{b}\sum_{i=1}^{b}\bigl(\widetilde Z_V[I]\,C\,\widetilde Z_L[J]^\top\bigr)_{i,\pi(i)},
\label{eq:pooled-wp}
\end{equation}
where $I,J$ index independent image and text batches of size $b$, and $\Pi_b$ is the set of permutations. WP alternates a Hungarian assignment for $\pi$ with an orthogonal Procrustes update for $C$. We write $\Hung(S)=\argmax_{\pi\in\Pi_b}\sum_i S_{i,\pi(i)}$ for the assignment of a score matrix $S$.
WP centers and row-normalizes the PCA features for this matching step. The native factors below use the raw PCA coordinates $Z_V,Z_L$. The two steps thus share a basis but use different normalization. Distribution retains the fitted matrix $C$, and native fitting means are stored separately for inference.
On COCO, both pooled and pairwise WP use 30 geometric centers, 30 initialization restarts, batches of 10,000 and 100 refinement updates. Pooling changes only the input spaces.

With $k$ paired anchors, we project them into the pools using the fixed PCA bases and include them in WP initialization and updates. All encoders share the same $k$ physical pairs. A VLM joins through its two towers, added to the corresponding pools before fitting a new bridge.

\paragraph{Native Maps and Inference.}
\label{sec:spokes}
The bridge acts on pooled coordinates, so each encoder needs a map into its pool. Known item identities give closed-form maps into each pool~\cite{Schonemann1966Procrustes,Higham1986Polar}:
\begin{equation}
 \begin{gathered}
 Q_v=\polar(X_v^\top Z_V),\qquad R_\ell=\polar(Y_\ell^\top Z_L),\\
 \boxed{W_{v\ell}=Q_vCR_\ell^\top}.
 \end{gathered}
\label{eq:spokes}
\end{equation}
For a thin SVD $A=U_A\Sigma_AV_A^\top$, $\polar(A)=U_AV_A^\top$. The factor $Q_v\in\R^{d_v\times d}$ maximizes $\operatorname{tr}(Q^\top X_v^\top Z_V)$ over semi-orthogonal $Q$, and $R_\ell$ is analogous. Their cross-covariances use observed row correspondences, so these fits need no estimated permutation. Each encoder requires one factor against its own pool, and all cross-modal search for alignment remains in $C$.

The bilinear score factorizes as $x_vW_{v\ell}y_\ell^\top=(x_vQ_vC)(y_\ell R_\ell)^\top$: every native pair uses the same bridge between its two pooled coordinates. At inference, we center $x_v,y_\ell$ with their fitting means and normalize their rows to $\hat x_v,\hat y_\ell$. Their score is $\cos(\hat x_vW_{v\ell},\hat y_\ell)$. We form $W_{v\ell}$ once; its rank is at most $d$. Other collection members affect the deployed pair through this matrix. Only the two target encoders run on a new query.
Cosine between $\hat x_vQ_vC$ and $\hat y_\ell R_\ell$ shares the deployed score's numerator but normalizes the projected vectors. The native-factor ablation in \cref{sec:exp-mechanisms} uses this factored cosine to compare distribution factors under fixed bridges.

\subsection{Weighting Heterogeneous Collections}
\label{sec:memberweights}
Different encoders can disagree about which images or texts are similar. We use agreement within each modality to reduce the influence of outlying members. The pooled Gram matrix shows how each member contributes before PCA:
\begin{equation}
 K_V=HT_VT_V^\top H=\frac{1}{M}\sum_{v\in\V}w_v^2K_v,\qquad K_v=HX_vX_v^\top H.
\label{eq:pooled-kernel}
\end{equation}
Independent orthogonal changes of native feature axes preserve each member's inner products and thus leave $K_V$ unchanged. Pooling combines how the members relate the same items, with weights controlling each contribution.

For centered Gram matrices, linear CKA is $\langle A,B\rangle_F/(\|A\|_F\|B\|_F)$. Expanding this inner product against any nonzero centered reference Gram matrix $K_\star$ on the same items gives:
\begin{equation}
 \operatorname{CKA}(K_V,K_\star)=\gamma\sum_v\alpha_v\operatorname{CKA}(K_v,K_\star),
\label{eq:pooled-cka}
\end{equation}
where $\alpha_v=w_v^2\|K_v\|_F/\sum_u w_u^2\|K_u\|_F$ sum to one and $\gamma=\sum_u w_u^2\|K_u\|_F/(M\|K_V\|_F)\ge1$ by the triangle inequality. The pooled kernel therefore has at least the norm-weighted mean CKA of its members. PCA retains its best rank-$d$ approximation $Z_VZ_V^\top$, and \cref{sec:exp-mechanisms} tests the geometry after this truncation. High-norm members contribute more to the mean, motivating weights that attenuate members with little shared structure:
\begin{equation}
 \kappa_v=\frac{1}{M-1}\sum_{u\ne v}\operatorname{CKA}(K_v,K_u),\quad
 w_v=\min(1,\kappa_v/\kappa_{\mathrm{med}})^{2},
\label{eq:member-weight}
\end{equation}
where CKA is linear centered kernel alignment~\cite{kornblith2019cka} and $\kappa_{\mathrm{med}}$ is the upper median member agreement. The weight enters \cref{eq:pooled-kernel} squared, attenuating outliers to the fourth power of their relative agreement. We compute text weights in the same way and use unit weights for fewer than three members.
The reference kernel $K_\star$ serves only the analysis. Weight computation uses within-modality features on known matching items. We estimate agreement on at most 8,000 shared fitting rows, selected with generator seed 0.

\subsection{Reusing the Bridge}
\label{sec:reuse}
Once the bridge is fitted, a new encoder only needs a map into its modality's pooled space. For a new corpus, we keep these coordinate systems and the bridge fixed while refitting the native factors.

A new vision encoder $u\notin\V$ embeds the fitting images and receives $Q_u=\polar(X_u^\top Z_V)$. Its maps $W_{u\ell}=Q_uCR_\ell^\top$ reuse the bridge and text factors. Text encoders attach likewise. Attachment keeps both source pools and the bridge fixed.

On a new corpus, we keep the source PCA bases, weights, bridge and concatenation means $\mu_V=\mathbf1^\top T_V/n_x$. Target features use their own fitting means and form $T'_V$ by \cref{eq:pooling}. We project them as $Z'_V=(T'_V-\mathbf1\mu_V)P_V$ and refit $Q'_v=\polar(X_v'^\top Z'_V)$. Text factors are refitted in the same way. These fits use only same-modality correspondences. Recomputing PCA on the target modalities would change the coordinate systems joined by $C$. Reuse preserves the source bases and adapts the native factors. Corpus adaptation extracts target features with every source encoder, then refits the native factors in the frozen coordinates.

\subsection{Assessing Maps Without Pairs}
\label{sec:quality}
The WP objective can be high even when image-text matches are wrong. To assess map quality without true pairs, we ask whether other encoder pairs support the same matches. This optional diagnostic uses separately fitted pairwise WP reference maps.

For $e=(v,\ell)$, let $\E_{-e}=\{(v',\ell'):v'\ne v,\ \ell'\ne\ell\}$ contain the pairs that share neither endpoint configuration with $e$. Reference families can overlap. Scores are indexed by the same item identities, so averaging takes place in a common item space despite differing feature dimensions. On common unpaired batches $(I,J)$ of size $b$, the reference score is:
\begin{equation}
 \bar S_{-e}=\frac{1}{|\E_{-e}|}\sum_{(v',\ell')\in\E_{-e}}X_{v'}[I]W^{\mathrm{WP}}_{v'\ell'}Y_{\ell'}[J]^\top.
\label{eq:reference}
\end{equation}
The assignment from \eqref{eq:reference} is $\pi_{-e}=\Hung(\bar S_{-e})$. A candidate $W$ gives $\pi_W=\Hung(X_v[I]WY_\ell[J]^\top)$ and agreement $a(W)=b^{-1}\sum_i\mathbf1[\pi_W(i)=\pi_{-e}(i)]$. Both use raw bilinear scores. Agreement measures support for the candidate's item matches from encoders outside that pair. On COCO, we fit 21 reference maps once per seed and share them across candidates. Each target uses its 12 encoder-disjoint references. These cheaper fits use batches of 5,000, 30 initialization restarts and 30 refinement updates. Their fitting and assignment costs are additional to alignment. \Cref{sec:exp-quality} evaluates how well agreement ranks candidate maps.

\section{Experiments}
\label{sec:experiments}
\begin{table*}[t]
\centering
\caption{\textbf{Alignment and downstream performance.} FOSCTTM mean$\pm$SD; downstream in \%. Bold: best zero-pair result. The paired reference uses 41,391 true training pairs.}
\label{tab:current-main}
\fontsize{10}{11}\selectfont
\setlength{\tabcolsep}{9pt}
\renewcommand{\arraystretch}{1.0}
\begin{tabular}{lcrcrrrr}
\toprule
& Fit & & Failed & & IN-1k & \multicolumn{2}{c}{Flickr30k R@1 $\uparrow$}\\
Method & seeds & FOSCTTM $\downarrow$ & cells $\downarrow$ & AVG10 $\uparrow$ & $\uparrow$ & I$\to$T & T$\to$I\\
\midrule
VecMap~\cite{artetxe2018vecmap} & 5 & $0.3352\!\pm\!0.0169$ & 105/105 & 3.5 & 0.4 & 0.3 & 0.3 \\
MUSE~\cite{conneau2018wordtranslation} & 5 & $0.4310\!\pm\!0.0190$ & 105/105 & 3.5 & 0.4 & 0.3 & 0.3 \\
GW + WP~\cite{alvarez2018gromov} & 5 & $0.4667\!\pm\!0.0396$ & 103/105 & 3.3 & 0.2 & 0.2 & 0.2 \\
Unaligned SCA~\cite{timilsina2024sca} & 5 & $0.4929\!\pm\!0.0038$ & 105/105 & 3.2 & 0.1 & 0.1 & 0.1 \\
vec2vec~\cite{jha2025vec2vec} & 5 & $0.4782\!\pm\!0.0242$ & 105/105 & 3.2 & 0.1 & 0.1 & 0.1 \\
mini-vec2vec~\cite{dar2025minivec2vec} & 5 & $0.2304\!\pm\!0.0082$ & 102/105 & 6.2 & 1.0 & 1.1 & 1.1 \\
GeoWarP~\cite{schnaus2026rosetta} & 5 & $0.1457\!\pm\!0.0151$ & 68/105 & 10.9 & 4.3 & 6.2 & 5.3 \\
Restarts-4 & 5 & $0.1274\!\pm\!0.0060$ & 65/105 & 11.6 & 4.4 & 5.9 & 5.2 \\
UCCA~\cite{benari2026ucca} & 5 & $0.3304\!\pm\!0.0139$ & 105/105 & 3.5 & 0.1 & 0.2 & 0.1 \\
\midrule
\rowcolor{oursrow} \textbf{\method} & 5 & $\mathbf{0.0099}\!\pm\!0.0040$ & \textbf{0/105} & \textbf{19.6} & \textbf{12.1} & \textbf{22.0} & \textbf{18.9} \\
\midrule
Paired orthogonal~\cite{Schonemann1966Procrustes} & 5 & $0.0036\!\pm\!0.0000$ & 0/105 & 24.7 & 19.3 & 36.0 & 29.2 \\
\bottomrule
\end{tabular}
\end{table*}

\begin{figure*}[t]
\centering
\includegraphics[width=\textwidth]{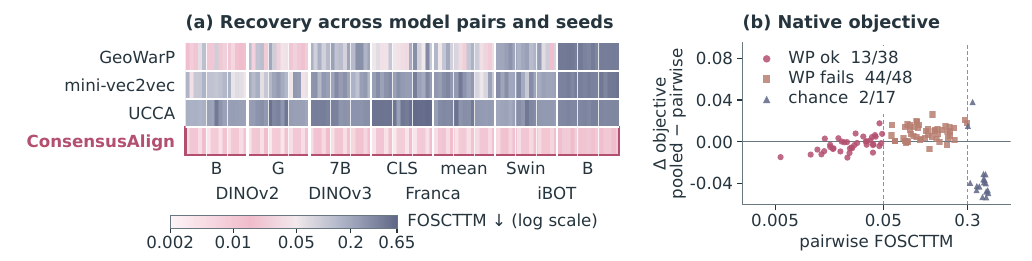}
\caption{\textbf{Pooling recovers the grid, while native objective preference depends on the failure regime.} (a) COCO validation FOSCTTM for seven vision configurations, three text encoders and five seeds (log color scale); \method\ uses the uniform $7\times3$ pool. (b) Native WP objective of $\polar(W_{v\ell})$ minus that of the pairwise map, against pairwise FOSCTTM (\CAMechCells\ cells with stored maps). The projected pooled map scores higher in \CAMechFailHigher/\CAMechFailN\ non-chance failures, but only \CAMechChanceHigher/\CAMechChanceN\ near-chance cells. Dashed lines mark the failure (0.05) and near-chance (0.3) cutoffs.}
\label{fig:current-recovery}
\end{figure*}

\subsection{Setup}
\label{sec:exp-setup}
\paragraph{Encoders and Data.}
We evaluate seven vision configurations and three text encoders on COCO, forming 21 vision-text pairs. The vision configurations are DINOv2-B/G~\cite{oquab2024dinov2}, DINOv3-7B~\cite{simeoni2026dinov3}, Franca-G with CLS and mean read-out~\cite{venkataramanan2026franca} and iBOT-Swin/B~\cite{zhou2022ibot}. The text encoders are MPNet~\cite{song2020mpnet}, Qwen3-Embedding-8B~\cite{zhang2025qwen3embedding} and Qwen3-8B generative features~\cite{yang2025qwen3}. The grid uses unimodally trained encoders. COCO train2014 is split into disjoint image and text halves of about 41k items each~\cite{lin2014coco,schnaus2026rosetta}, so no fitting image has its caption in the text half. The 40,504 COCO validation pairs serve for development and evaluation. The base method pools all ten encoders with unit weights and $d=768$.

\paragraph{Baselines.}
Zero-pair: GeoWarP~\cite{schnaus2026rosetta}, Restarts-4 (the best of four shorter WP fits by objective), mini-vec2vec~\cite{dar2025minivec2vec}, vec2vec~\cite{jha2025vec2vec}, UCCA~\cite{benari2026ucca}, unaligned shared-component analysis (SCA)~\cite{timilsina2024sca}, and the classical bilingual aligners VecMap~\cite{artetxe2018vecmap}, MUSE~\cite{conneau2018wordtranslation} and Gromov-Wasserstein followed by WP~\cite{alvarez2018gromov}. Few-pair: ASIF~\cite{norelli2023asif}, LocalCKA~\cite{maniparambil2024localcka}, SUE~\cite{yacobi2025sue}, STRUCTURE~\cite{groger2025structure} and SOTAlign~\cite{roschmann2026sotalign}, plus linear and orthogonal maps. Same-resource controls use these within-modality correspondences with alternative constructions: generalized Procrustes pooling, known-ID hyperalignment~\cite{alaux2019unsupervised,achara2026multiway} and a graph coordination of pairwise maps. All baseline scores come from our runs; UCCA uses the settings of its paper.

\paragraph{Protocols and Metrics.}
FOSCTTM is the fraction of texts ranked above an image's true caption, so $0$ is perfect and $0.5$ is chance, and a pair-seed cell fails when it exceeds $0.05$. Grid results report the mean and sample SD of five seed-level means over 21 pairs (\cref{tab:current-main}). Downstream evaluation uses the six vision configurations other than DINOv3, crossed with MPNet and Qwen3-Embedding-8B, for which evaluation features were prepared. AVG10 averages top-1 accuracy on ImageNet-1k, Food-101, Flowers102, DTD, EuroSAT, Oxford-IIIT Pets, Caltech101, FGVC Aircraft, STL-10 and GTSRB. We also report ImageNet-1k alone and R@1 on the 1k-image/5k-caption Flickr30k split~\cite{young2014flickr30k}. Few-pair comparisons use DINOv2-B/Qwen3-generative, $k$ known pairs and three seeds~\cite{schnaus2026rosetta} (\cref{fig:current-interfaces}a).

\subsection{Main Results}
\label{sec:exp-main}
Pooling improves both failed and usable interfaces (\cref{tab:current-main,fig:current-recovery}). The iBOT-B row shows independent fits remaining poor across its text encoders while pooling aligns the same frozen features. Even the worst pooled cell has FOSCTTM $0.0256$. On the \CADownSuccessCells\ downstream pair-seed cells where WP already succeeds, AVG10 rises from \CADownSuccessWPAVG\% to \CADownSuccessOursAVG\% and image-to-text R@1 from \CADownSuccessWPIT\% to \CADownSuccessOursIT\%, averaging first within each seed and then over five seeds. Classification improves on all ten datasets; STL-10 and Caltech101 gain most, while Aircraft and GTSRB remain weak even with paired orthogonal fitting.

The expensive search is shared across native interfaces. From stored features on one 20-core CPU at seed 0, the pooled pipeline takes \CADownPipelineMinutes\ min, compared with \CADownPairwiseMinutes\ min for one pairwise WP fit. Native-factor construction takes \CAPipeDistributeSeconds\ s; the factors are then reused across encoder pairs. Optional diagnosis requires additional reference fits.

\begin{figure*}[t]
\centering
\includegraphics[width=\textwidth]{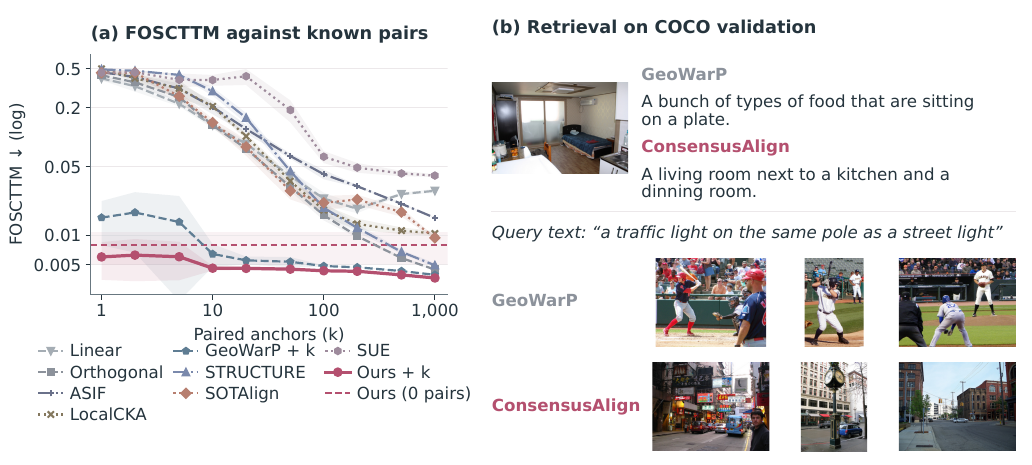}
\caption{\textbf{Few-pair alignment and bidirectional retrieval.} (a) Few-pair alignment on DINOv2-B/Qwen3-generative for all eight paired-anchor baselines and \method+$k$, three seeds per curve and budget, with shading for the sample SD. On this target pair, the zero-pair pooled fit (dashed) matches orthogonal fitting with 200 to 500 true pairs. (b) Retrieval with iBOT-B/MPNet on the COCO validation gallery for method-independent medoid queries: the rank-1 caption of an image query, and the top-3 images of a caption query.}
\label{fig:current-interfaces}
\end{figure*}

\label{sec:exp-fewpair}
On DINOv2-B/Qwen3-generative, paired anchors refine usable pairwise and pooled zero-pair interfaces (\cref{fig:current-interfaces}a). Anchors fix some cross-modal identities; pooling changes the geometry used to infer the remaining matches. Their combination gives the lowest error at every shared budget, also against an orthogonal map on the same pooled features and pairs. On the harder iBOT-B/MPNet pair, WP with 1,000 anchors reaches \CAHPIbotWPThousand, compared with \CAHPIbotOursZero\ for zero-pair pooling.

\subsection{Mechanism and Reliability}
\label{sec:exp-mechanisms}
The controls show that the benefit survives different uses of the within-modality correspondences (\cref{tab:collective-controls}). WP remains more reliable than the tested bridge replacements on the same concat-PCA pools. Propagating one pairwise bridge through the same correspondences gives a grid mean above 0.05 for 64 of 103 hub choices. Reducing each encoder to 768 dimensions before pairwise WP leaves 13 of 21 seed-0 cells failed. Projection size is flexible within the tested range: a matched two-seed sweep from 256 to 1,536 dimensions retains zero failures, with FOSCTTM between 0.0062 and 0.0085.

The geometry measurements explain why pooling helps this search. On COCO validation, the base pooled spaces have linear CKA \CAMechCKAPooled, above the \CAMechCKANativeMin--\CAMechCKANativeMax\ range of native pairs~\cite{kornblith2019cka,schnaus2026rosetta}. Across \CACKASubsets\ unit-weight vision-subset fits spanning \CACKASeeds\ seeds, CKA and grid-mean FOSCTTM have Spearman correlation $-\CACKARho$. Subset CKA tracks the member mean rather than the best member, consistent with the aggregate geometry in \cref{eq:pooled-cka}. The collection's composition therefore matters to the pooled search.

\begin{figure*}[t]
\centering
\includegraphics[width=\textwidth]{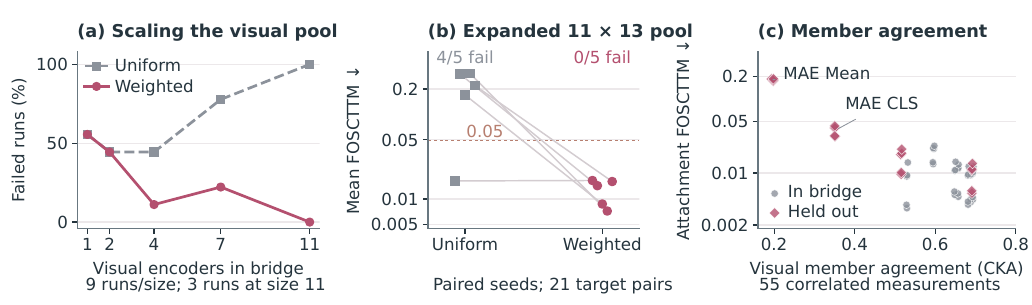}
\caption{\textbf{Agreement weighting prevents collapse of the tested $11\times13$ collection.} (a) Vision pools of growing size drawn from the eleven-encoder collection (text pool fixed), evaluated on the original 21 target pairs, where counts are failed runs (grid-mean FOSCTTM above $0.05$) over available runs. (b) The full $11\times13$ collection on five matched seeds, with and without agreement weights, with lines joining the same seed. (c) Member agreement (mean CKA with the other members) against attachment quality, eleven vision configurations over five seeds (55 points), where MAE is the outlier that weighting attenuates.}
\label{fig:current-scaling}
\end{figure*}

\label{sec:exp-scale}
This dependence on membership is visible in \cref{fig:current-scaling}. Adding encoders does not improve alignment monotonically. The two MAE configurations~\cite{he2022mae} have both low member agreement and poor attachment quality; unit weights let their geometry enter the pool at full strength. Agreement weighting attenuates those contributions and removes the observed collapses in the expanded collection. Its mean FOSCTTM remains \CAFullWeighted, against \CAPooledMean\ for the base pool. In this expansion, weighting improves tolerance of heterogeneous additions, while the smaller base collection remains the more accurate interface.

The changed geometry also reduces initialization sensitivity. Four pooled restarts reach FOSCTTM \CABudgetPFourGrid\ with no failed cell; one restart fails on \CABudgetPOneFails/\CABudgetPOneCells\ cells, against \CABudgetNOneFails/\CABudgetNOneCells\ for native fits on three seeds. The native-objective analysis in \cref{fig:current-recovery}b further distinguishes search failures from objective mismatch. We evaluate $\polar(W_{v\ell})$ to satisfy native WP's semi-orthogonal constraint when both native dimensions exceed $d$; the maximum FOSCTTM change is about $5\times10^{-6}$. In the \CAMechFailN\ non-chance failures, five native WP updates from the projected pooled map reach a higher objective than the original pairwise solution; updates from that solution change its objective by less than $0.0006$. The pooled construction supplies better native initializations.

\label{sec:exp-quality}
Near-chance failures also show objective mismatch: the native objective can prefer the incorrect map to the projected pooled map (\cref{fig:current-recovery}b). Further optimization of that objective need not improve semantic matching. Assignment agreement provides evidence from other encoder pairs when objective value and alignment quality disagree. On \CADiagMaps\ separately fitted COCO candidates, \CADiagFailures\ fail; agreement ranks failures against successes with AUROC \CADiagAUROCAgree, compared with \CADiagAUROCObj\ for the objective. This tests cross-encoder support for item matches, using the encoder-disjoint references defined in \cref{sec:quality}.

Distributing the bridge to native maps incurs a quality cost. On five matched seeds with fixed cached bridges, pooled-space retrieval using every encoder at inference reaches FOSCTTM \CANonlinPooledMean. Native-factor arms score $\cos(f_v(\hat x_v)C,g_\ell(\hat y_\ell))$: orthogonal factors on one encoder pair give \CANonlinPolarMean, while least squares and an MLP both give \CANonlinRidgeMean. These regressors use only the known native-to-pool correspondences. Improving the factors narrows the retrieval gap without changing the bridge, and all arms retain zero alignment failures.

\begin{table}[t]
\centering\small
\caption{\textbf{Collection and bridge controls.} Five seeds on 21 COCO pairs; FOSCTTM mean$\pm$SD and failed cells ($>0.05$).}
\label{tab:collective-controls}
\setlength{\tabcolsep}{2pt}
\renewcommand{\arraystretch}{1.0}
\begin{tabular*}{\columnwidth}{@{\extracolsep{\fill}}lrr@{}}
\toprule
Method & FOSCTTM $\downarrow$ & Failed\\
\midrule
\multicolumn{3}{@{}l}{\textit{Collective construction using WP}~\cite{schnaus2026rosetta}}\\
\textbf{\method} & $\mathbf{0.0099}\!\pm\!0.0040$ & \textbf{0/105} \\
GPA~\cite{achara2026multiway} & $0.0104\!\pm\!0.0045$ & \textbf{0/105} \\
Known-ID hyperalignment~\cite{alaux2019unsupervised} & $0.0118\!\pm\!0.0049$ & \textbf{0/105} \\
Graph coordination (control) & $0.0151\!\pm\!0.0064$ & \textbf{0/105} \\
\midrule
\multicolumn{3}{@{}l}{\textit{Bridge on fixed concat-PCA pools}}\\
VecMap~\cite{artetxe2018vecmap} & $0.3424\!\pm\!0.1019$ & 105/105 \\
MUSE~\cite{conneau2018wordtranslation} & $0.4750\!\pm\!0.2491$ & 105/105 \\
GW + WP~\cite{alvarez2018gromov} & $0.3811\!\pm\!0.1369$ & 105/105 \\
mini-vec2vec~\cite{dar2025minivec2vec} & $0.1072\!\pm\!0.0351$ & 94/105 \\
vec2vec~\cite{jha2025vec2vec} & $0.4358\!\pm\!0.0734$ & 105/105 \\
UCCA~\cite{benari2026ucca} & $0.2015\!\pm\!0.0084$ & 105/105 \\
\bottomrule
\end{tabular*}
\end{table}

\begin{table*}[t]
\centering
\caption{\textbf{Bridge reuse.} (a) New encoders: five seeds, mean$\pm$SD. (b) New corpora: five seeds. AVG10 in \%. Bold: best; tint: reuse.}
\label{tab:current-interfaces}%
\begin{minipage}[t]{.48\textwidth}
\centering\fontsize{10}{11}\selectfont
\textbf{(a) New vision encoders}\\[3pt]
\setlength{\tabcolsep}{2pt}
\renewcommand{\arraystretch}{1.0}
\begin{tabular*}{\linewidth}{@{\extracolsep{\fill}}lrr@{}}
\toprule
Encoder & FOSCTTM $\downarrow$ & AVG10 $\uparrow$\\
\midrule
DINOv2-S~\cite{oquab2024dinov2} & $0.0093\!\pm\!0.0036$ & 19.8 \\
I-JEPA-H~\cite{assran2023ijepa} & $0.0154\!\pm\!0.0052$ & 16.0 \\
MAE-L (CLS)~\cite{he2022mae} & $0.0372\!\pm\!0.0054$ & 13.9 \\
MAE-L (mean)~\cite{he2022mae} & $0.1808\!\pm\!0.0049$ & 7.0 \\
\bottomrule
\end{tabular*}
\end{minipage}\hfill%
\begin{minipage}[t]{.49\textwidth}
\centering\fontsize{10}{11}\selectfont
\textbf{(b) One COCO bridge, new corpora}\\[3pt]
\setlength{\tabcolsep}{2pt}
\renewcommand{\arraystretch}{1.0}
\begin{tabular*}{\linewidth}{@{\extracolsep{\fill}}lrr>{\columncolor{oursrow}}rr@{}}
\toprule
& \multicolumn{3}{c}{COCO FOSCTTM $\downarrow$} & Held-out\\
& Pairwise & \multicolumn{2}{c}{\small\method} & target\\
Corpus & WP~\cite{schnaus2026rosetta} & fit & reuse & (reuse)\\
\midrule
SPC & 0.2312 & \textbf{0.0104} & 0.0105 & 0.0164 \\
DCI & 0.3595 & 0.3573 & \textbf{0.0152} & 0.0413 \\
DOCCI & 0.3546 & 0.3068 & \textbf{0.0142} & 0.0335 \\
\bottomrule
\end{tabular*}
\end{minipage}%
\end{table*}

\begin{table}[t]
\centering
\caption{\textbf{Alignment beyond COCO.} FOSCTTM and assignment agreement (random: $0.0002$). Ours: \method. \emph{Paired} fits orthogonal maps with training pairs; WP~\cite{schnaus2026rosetta} uses seed 0 on TED.}
\label{tab:beyond}
\small
\setlength{\tabcolsep}{2pt}
\renewcommand{\arraystretch}{1.0}
\begin{tabular*}{\columnwidth}{@{\extracolsep{\fill}}lrrrr@{}}
\toprule
& \multicolumn{3}{c}{FOSCTTM $\downarrow$} & Agreement\\
\cmidrule(lr){2-4}
Corpus (pairs, seeds) & WP & Ours & Paired & Ours\\
\midrule
SPC+DCI+DOCCI (\CAUnionPairs, \CAUnionSeeds) & \CAUnionPairMean & $\mathbf{\CAUnionOursMean}$ & \CAUnionCeilMean & \CAUnionAgreeOursLow--\CAUnionAgreeOursHigh\\
TED En-Zh (\CAXlingPairs, \CAXlingSeeds) & \CAXlingPairMean & $\mathbf{\CAXlingOursMean}$ & \CAXlingCeilMean & \CAXlingAgreeOursLow--\CAXlingAgreeOursHigh\\
ROCOv2 (\CAROCOPairsPerSeed, \CAROCOSeeds) & \CAROCOFOSPair & \CAROCOFOSOurs & \CAROCOFOSSup & \CAROCOAgreeOurs\\
Molecules (\CAMolPairs, \CAMolSeeds) & \CAMolPairMean & \CAMolOursMean & \CAMolCeilMean & $\le$\CAMolAgreeMaxOurs\\
\bottomrule
\end{tabular*}
\end{table}

\subsection{Generalization and Limitations}
\label{sec:exp-transfer}
Reuse separates cross-modal search from native calibration. New encoders in \cref{tab:current-interfaces}a fit only their same-modality factor, keeping the source pools and bridge fixed. The cross-product and polar fit take \CAAttachSecondsLow--\CAAttachSecondsHigh~s after feature extraction. Attachment FOSCTTM averages three text targets per seed; AVG10 averages two. I-JEPA-H~\cite{assran2023ijepa} attaches below the failure threshold, while MAE remains weak. On SPC, DCI and DOCCI~\cite{krause2017paragraphs,urbanek2024dci,onoe2024docci} (8k--20k images before splitting), \cref{tab:current-interfaces}b compares fresh bridges with COCO-bridge reuse and refitted native factors, all scored on COCO validation. Fresh pooled fits fail on DCI and DOCCI, while reuse succeeds. The held-out column scores target pairs entering no fit: its mean stays below 0.05 on every corpus, while fresh pooled fits exceed 0.1 on the same target split for DCI and DOCCI. Fixed source coordinates let unlabeled target inputs calibrate native factors without a new cross-modal search. Pooling the corpus union also yields a successful fresh fit (\cref{tab:beyond}).

A separate English-Chinese experiment tests the construction with sentence encoders~\cite{conneau2018wordtranslation,reimers2020multilingual}. We fit a new bridge between five encoders per language on disjoint TED halves of \CAXlingHalfK\ sentences. The pooled fits recover the language grid over five seeds (\cref{tab:beyond}). At seed 0, English-to-Chinese R@1 is \CAXlingOursRone\%, close to the \CAXlingCeilRone\% paired reference. The same diagnostic separates failed maps, scoring at most \CAXlingAgreeFailMax, from successful maps, scoring at least \CAXlingAgreePassMin. Both pooling and diagnostic voting use item identities within each language, so their definitions carry over. A cutoff of $0.0207$, chosen using COCO evaluation labels, separates the seed-0 language fits but flags 13 of 20 successful pooled maps on the corpus union. Agreement's absolute scale therefore depends on the corpus and its reference ensemble.

Adding CLIP's towers~\cite{radford2021clip} tests whether the collection remains useful when a paired-pretraining resource is available. Across six self-supervised (SSL) vision configurations and five seeds, the new bridge raises AVG10 to \CAVLMCollAVG\%, against \CAVLMTTCAVG\% for Text-to-Concept~\cite{moayeri2023textconcept}. Image-to-text retrieval also improves, but text-to-image R@1 is lower (\CAVLMCollTI\% against \CAVLMTTCTI\%). Both methods use the SSL image tower and the same CLIP ViT-L/14 text tower at inference. For the evaluated collection, the preferred interface therefore depends on whether retrieval starts from an image query or a text query.

ROCOv2 radiology~\cite{ruckert2024rocov2} and molecule-text identify a boundary of the construction (\cref{tab:beyond}). The all-pairs orthogonal reference shows how well a linear interface aligns the frozen features when pairing is supplied. Pooling and pairwise WP both remain near chance, and the unpaired objective prefers these maps to the all-pairs orthogonal references. These cases differ from the native search failures recovered on COCO: the criterion itself rewards poor semantic matches. Assignment agreement stays near its random reference level, so the diagnostic flags these maps.

\FloatBarrier
\section{Conclusion}
\label{sec:conclusion}
\method\ uses exact within-modality correspondences to replace pairwise cross-modal searches with one WP bridge. Across five seeds, each pooled fit aligns all 21 COCO pairs, with two encoders and one matrix at inference. On the subset of successful pairwise WP fits, AVG10 rises by six percentage points, with gains in image-to-text and text-to-image retrieval. New encoders and corpora reuse it through same-modality fitting. The controls retain the gains across collective constructions and identify the bridge solver's role. Geometry and search experiments locate the improvement before native distribution; encoder agreement weights heterogeneous pools and ranks maps against reference fits. Alignment reliability depends on collection composition, the size of the unpaired fitting pools, and similar semantic coverage across modalities. On ROCOv2 and molecule-text, the unpaired objective favors near-chance maps over the paired references, limiting alignment even after pooling. Separating the shared cross-modal relation from native calibration lets the model library grow without repeating the search for each new interface.

\clearpage
\section*{LLM Usage Statement}
OpenAI Codex was used to assist with polishing the manuscript and monitoring experimental progress. Anthropic Claude was used to assist with code implementation and experiment execution. The authors developed the method, designed the experiments, verified all reported numbers against the raw experiment logs, and were responsible for all other aspects of the work. The authors take full responsibility for all claims, citations, and results.
{\small
\bibliographystyle{ieeenat_fullname}
\bibliography{main}}
\clearpage
\onecolumn
\appendix
\setcounter{table}{0}
\setcounter{figure}{0}
\setcounter{equation}{0}
\renewcommand{\thetable}{S\arabic{table}}
\renewcommand{\thefigure}{S\arabic{figure}}
\renewcommand{\theequation}{S\arabic{equation}}
\renewcommand{\theHtable}{supp.\arabic{table}}
\renewcommand{\theHfigure}{supp.\arabic{figure}}
\renewcommand{\theHequation}{supp.\arabic{equation}}
\makeatletter
\setlength{\@fptop}{0pt}
\setlength{\@fpsep}{16pt}
\setlength{\@fpbot}{0pt plus 1fil}
\makeatother
\raggedbottom
\section*{Supplementary Material}
This supplement gives the implementation of \method\ (\cref{supp:implementation}) and the proof of the pooled CKA identity (\cref{supp:cka-identity-proof}), the experimental setup (\cref{supp:setup}), complete results for every experiment of the main paper (\cref{supp:current-results,supp:recent-baselines,supp:mechanism,supp:collections,supp:corpora,supp:diagnosis}), the nonlinear extensions (\cref{supp:nonlinear}) and an expanded discussion of the limitations (\cref{sec:limitations}).

\section{Implementation Details}
\label{supp:implementation}
\paragraph{Native Features and Pooled Features.}
The released loader normalizes input embeddings; for multiple captions it normalizes each valid caption, averages them and normalizes again. Each encoder then estimates its own mean from its fitting half. We subtract that mean and normalize rows to obtain $X_v,Y_\ell$. The main B runs load the saved generative-text pooled alias through the same loader. Stored preprocessing conventions are part of the experiment record.

Within each modality we concatenate these matrices with scale $1/\sqrt{M}$, optionally multiplying each block by its amplitude weight. We subtract the concatenated matrix's own column mean and run a full thin SVD, retaining the first 768 right singular vectors. PCA does not whiten the output, and the resulting $Z_V,Z_L$ are not row-normalized when fitting native-to-pooled maps.

\paragraph{Member Agreement.}
The linear CKA of matrices $A,B$ uses column-centered rows:
\begin{equation}
 \operatorname{CKA}(A,B)=
 \frac{\|A_c^\top B_c\|_F^2}
 {\|A_c^\top A_c\|_F\,\|B_c^\top B_c\|_F}.
\end{equation}
Each modality uses a shared subset of at most 8,000 fitting rows, selected by a generator seeded with zero. Mean agreement is averaged over other members. The weight denominator is the sorted value at index $\lfloor M/2\rfloor+1$ in one-based indexing, that is, the upper median for even $M$. Weights are capped at one and squared before multiplying the features. Consequently their contribution to the concatenated Gram matrix is squared again. Uniform pooling is the seven-by-three base experiment; agreement weighting is the heterogeneous-collection extension. Neither uses image-text pairs for its weights.

\paragraph{One Cross-Modal Fitting Problem.}
The bridge uses GeoWarP's released WP implementation~\cite{schnaus2026rosetta}, with 30 geometric centers, 30 initialization restarts, batches of 10,000 and 100 refinement updates. The run uses seed $s+7$. The 30 restarts belong to the geometric initialization, whose correspondence estimates are averaged; they are not 30 complete WP fits followed by best-fit selection.

WP computes its own means of $Z_V,Z_L$ and centers and row-normalizes its inputs before matching and polar updates. The distribution step retains only the learned matrix $C$. Native maps are fitted to the raw PCA latents, as stated in the main paper. Inference uses the stored native fitting means. The five seeds are those of \cref{supp:setup}.

\paragraph{Full Procedure.}
\Cref{alg:consensusalign} lists the steps of the main paper's method; equation numbers refer to the main paper.

\begin{algorithm}[H]
\caption{\method: one pooled bridge, many native maps}
\label{alg:consensusalign}
\footnotesize
\begin{algorithmic}[1]
\Require Frozen encoder features with shared inputs within each modality
\State Store each encoder's fitting mean; center and normalize its rows
\State Set $w_v=w_\ell=1$, or compute agreement weights by main-paper Eq.~(6)
\State Pool each modality by Eq.~(1); obtain $Z_V,Z_L$
\State $C\gets\Call{WassersteinProcrustes}{Z_V,Z_L}$ by Eq.~(2)
\State $Q_v\gets\polar(X_v^\top Z_V)$ for each $v\in\V$
\State $R_\ell\gets\polar(Y_\ell^\top Z_L)$ for each $\ell\in\Lset$
\State $W_{v\ell}\gets Q_vCR_\ell^\top$ for each requested pair $(v,\ell)$ by Eq.~(3)
\State \Return $\{W_{v\ell}\}$ and fitting means; keep $P_V,P_L,\mu_V,\mu_L,C$ for reuse
\end{algorithmic}
\end{algorithm}

\paragraph{Rectangular Factors.}
For $A=X_v^\top Z_V\in\R^{d_v\times d}$, the thin SVD retains $r=\min(d_v,d)$ singular vectors: $U_A\in\R^{d_v\times r}$ and $V_A\in\R^{d\times r}$. The map $Q_v=U_AV_A^\top$ maximizes $\operatorname{tr}(Q^\top A)$ over the corresponding semi-orthogonal set. If $d_v\ge d$, then $Q_v^\top Q_v=I_d$; if $d_v\le d$, then $Q_vQ_v^\top=I_{d_v}$. The same applies to text maps. For raw features and $d_v>d$, this trace fit is not generally the minimizer of $\|X_vQ_v-Z_V\|_F^2$, since $\operatorname{tr}(Q_v^\top X_v^\top X_vQ_v)$ depends on $Q_v$. The factors use the raw PCA coordinates and the bridge uses normalized coordinates, as specified above.

The composition satisfies $\operatorname{rank}(W_{v\ell})\le\min(d_v,d_\ell,d)$. When both native dimensions exceed $d$, its rank is $d$. In the native-objective comparison, the SVD of $W_{v\ell}$ retains all $\min(d_v,d_\ell)$ singular-vector pairs, including those associated with zero singular values. Their bases are supplied by the numerical SVD, giving a semi-orthogonal completion. The deployed map remains $Q_vCR_\ell^\top$; this completion is used only for the native-objective test of \cref{supp:mechanism}, where the maximum FOSCTTM change is approximately $5\times10^{-6}$ over the 103 stored COCO cells.

\paragraph{Native and Factored Cosine Scores.}
For unit-normalized native inputs, let $s_{\rm native}=\cos(\hat x_vW_{v\ell},\hat y_\ell)$ and $s_{\rm factor}=\cos(\hat x_vQ_vC,\hat y_\ell R_\ell)$. When $d_\ell\ge d$, $R_\ell^\top R_\ell=I_d$ and $C$ is orthogonal, so
\begin{equation}
 \|\hat x_vW_{v\ell}\|_2=\|\hat x_vQ_v\|_2,\qquad
 s_{\rm native}=s_{\rm factor}\,\|\hat y_\ell R_\ell\|_2.
\end{equation}
The candidate-text norm can affect image-to-text rankings. It is constant for a fixed text query, so these two scores give the same text-to-image ranking. For $d_\ell=d$, the text factor is square orthogonal and the two scores coincide. Bidirectional retrieval uses one score matrix and its transpose. The distribution-factor ablation uses the factored cosine; \cref{sec:scoring} separately evaluates query direction and input normalization on cached maps.

\paragraph{Scope of the Pooled CKA Identity.}
The identity proved in \cref{supp:cka-identity-proof} concerns the centered concatenated Gram matrix before PCA. The lower bound applies to that pre-PCA kernel. Geometry after PCA and row normalization is measured empirically: \cref{sec:scoring} evaluates the inputs actually used by WP. The polar factors, CKA and WP's row normalization are invariant to a common positive scale of the concatenated features; the relative weights of individual encoder blocks still affect the construction.

\paragraph{Attachment and Corpus Reuse.}
Attachment holds the original pooled representations and $C$ fixed. A new encoder estimates its native mean from the same fitting inputs and fits one native-to-pooled factor. Its features do not enter the original bridge. For reported corpus reuse, the encoder collection and block order stay fixed. Target native means are recomputed; target features are projected with the source concatenation means and PCA bases. Target same-modality factors are then fitted against these projected target inputs, while $C$ remains the COCO bridge. The reported transfer uses the uniform source construction. This procedure preserves the source coordinate system; refitting PCA independently on the target corpus would require additional coordinate alignment. For $\mu'_V=\mathbf1_{n'_x}^\top T'_V/n'_x$, the target latent mean is $(\mu'_V-\mu_V)P_V$, which need not vanish. The native factor uses $X_v'^\top Z'_V$ with this frozen source offset. The held-out protocol of \cref{sec:r4holdout} keeps target evaluation pairs out of every fit; earlier evaluation that includes attachment inputs is transductive.

\paragraph{Optional Paired Anchors and VLM Members.}
The few-pair extension keeps the unpaired fitting projections fixed and maps $k$ known image-caption pairs through those projections. It supplies the resulting anchor coordinates to WP. The same $k$ physical item identities are shared across the encoder collection. Known pairs affect both geometric initialization and later polar updates. In a Procrustes update with $b'$ matched unpaired rows and $k>0$ anchors, the anchor cross-product is multiplied by $b'/k$, following the released solver. The anchor budget is separate from the zero-pair experiment.

The VLM-assisted extension includes CLIP's visual and text towers in their corresponding pools before fitting a new bridge. Although the fitting halves remain unpaired, the towers retain CLIP's paired-pretraining resource. These measurements are separate from unimodal-only results.

\paragraph{Diagnostic Scores.}
The agreement diagnostic is optional and requires reference fits in addition to the pooled bridge. Its COCO references are the cheaper pairwise WP fits: batches of 5,000, 30 geometric-initialization restarts and 30 refinement updates. All 21 reference fits are shared across targets. Each target uses the 12 edges sharing neither endpoint encoder. Their cost is not part of the reported alignment runtime. For seed 0 on one 20-core CPU with no other job running, the 21 reference fits take 136 minutes in total, 4.7 to 11.8 minutes each. Scoring all 21 targets then takes 3 minutes. Data loading adds 13 minutes. The diagnostic uses raw bilinear scores for both candidates and references, without mapped-output normalization or per-edge score standardization. The COCO cohort uses a 5,000-row batch with a generator seeded at $s+5$. Reference edges still share fitting data and some models with one another. The COCO evaluation scores 161 independently fitted candidates (\cref{supp:diagnosis}). ROCO uses ten encoder-disjoint reference edges per target on its six-by-three grid. The cutoff is calibrated on COCO, and \cref{tab:calibration} reports its cross-corpus behavior.

\paragraph{Feature Extraction and Fitting Cost.}
The reported CPU timings start from stored features and exclude encoder feature extraction and the optional diagnostic fits. Direct deployment of a frozen $W_{v\ell}$ runs only its two target encoders. Attaching a new encoder on the source items requires its features and the cached pooled coordinates. Recalibration on a new corpus constructs $T'_V,T'_L$ using every encoder in the source collection before projecting with the frozen PCA bases and refitting native factors. It therefore requires target features from the whole collection, even when only one final interface is requested.

\section{Proof of the Pooled CKA Identity}
\label{supp:cka-identity-proof}
\label{sec:cka-identity}
\begin{proposition}
Let $K_v$ be nonzero centered positive-semidefinite kernels on the same items, with weights $w_v>0$, and let $K_V=M^{-1}\sum_v w_v^2K_v$. For any nonzero centered positive-semidefinite kernel $K_L$ on these items and the Frobenius norm $\|\cdot\|$,
\begin{equation}
 \operatorname{CKA}(K_V,K_L)=\gamma\sum_v\alpha_v\operatorname{CKA}(K_v,K_L),
 \label{eq:cka-identity-supp}
\end{equation}
where $\alpha_v=w_v^2\|K_v\|/\sum_u w_u^2\|K_u\|$ sum to one and $\gamma=\sum_u w_u^2\|K_u\|/(M\|K_V\|)\ge1$. Equality $\gamma=1$ holds if and only if the member kernels are proportional. This identity concerns the full pooled kernel before PCA truncation.
\end{proposition}

\paragraph{Proof.} Write $\langle A,B\rangle=\operatorname{tr}(A^\top B)$ and $\|A\|=\langle A,A\rangle^{1/2}$, so that $\operatorname{CKA}(A,B)=\langle A,B\rangle/(\|A\|\|B\|)$. Linearity gives $\langle K_V,K_L\rangle=\frac1M\sum_v w_v^2\|K_v\|\,\|K_L\|\operatorname{CKA}(K_v,K_L)$. Dividing by $\|K_V\|\,\|K_L\|$ gives \eqref{eq:cka-identity-supp}. The triangle inequality gives $\|K_V\|\le\frac1M\sum_vw_v^2\|K_v\|$, hence $\gamma\ge1$, with equality exactly when the positive-semidefinite member kernels are proportional. Since their CKAs are nonnegative and $\sum_v\alpha_v=1$, the pooled CKA is at least the norm-weighted member mean. On the subsets below, rank-$d$ PCA truncation changes pooled CKA by at most \CAIdTrunc.

\paragraph{Numerical Verification.} On the \CAIdSubsets\ vision subsets of the collection-size study (\cref{sec:ckascale}; \CAIdSeeds\ seeds, COCO validation, the pooled text space as the other kernel), the directly computed pooled CKA and the right-hand side of \eqref{eq:cka-identity-supp} agree to $\CAIdMaxDev$, $\gamma$ lies in $\CAIdGammaMin$ to $\CAIdGammaMax$, and the pooled CKA is never below the norm-weighted member mean (\cref{tab:cka-identity,tab:cka-identity-weighted}). Of the \CAIdBelowMean\ unit-weight subsets whose pooled CKA falls below the plain member mean, \CAIdBelowMeanExplained\ have a largest-Gram-norm member with a below-median CKA, consistent with the norm weighting in \eqref{eq:cka-identity-supp}. The size means of $\gamma$ are \CAIdGammaTwo, \CAIdGammaFour, \CAIdGammaSeven\ and \CAIdGammaEleven\ for $M=2,4,7,11$. One expectation did not hold: with agreement weights the $\alpha$ of below-median members decreases in \CAIdWAlphaDown\ of the \CAIdMulti\ multi-member subsets and the pooled CKA rises in \CAIdWUp\ (predicted at least 25). Within-modality agreement and alignment with text therefore remain distinct quantities.

\begin{table}[H]\centering\small\setlength{\tabcolsep}{4pt}\caption{Pooled CKA identity, unit weights: size averages of $\gamma$, the pooled CKA (full pooled kernel), the norm-weighted member mean $\sum_v\alpha_v\mathrm{CKA}_v$ (so that pooled $=\gamma\times$ this column), the plain member mean and the best member.}\label{tab:cka-identity}\begin{tabular}{rrccccc}\toprule $M$ & subsets & $\gamma$ & pooled & $\sum_v\alpha_v\,\mathrm{CKA}_v$ & mean & best\\\midrule
1 & 9 & 1.000 & 0.535 & 0.535 & 0.535 & 0.535 \\
2 & 9 & 1.142 & 0.502 & 0.444 & 0.490 & 0.640 \\
4 & 9 & 1.233 & 0.575 & 0.477 & 0.559 & 0.701 \\
7 & 9 & 1.362 & 0.570 & 0.424 & 0.547 & 0.706 \\
11 & 3 & 1.412 & 0.600 & 0.425 & 0.548 & 0.718 \\
\bottomrule\end{tabular}\end{table}

\begin{table}[H]\centering\small\setlength{\tabcolsep}{4pt}\caption{As \cref{tab:cka-identity}, with agreement weights.}\label{tab:cka-identity-weighted}\begin{tabular}{rrccccc}\toprule $M$ & subsets & $\gamma$ & pooled & $\sum_v\alpha_v\,\mathrm{CKA}_v$ & mean & best\\\midrule
1 & 9 & 1.000 & 0.533 & 0.533 & 0.533 & 0.533 \\
2 & 9 & 1.142 & 0.501 & 0.444 & 0.490 & 0.640 \\
4 & 9 & 1.137 & 0.699 & 0.620 & 0.558 & 0.699 \\
7 & 9 & 1.140 & 0.720 & 0.632 & 0.546 & 0.704 \\
11 & 3 & 1.132 & 0.723 & 0.639 & 0.547 & 0.716 \\
\bottomrule\end{tabular}\end{table}

\section{Experimental Setup}
\label{supp:setup}
\paragraph{Data and Splits.}
We use the stored GeoWarP embeddings and its disjoint-split protocol~\cite{schnaus2026rosetta}. A seeded permutation divides the 82,783 COCO train2014 items into 41,391 image items and 41,392 different text items, and all encoders use the same permutation. The 40,504 COCO val2014 pairs form the evaluation set. For items with several captions, each caption vector is normalized, valid captions are averaged and the average is normalized again. Each encoder's fitting mean is estimated on its own fitting half and reused at evaluation. Mapped image vectors are normalized again before they are compared with text vectors, because a rectangular map can change vector lengths.

\begin{table}[H]
\centering
\small
\setlength{\tabcolsep}{3pt}
\begin{tabular}{p{0.82\columnwidth}r}
\toprule
Frozen representation & Dim.\\
\midrule
DINOv2 ViT-B/14, 224, token mean & 768\\
DINOv2 ViT-g/14, 224, token mean & 1536\\
DINOv3 ViT-7B/16, 512, token mean & 4096\\
Franca ViT-g/14, LAION, CLS token & 1536\\
Franca ViT-g/14, LAION, token mean & 1536\\
iBOT Swin-T, token mean & 768\\
iBOT ViT-B/16, token mean & 768\\
\midrule
all-mpnet-base-v2 & 768\\
Qwen3-Embedding-8B & 4096\\
Qwen3-8B, generative token pooling & 4096\\
\bottomrule
\end{tabular}
\caption{The complete alignment grid. Dimensions were checked against the stored COCO embedding tensors. Franca's two outputs are read-outs of the same model.}
\label{tab:supp-encoders}
\end{table}

\paragraph{Encoders.}
\Cref{tab:supp-encoders} lists the grid. DINOv2 and Franca token means include the class, register and patch tokens, and iBOT averages its last-layer tokens. Qwen3 generative features average the hidden states of up to 128 greedily decoded tokens. All results use the released stored vectors. The larger collection of Sec.~4.3 of the main paper adds four vision configurations, MAE-L with CLS and mean read-out~\cite{he2022mae}, I-JEPA-H~\cite{assran2023ijepa} and DINOv2-S, and ten text encoders: BGE-base, BGE-large, Contriever, DistilRoBERTa, E5-large-v2, GTE-large, GTR-T5-large, MiniLM-L6, MiniLM-L12 and Qwen3-Embedding-0.6B.

\paragraph{Seeds and Pairwise WP.}
The five seeds come from \texttt{derive\_seeds(42,5)} in the released WP code and are numbered 0 to 4. A seed fixes the disjoint split and the randomness of every fit. Pairwise WP uses the defaults of the release: 30 clusters, 30 initialization restarts, 100 refinement steps and batches of 10,000.

\paragraph{Downstream Protocol.}
AVG10 averages zero-shot top-1 accuracy over ImageNet-1k, Caltech101, STL-10~\cite{Deng2009ImageNet,FeiFei2004Caltech101,Coates2011STL10}, Food-101, Flowers102, Pets, Aircraft~\cite{Bossard2014Food101,Nilsback2008Flowers102,Parkhi2012Pets,Maji2013Aircraft}, DTD, EuroSAT and GTSRB~\cite{Cimpoi2014DTD,Helber2019EuroSAT,Stallkamp2011GTSRB}. We use the test sets where provided, the ImageNet validation set, DTD partition 1 and the full EuroSAT and Caltech101 collections. Each class uses 18 prompt templates whose vectors are normalized, averaged and normalized again. Method averages weight twelve pairs equally: the six vision configurations other than DINOv3, crossed with MPNet and Qwen3-Embedding-8B. Flickr30k uses the Karpathy 1k test split of 1,000 images with five captions each~\cite{young2014flickr30k}. Image-to-text recall asks whether any of an image's captions is in the top $k$ of 5,000 captions, and text-to-image recall asks whether a caption's image is in the top $k$ of 1,000 images.

Checkpoint identifiers and the exact class templates are given in \cref{sec:checkpoint-prompts}.

\paragraph{Record Selection.} We canonicalize the generative Qwen text alias, retain the pre-reduced alias when both forms occur, remove exact repeated records, and require complete 21-pair alignment or twelve-pair downstream seeds. Thus each logical target has one value.

\paragraph{Baseline Resources.} WP and UCCA use their released implementations; UCCA retains the official fixed PCA-10/output-2 configuration. Our GeoWarP reproduction gives FOSCTTM $\CACurrentWP$, close to the authors\textquotesingle\ reported $0.154$. Unaligned SCA uses the authors\textquotesingle\ word-alignment code with a 256-dimensional shared component, batch size 32 and five epochs of one million samples, scored after the final epoch. The mini-vec2vec and vec2vec entries use the adaptations released with GeoWarP, including rectangular maps and adaptation-specific optimization. Few-pair baselines use the adaptations in the same release. VecMap, MUSE and GW are our adaptations of the bilingual-lexicon methods to the encoder grid. Restarts-4 selects among four shorter pairwise fits using the unpaired objective. Generalized Procrustes and known-ID hyperalignment are same-input multiway controls. Graph coordination fits all pairwise maps and makes them consistent through the shared encoders.

\paragraph{GPUA Implementation.} Our port calls the official geometry-preserving alignment and pseudo-label functions. It concatenates DINOv2-B and CLIP visual features for the query representation, uses CLIP class prototypes, and uses the evaluation images as its unlabeled fitting pool. Labels are used for scoring. It is run with one seed and uses CLIP\textquotesingle s paired-pretrained image tower, so it is not a zero-pair method.

\subsection{Checkpoint Identifiers and Class Prompts}
\label{sec:checkpoint-prompts}
The additional English text checkpoints are listed in \cref{tab:extra-checkpoints}. The E5 prefix is followed by one space before the text; the other listed models receive the text without this prefix. The TED English subset uses MPNet and the BGE-large, E5-large-v2, GTE-large and MiniLM-L12 checkpoints. The molecular text subset uses MPNet, Qwen3-Embedding-0.6B, BGE-large and E5-large-v2. The five Chinese checkpoints and the additional domain-specific checkpoints appear in \cref{tab:domain-checkpoints}.

\begin{table}[H]\centering\small\setlength{\tabcolsep}{5pt}
\caption{Additional English text checkpoints used for the expanded pool and attachment. Prefix denotes text prepended by the extraction code.}
\label{tab:extra-checkpoints}
\begin{tabular}{lll}\toprule
Encoder & Checkpoint identifier & Prefix\\\midrule
minilm-l6 & \texttt{sentence-transformers/all-MiniLM-L6-v2} & None\\
minilm-l12 & \texttt{sentence-transformers/all-MiniLM-L12-v2} & None\\
distilroberta & \texttt{sentence-transformers/all-distilroberta-v1} & None\\
contriever & \texttt{facebook/contriever} & None\\
bge-base & \texttt{BAAI/bge-base-en-v1.5} & None\\
bge-large & \texttt{BAAI/bge-large-en-v1.5} & None\\
gte-large & \texttt{thenlper/gte-large} & None\\
gtr-t5-large & \texttt{sentence-transformers/gtr-t5-large} & None\\
e5-large-v2 & \texttt{intfloat/e5-large-v2} & \texttt{query: }\\
qwen3-embedding-0.6b & \texttt{Qwen/Qwen3-Embedding-0.6B} & None\\
\bottomrule\end{tabular}\end{table}

MAE-L CLS and mean use \path{facebook/vit-mae-large} with mask ratio zero. I-JEPA-H uses \path{facebook/ijepa_vith14_1k}. DINOv2-S uses the \texttt{dinov2\_vits14} entry point at the same pinned DINOv2 code revision as B and G. Their output dimensions are 1,024, 1,280 and 384, respectively. MPNet is \path{sentence-transformers/all-mpnet-base-v2}; the base Qwen checkpoints are \path{Qwen/Qwen3-Embedding-8B} and \path{Qwen/Qwen3-8B}. The generative read-out is the latter's generated-token pooling, as specified above.

\begin{table}[H]\centering\small\setlength{\tabcolsep}{5pt}
\caption{Additional Chinese, radiology-text and molecule checkpoints. Molecular ChemBERTa outputs average the last hidden states over non-padding tokens; MoLFormer uses its pooler output.}
\label{tab:domain-checkpoints}
\begin{tabular}{ll}\toprule
Encoder & Checkpoint identifier\\\midrule
bge-base-zh & \texttt{BAAI/bge-base-zh-v1.5}\\
bge-large-zh & \texttt{BAAI/bge-large-zh-v1.5}\\
text2vec-zh & \texttt{shibing624/text2vec-base-chinese}\\
m3e-base & \texttt{moka-ai/m3e-base}\\
gte-base-zh & \texttt{thenlper/gte-base-zh}\\
PubMedBERT & \texttt{NeuML/pubmedbert-base-embeddings}\\
\midrule
chemberta-10m-mtr & \texttt{DeepChem/ChemBERTa-10M-MTR}\\
chemberta2-mtr & \texttt{DeepChem/ChemBERTa-77M-MTR}\\
chemberta2-mlm & \texttt{DeepChem/ChemBERTa-77M-MLM}\\
chemberta-zinc & \texttt{seyonec/ChemBERTa-zinc-base-v1}\\
MoLFormer-XL & \texttt{ibm/MoLFormer-XL-both-10pct}\\
\bottomrule\end{tabular}\end{table}

\paragraph{Class Prompts and the Downstream Subset.}
The twelve downstream targets are the six base vision configurations other than DINOv3, crossed with MPNet and Qwen3-Embedding-8B. This is the collection for which the ten-dataset and Flickr30k evaluation features were prepared. Attachment AVG10 uses the same two text targets, and the VLM comparison uses the same six SSL vision configurations. The 18 class templates are listed in \cref{tab:class-templates}; \texttt{\{\}} is replaced with the stored class name. Each prompt vector is normalized, the 18 vectors are averaged and the result is normalized again.

\begin{table}[H]\centering\small\setlength{\tabcolsep}{5pt}
\caption{The 18 classification templates used by AVG10. Each row pairs the indefinite-article and definite-article forms.}
\label{tab:class-templates}
\begin{tabular}{ll}\toprule
Templates 1--9 & Templates 10--18\\\midrule
a photo of a \{\}. & a photo of the \{\}.\\
a blurry photo of a \{\}. & a blurry photo of the \{\}.\\
a black and white photo of a \{\}. & a black and white photo of the \{\}.\\
a low contrast photo of a \{\}. & a low contrast photo of the \{\}.\\
a high contrast photo of a \{\}. & a high contrast photo of the \{\}.\\
a bad photo of a \{\}. & a bad photo of the \{\}.\\
a good photo of a \{\}. & a good photo of the \{\}.\\
a photo of a small \{\}. & a photo of the small \{\}.\\
a photo of a big \{\}. & a photo of the big \{\}.\\
\bottomrule\end{tabular}\end{table}

\subsection{Control Protocols}
\label{sec:control-protocols}
\paragraph{Graph Coordination.}
The graph control in Table~2 of the main paper is the final arm of the multi-edge procedure below. All 21 native pairs first receive a cheap WP fit with batches of 5,000, 30 geometric-initialization restarts and 30 refinement updates. DINOv2-B and MPNet define canonical coordinates. Known within-modality item IDs give
\begin{equation}
 \begin{gathered}
 A_v=\polar(X_v^\top X_c),\qquad B_\ell=\polar(Y_c^\top Y_\ell),\\
 H_{v\ell}=A_v^\top W_{v\ell}^{\rm WP}B_\ell^\top.
 \end{gathered}
\end{equation}
Each $H_e$ is scored by the WP matching objective on the canonical features, using one 10,000-row batch sampled with seed 0. With $o_{\max}=\max_e o_e$ and $\tau=0.02$, the retained set is $\mathcal G_\tau=\{e:o_e\ge o_{\max}-\tau\}$. Its canonical map is
\begin{equation}
 G=\polar\!\left(\sum_{e\in\mathcal G_\tau}\omega_e H_e\right),\qquad
 \omega_e=\frac{o_e-o_{\max}+\tau}{\sum_{j\in\mathcal G_\tau}(o_j-o_{\max}+\tau)}.
\end{equation}
Two synchronization rounds initialize every native edge at $A_vGB_\ell$, apply 20 Hungarian/polar updates on 5,000-item batches, then lift the maps, rescore them and recompute $G$ by the same rule. Refinement round $t\in\{0,1\}$ uses seed $s+t$.

Six further rounds use shared pseudo-pairs. A generator seeded with $s+11$ draws image and text batches of 10,000 items. The average native score is
\begin{equation}
 S=\frac{1}{21}\sum_{v,\ell}X_v[I]A_vGB_\ell Y_\ell[J]^\top.
\end{equation}
One Hungarian assignment $\pi$ is used by every edge. Each edge is refitted as $\widehat W_{v\ell}=\polar(X_v[I]^\top Y_\ell[J]_\pi)$, where the subscript permutes text rows, and the next canonical map is $G=\polar(21^{-1}\sum_{v,\ell}A_v^\top\widehat W_{v\ell}B_\ell^\top)$. The final maps are $A_vGB_\ell$. The six rounds are fixed; there is no label-based stopping or selection. This control uses the same encoder features and within-modality correspondences as pooling, but fits all native edges and coordinates their maps.

\paragraph{Restarts-4.}
This baseline runs four complete cheap pairwise WP fits, each with 30 geometric centers, 30 initialization restarts, 5,000-item batches and 30 refinement updates. Fit $k\in\{0,1,2,3\}$ uses seed $s+7+101k$. The fit with the largest unpaired WP objective is retained. Objective selection uses one 10,000-row batch with generator seed 0. These four complete fits are distinct from the single pooled fit with four geometric-initialization restarts in \cref{supp:mechanism}.

\paragraph{Resources in the Few-Pair Comparison.}
The curves use the same target pair, the same $k$ physical image-caption anchors and the same three seeds. Native pairwise baselines use the two target encoders, while \method\ uses the collection and the unpaired fitting halves. The pooled-feature Orthogonal control uses the same collection, PCA bases, native factors and anchors as \method\ + $k$, but fits its bridge from the anchors alone. This control holds the encoder and data resources fixed while varying bridge fitting. The full-collection controls in Table~2 of the main paper compare alternative uses of the within-modality correspondences without anchors.

\section{Complete COCO Results}
\label{supp:current-results}
\Cref{tab:current-seeds,tab:current-pairs} give the seed-level and per-pair numbers behind Table~1 of the main paper. \Cref{tab:current-classical} adds the classical bilingual-lexicon methods adapted to the grid, \cref{tab:current-extra-ablations} further ablations of the pooled fit, and \cref{tab:karpathy} retrieval on the Karpathy COCO test split.
\begin{table}[H]
\centering\small
\caption{Complete per-seed 21-target means. Columns S0 to S4 follow the five seed IDs listed in the implementation section. Missing seeds are not imputed. Bold: best zero-pair mean per seed, including rounded ties.}
\label{tab:current-seeds}
\setlength{\tabcolsep}{4pt}
\begin{tabular}{lrrrrrr}
\toprule
Method & S0 & S1 & S2 & S3 & S4 & Failures \\
\midrule
GeoWarP~\cite{schnaus2026rosetta} & 0.1686 & 0.1269 & 0.1401 & 0.1466 & 0.1463 & 68/105 \\
Restarts-4 & 0.1248 & 0.1296 & 0.1361 & 0.1200 & 0.1265 & 65/105 \\
mini-vec2vec~\cite{dar2025minivec2vec} & 0.2382 & 0.2316 & 0.2269 & 0.2369 & 0.2181 & 102/105 \\
UCCA (official 10/2)~\cite{benari2026ucca} & 0.3390 & 0.3092 & 0.3429 & 0.3236 & 0.3373 & 105/105 \\
vec2vec~\cite{jha2025vec2vec} & 0.5069 & 0.4522 & 0.4693 & 0.5005 & 0.4622 & 105/105 \\
ConsensusAlign & 0.0055 & 0.0116 & 0.0119 & 0.0060 & 0.0147 & 0/105 \\
ConsensusAlign + 10 pairs & 0.0056 & 0.0061 & 0.0057 & 0.0061 & 0.0056 & 0/105 \\
ConsensusAlign + head & \textbf{0.0020} & 0.0091 & \textbf{0.0093} & \textbf{0.0025} & \textbf{0.0110} & 0/105 \\
GPA + WP & 0.0117 & \textbf{0.0054} & 0.0147 & 0.0058 & 0.0142 & 0/105 \\
Known-ID hyperalignment & 0.0136 & 0.0063 & 0.0163 & 0.0068 & 0.0159 & 0/105 \\
Graph coordination & 0.0065 & 0.0134 & 0.0170 & 0.0243 & 0.0141 & 0/105 \\
\bottomrule
\end{tabular}
\end{table}

\begin{table}[H]
\centering\small
\caption{Per-target FOSCTTM, mean and sample SD over five seeds. Bold: best zero-pair mean per row, including rounded ties; +10 is a separate reference. Main-table SD is over seed-level grid means.}
\label{tab:current-pairs}
\setlength{\tabcolsep}{4pt}
\begin{tabular}{lrrrrrr}
\toprule
Target pair & WP & Restarts-4 & mini & UCCA~\cite{benari2026ucca} & Ours & Ours+10 \\
\midrule
DINOv2-B / MPNet & $0.043\!\pm\!0.048$ & $0.013\!\pm\!0.007$ & $0.109\!\pm\!0.047$ & $0.164\!\pm\!0.006$ & $\mathbf{0.009}\!\pm\!0.004$ & $0.004\!\pm\!0.000$ \\
DINOv2-B / Qwen3-Emb & $0.035\!\pm\!0.030$ & $0.019\!\pm\!0.007$ & $0.118\!\pm\!0.015$ & $0.277\!\pm\!0.107$ & $\mathbf{0.007}\!\pm\!0.003$ & $0.004\!\pm\!0.000$ \\
DINOv2-B / Qwen3-Gen & $0.014\!\pm\!0.006$ & $0.015\!\pm\!0.006$ & $0.079\!\pm\!0.048$ & $0.227\!\pm\!0.007$ & $\mathbf{0.008}\!\pm\!0.003$ & $0.005\!\pm\!0.000$ \\
DINOv2-G / MPNet & $0.078\!\pm\!0.058$ & $0.077\!\pm\!0.049$ & $0.129\!\pm\!0.051$ & $0.266\!\pm\!0.030$ & $\mathbf{0.008}\!\pm\!0.004$ & $0.004\!\pm\!0.000$ \\
DINOv2-G / Qwen3-Emb & $0.079\!\pm\!0.070$ & $0.039\!\pm\!0.006$ & $0.182\!\pm\!0.046$ & $0.402\!\pm\!0.097$ & $\mathbf{0.007}\!\pm\!0.004$ & $0.003\!\pm\!0.000$ \\
DINOv2-G / Qwen3-Gen & $0.022\!\pm\!0.004$ & $0.024\!\pm\!0.003$ & $0.130\!\pm\!0.112$ & $0.224\!\pm\!0.014$ & $\mathbf{0.008}\!\pm\!0.003$ & $0.004\!\pm\!0.000$ \\
DINOv3-7B / MPNet & $0.126\!\pm\!0.041$ & $0.165\!\pm\!0.049$ & $0.240\!\pm\!0.061$ & $0.408\!\pm\!0.011$ & $\mathbf{0.010}\!\pm\!0.004$ & $0.005\!\pm\!0.000$ \\
DINOv3-7B / Qwen3-Emb & $0.148\!\pm\!0.061$ & $0.098\!\pm\!0.048$ & $0.257\!\pm\!0.057$ & $0.327\!\pm\!0.064$ & $\mathbf{0.008}\!\pm\!0.004$ & $0.004\!\pm\!0.000$ \\
DINOv3-7B / Qwen3-Gen & $0.218\!\pm\!0.079$ & $0.156\!\pm\!0.059$ & $0.180\!\pm\!0.059$ & $0.292\!\pm\!0.011$ & $\mathbf{0.010}\!\pm\!0.004$ & $0.006\!\pm\!0.000$ \\
Franca-G (CLS) / MPNet & $0.059\!\pm\!0.025$ & $0.045\!\pm\!0.006$ & $0.278\!\pm\!0.052$ & $0.497\!\pm\!0.002$ & $\mathbf{0.009}\!\pm\!0.006$ & $0.004\!\pm\!0.000$ \\
Franca-G (CLS) / Qwen3-Emb & $0.065\!\pm\!0.060$ & $0.048\!\pm\!0.007$ & $0.250\!\pm\!0.029$ & $0.429\!\pm\!0.120$ & $\mathbf{0.007}\!\pm\!0.004$ & $0.003\!\pm\!0.000$ \\
Franca-G (CLS) / Qwen3-Gen & $0.172\!\pm\!0.135$ & $0.132\!\pm\!0.064$ & $0.280\!\pm\!0.097$ & $0.577\!\pm\!0.009$ & $\mathbf{0.008}\!\pm\!0.004$ & $0.004\!\pm\!0.000$ \\
Franca-G (mean) / MPNet & $0.113\!\pm\!0.056$ & $0.137\!\pm\!0.050$ & $0.285\!\pm\!0.053$ & $0.320\!\pm\!0.050$ & $\mathbf{0.009}\!\pm\!0.004$ & $0.005\!\pm\!0.000$ \\
Franca-G (mean) / Qwen3-Emb & $0.079\!\pm\!0.054$ & $0.043\!\pm\!0.012$ & $0.290\!\pm\!0.092$ & $0.371\!\pm\!0.104$ & $\mathbf{0.008}\!\pm\!0.003$ & $0.004\!\pm\!0.000$ \\
Franca-G (mean) / Qwen3-Gen & $0.043\!\pm\!0.028$ & $0.047\!\pm\!0.009$ & $0.103\!\pm\!0.064$ & $0.249\!\pm\!0.008$ & $\mathbf{0.009}\!\pm\!0.003$ & $0.005\!\pm\!0.000$ \\
iBOT-Swin / MPNet & $0.219\!\pm\!0.038$ & $0.177\!\pm\!0.014$ & $0.243\!\pm\!0.059$ & $0.207\!\pm\!0.014$ & $\mathbf{0.011}\!\pm\!0.005$ & $0.006\!\pm\!0.000$ \\
iBOT-Swin / Qwen3-Emb & $0.172\!\pm\!0.038$ & $0.110\!\pm\!0.056$ & $0.229\!\pm\!0.103$ & $0.348\!\pm\!0.076$ & $\mathbf{0.009}\!\pm\!0.004$ & $0.005\!\pm\!0.000$ \\
iBOT-Swin / Qwen3-Gen & $0.096\!\pm\!0.083$ & $0.078\!\pm\!0.003$ & $0.208\!\pm\!0.014$ & $0.254\!\pm\!0.009$ & $\mathbf{0.009}\!\pm\!0.003$ & $0.006\!\pm\!0.000$ \\
iBOT-B / MPNet & $0.447\!\pm\!0.021$ & $0.430\!\pm\!0.014$ & $0.395\!\pm\!0.026$ & $0.359\!\pm\!0.008$ & $\mathbf{0.020}\!\pm\!0.005$ & $0.015\!\pm\!0.000$ \\
iBOT-B / Qwen3-Emb & $0.400\!\pm\!0.041$ & $0.390\!\pm\!0.050$ & $0.400\!\pm\!0.012$ & $0.378\!\pm\!0.021$ & $\mathbf{0.017}\!\pm\!0.005$ & $0.013\!\pm\!0.000$ \\
iBOT-B / Qwen3-Gen & $0.428\!\pm\!0.008$ & $0.432\!\pm\!0.011$ & $0.451\!\pm\!0.051$ & $0.362\!\pm\!0.008$ & $\mathbf{0.019}\!\pm\!0.004$ & $0.014\!\pm\!0.001$ \\
\bottomrule
\end{tabular}
\end{table}

\begin{table}[H]
\centering\small
\caption{Additional adapted baselines on all 21 targets, with the number of complete seeds. }
\label{tab:current-classical}
\setlength{\tabcolsep}{4pt}
\begin{tabular}{lrrrr}
\toprule
Adaptation & Seeds & FOSCTTM & Failures & Worst \\
\midrule
VecMap & 5 & 0.3352 & 105/105 & 0.522 \\
MUSE & 5 & 0.4310 & 105/105 & 0.691 \\
GW + WP & 5 & 0.4667 & 103/105 & 0.697 \\
Entropic GW, eps=0.001 & 1 & 0.4439 & 21/21 & 0.662 \\
Unaligned SCA & 5 & 0.4929 & 105/105 & 0.563 \\
\bottomrule
\end{tabular}
\end{table}

\begin{table}[H]
\centering\small
\caption{Additional ablations at their complete-seed coverage, all three PCA widths at five seeds. Best-of-four fits use four full pooled fitting problems. The matched comparison between pooled-space retrieval and distributed native maps is in \cref{tab:nonlin}.}
\label{tab:current-extra-ablations}
\setlength{\tabcolsep}{4pt}
\begin{tabular}{lrrrr}
\toprule
Configuration & Known pairs & Seeds & FOSCTTM & Failures \\
\midrule
PCA 768 & 0 & 5 & $0.0099\!\pm\!0.0040$ & 0/105 \\
PCA 256 & 0 & 5 & $0.0079\!\pm\!0.0025$ & 0/105 \\
PCA 1536 & 0 & 5 & $0.0099\!\pm\!0.0036$ & 0/105 \\
Pooled bridge best-of-4 & 0 & 5 & $0.0072\!\pm\!0.0032$ & 0/105 \\
ConsensusAlign + 1 pairs & 1 & 5 & $0.0083\!\pm\!0.0035$ & 0/105 \\
ConsensusAlign + 10 pairs & 10 & 5 & $0.0058\!\pm\!0.0002$ & 0/105 \\
ConsensusAlign + 100 pairs & 100 & 5 & $0.0056\!\pm\!0.0001$ & 0/105 \\
ConsensusAlign + 1000 pairs & 1000 & 5 & $0.0045\!\pm\!0.0000$ & 0/105 \\
\bottomrule
\end{tabular}
\end{table}

\begin{table}[H]
\centering\small
\caption{Retrieval on the Karpathy COCO test split (5,000 images, 25,000 captions) with the same maps as Table~1 of the main paper, so that the numbers compare with the literature. Means over seven vision encoders and the five seeds, restricted to the 68 vision-text-seed cells for which both methods have a stored map. Bold: \method.}
\label{tab:karpathy}
\setlength{\tabcolsep}{4pt}
\begin{tabular}{llrrrrrr}
\toprule
& & \multicolumn{3}{c}{I$\to$T R@$k$} & \multicolumn{3}{c}{T$\to$I R@$k$}\\
\cmidrule(lr){3-5}\cmidrule(lr){6-8}
Text encoder & Method & 1 & 5 & 10 & 1 & 5 & 10\\
\midrule
MPNet & GeoWarP & 3.0 & 8.5 & 12.6 & 2.4 & 7.7 & 11.9 \\
MPNet & \method & \textbf{12.7} & \textbf{30.5} & \textbf{41.3} & \textbf{9.6} & \textbf{25.8} & \textbf{36.5} \\
\midrule
Qwen3-Emb & GeoWarP & 3.2 & 8.8 & 12.9 & 2.6 & 8.1 & 12.3 \\
Qwen3-Emb & \method & \textbf{14.9} & \textbf{33.6} & \textbf{44.8} & \textbf{11.3} & \textbf{28.6} & \textbf{39.7} \\
\bottomrule
\end{tabular}
\end{table}

\subsection{Downstream Recognition and Retrieval}
We separate improvement on already usable interfaces from recovery of failed searches in \cref{tab:downstream-conditional}. The 60 downstream pair-seed cells use six vision configurations and two text encoders over five seeds. A cell belongs to the successful group when its pairwise WP COCO FOSCTTM is at most 0.05, and to the failed group otherwise. Both methods are evaluated on the same cells in each group. We average over the group's cells within each seed, then report the mean and sample SD of the five seed means. The successful group contains 3, 6, 5, 4 and 3 cells in seeds 0 to 4; the failed group contains 9, 6, 7, 8 and 9. Group membership is used only for this analysis.

\begin{table}[H]
\centering\fontsize{10}{11}\selectfont
\caption{Downstream results conditioned on pairwise WP alignment success. All scores are percentages, mean$\pm$SD over five seed means. Counts are pair-seed cells. Bold: the higher mean within each group.}
\label{tab:downstream-conditional}
\setlength{\tabcolsep}{5pt}
\begin{tabular}{lrlrrrr}
\toprule
WP status & Cells & Method & AVG10 & IN-1k & I$\to$T R@1 & T$\to$I R@1\\
\midrule
WP succeeds & 21 & WP~\cite{schnaus2026rosetta} & $15.4\!\pm\!0.5$ & $8.2\!\pm\!0.8$ & $12.2\!\pm\!1.6$ & $10.5\!\pm\!1.0$ \\
 &  & \method & $\mathbf{21.4}\!\pm\!1.3$ & $\mathbf{14.1}\!\pm\!1.9$ & $\mathbf{26.3}\!\pm\!3.2$ & $\mathbf{22.7}\!\pm\!1.9$ \\
\midrule
WP fails & 39 & WP~\cite{schnaus2026rosetta} & $8.4\!\pm\!0.2$ & $2.2\!\pm\!0.2$ & $2.8\!\pm\!0.4$ & $2.5\!\pm\!0.3$ \\
 &  & \method & $\mathbf{18.7}\!\pm\!1.0$ & $\mathbf{10.8}\!\pm\!0.9$ & $\mathbf{19.6}\!\pm\!1.5$ & $\mathbf{16.8}\!\pm\!1.3$ \\
\bottomrule
\end{tabular}
\end{table}

On the WP-successful cells, \method\ improves AVG10 in 20 of 21 cells and improves ImageNet-1k accuracy and both retrieval directions in all 21. The mean improves in each of the five seeds for every metric. On the WP-failed cells, all four metrics improve in all 39 cells. The collection therefore improves usable interfaces as well as recovering failed searches.

\Cref{tab:downstream-datasets-current} gives the ten classification datasets separately. Uniform chance is $100/C$ percent for $C$ classes in the stored evaluation labels. The paired orthogonal reference uses all 41,391 true image-caption pairs from the image fitting half, fits each native map by orthogonal Procrustes~\cite{Schonemann1966Procrustes}, and uses the same downstream evaluation features and prompts. All five seeds cover the same twelve encoder pairs. Its classification and retrieval results provide empirical references for these features and maps.

\begin{table}[H]
\centering\fontsize{10}{11}\selectfont
\caption{Per-dataset top-1 accuracy in percent, mean$\pm$SD over five seed means, each averaging the same twelve encoder pairs. Bold: \method\ compared with unpaired WP. The final column uses true pairs for fitting.}
\label{tab:downstream-datasets-current}
\setlength{\tabcolsep}{6pt}
\begin{tabular}{lrrrrr}
\toprule
Dataset & Classes & Chance & WP~\cite{schnaus2026rosetta} & \method & Paired orthogonal\\
\midrule
ImageNet-1k & 1000 & 0.10 & $4.3\!\pm\!0.7$ & $\mathbf{12.1}\!\pm\!0.6$ & $19.3\!\pm\!0.1$ \\
Food-101 & 101 & 0.99 & $3.9\!\pm\!0.5$ & $\mathbf{10.3}\!\pm\!0.5$ & $14.2\!\pm\!0.2$ \\
Flowers102 & 102 & 0.98 & $2.3\!\pm\!0.4$ & $\mathbf{4.6}\!\pm\!0.3$ & $6.6\!\pm\!0.2$ \\
DTD & 47 & 2.13 & $5.6\!\pm\!0.7$ & $\mathbf{11.5}\!\pm\!0.5$ & $17.8\!\pm\!0.2$ \\
EuroSAT & 10 & 10.00 & $18.8\!\pm\!1.4$ & $\mathbf{23.6}\!\pm\!1.0$ & $27.3\!\pm\!0.3$ \\
Oxford-IIIT Pets & 37 & 2.70 & $3.9\!\pm\!0.3$ & $\mathbf{6.1}\!\pm\!0.5$ & $10.7\!\pm\!0.2$ \\
Caltech101 & 101 & 0.99 & $18.4\!\pm\!2.7$ & $\mathbf{37.9}\!\pm\!6.7$ & $56.0\!\pm\!0.4$ \\
FGVC Aircraft & 100 & 1.00 & $1.7\!\pm\!0.1$ & $\mathbf{2.2}\!\pm\!0.1$ & $2.4\!\pm\!0.1$ \\
STL-10 & 10 & 10.00 & $48.0\!\pm\!3.8$ & $\mathbf{85.0}\!\pm\!0.7$ & $88.2\!\pm\!0.9$ \\
GTSRB & 43 & 2.33 & $2.1\!\pm\!0.4$ & $\mathbf{3.2}\!\pm\!1.0$ & $5.1\!\pm\!0.1$ \\
\midrule
AVG10 & & & $10.9\!\pm\!0.9$ & $\mathbf{19.6}\!\pm\!0.9$ & $24.7\!\pm\!0.1$ \\
\bottomrule
\end{tabular}
\end{table}

\subsection{Qualitative Retrieval}
\label{supp:qualitative}
\begin{figure}[H]\centering\includegraphics[width=\textwidth,trim=0 246.46bp 0 0,clip]{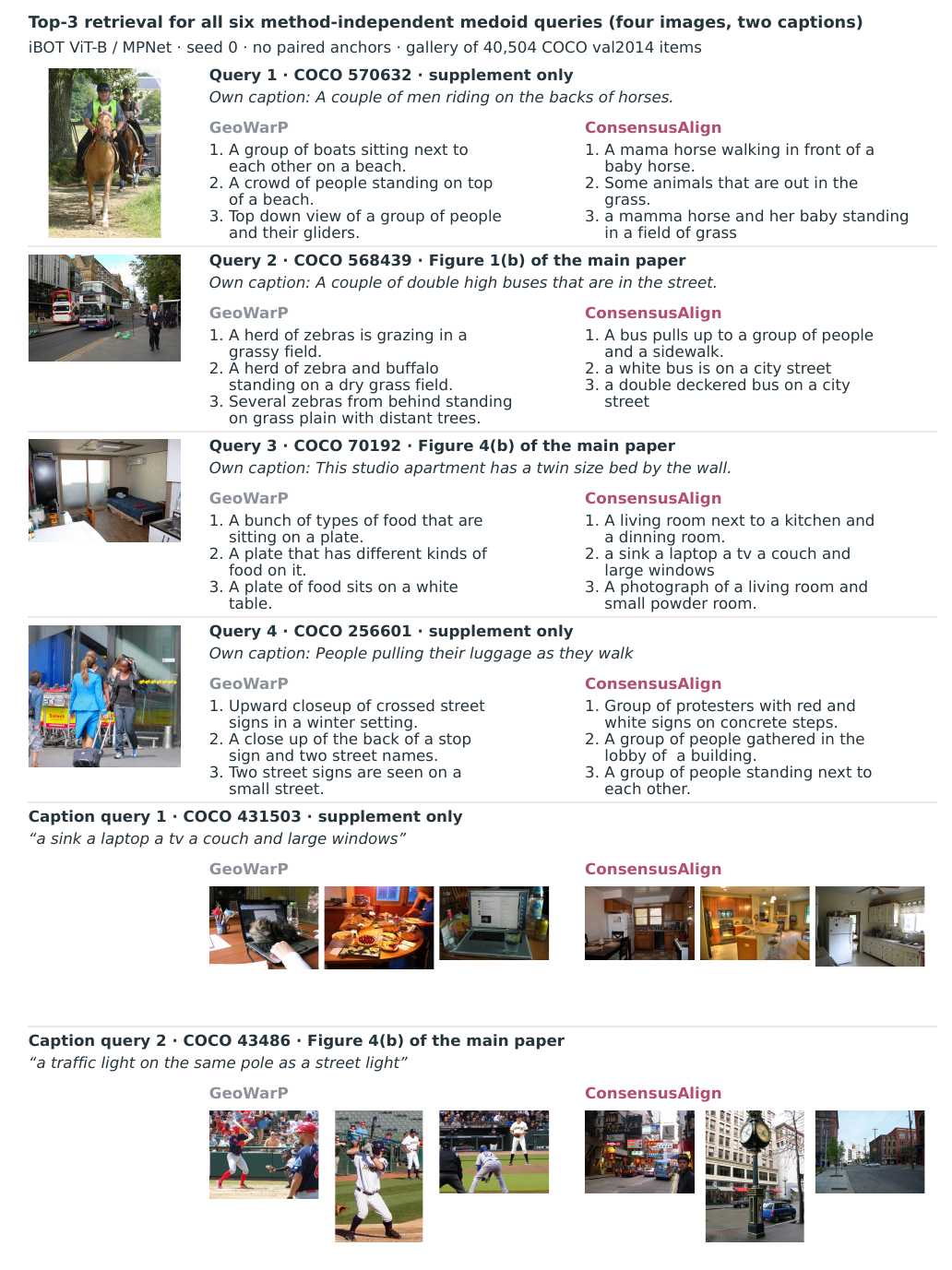}\caption{Image-to-text top-3 retrieval with iBOT-B/MPNet for the four method-independent image medoids (seed 0, no pairs). Each result gives the verbatim first COCO caption of the retrieved image. No retrieved item is the query\textquotesingle s own, so these examples show semantic relevance. The two caption queries are in \cref{fig:current-qual-text}.}\label{fig:current-qual-all}\end{figure}

The image queries are the four method-independent medoid queries of iBOT-B/MPNet at seed 0: k-means medoids of the raw iBOT-B embeddings, clusters in order of size, near-duplicates skipped. Figure~1(b) of the main paper shows the second query (COCO image ID 568439, a landscape photo that fits the panel), Figure~4(b) of the main paper shows the third image query (70192) and the second caption query (43486), and \cref{fig:current-qual-all} of this supplement shows all four with the top-3 captions of both methods. Images and ranked captions come from the actual COCO validation gallery and saved retrieval indices. The pooled captions are semantic retrieval examples, not necessarily the exact caption identity used by FOSCTTM.

\begin{figure}[H]\centering\includegraphics[width=\textwidth,trim=0 0 0 414.5bp,clip]{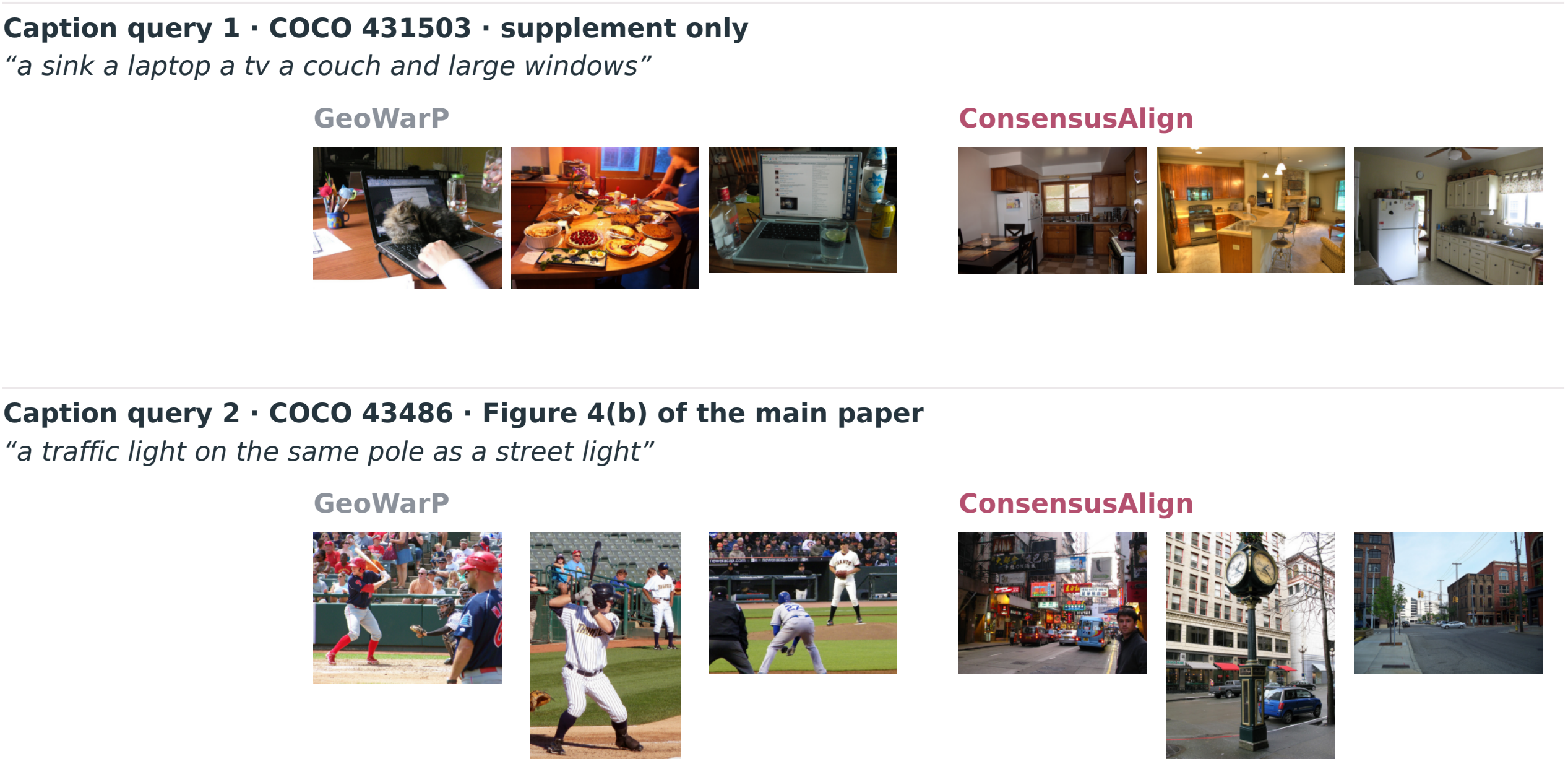}\caption{Text-to-image top-3 retrieval with iBOT-B/MPNet for the two caption queries from the same six-query cohort as \cref{fig:current-qual-all} (seed 0, no pairs, COCO validation gallery). No retrieved image is the caption\textquotesingle s own; the panels show semantic relevance.}\label{fig:current-qual-text}\end{figure}

\section{Few-Pair Results}
\label{supp:recent-baselines}
\begin{figure}[H]\centering\includegraphics[width=\textwidth]{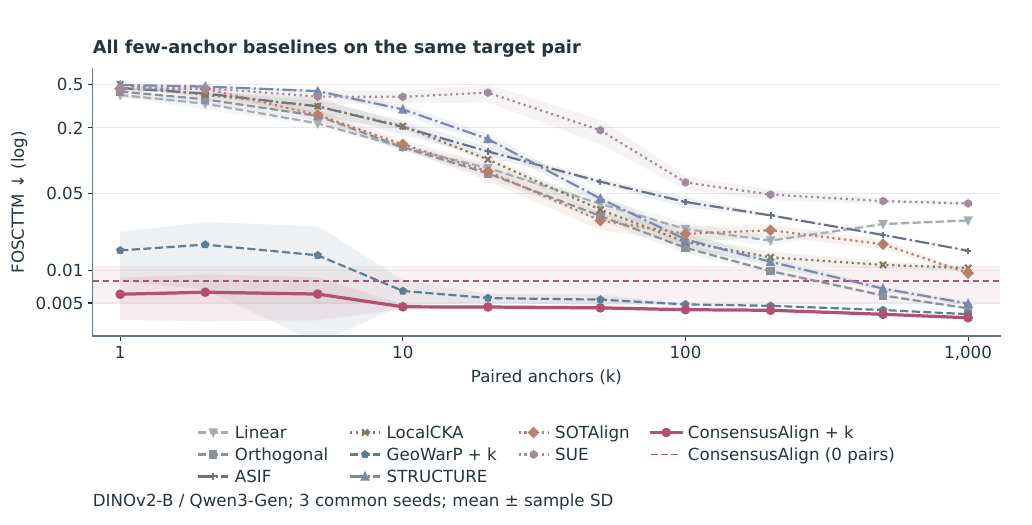}\caption{All eight adapted few-pair baselines and the pooled anchor extension on the fixed DINOv2-B/Qwen3-generative target. This supplementary plot fixes the common complete seeds across all displayed methods, matching the three-seed cohort of the main comparison table and selected curves. The zero-pair pooled reference uses that same intersection.}\label{fig:current-few-all}\end{figure}

All eight adapted methods and our extension use DINOv2-B/Qwen3-generative. Pooled features + Orthogonal is a same-resource control. It uses the pooled features and native factors of \method\ and the same $k$ pairs, and it fits the bridge by orthogonal Procrustes on these pairs alone, with no unpaired fit. It stays above \method\ + $k$ at every shared budget, and it falls below native Orthogonal only from $k=200$. The common three seeds are 0, 1 and 2. Figure~4(a) of the main paper and \cref{fig:current-few-all} use the same seeds.

The zero-pair pooled reference is $0.0079\!\pm\!0.0030$. \Cref{tab:recent-curves-1} reports all methods at $k=1$, 10, 100 and 1,000. \Cref{tab:recent-curves-2} reports them at six additional budgets.

\begin{table}[H]\centering\fontsize{10}{11}\selectfont\setlength{\tabcolsep}{4pt}
\caption{Few-pair alignment at shared budgets on DINOv2-B/Qwen3-generative. FOSCTTM mean$\pm$sample SD over seeds 0 to 2. Every arm uses the same $k$ true pairs. Bold: lowest reported mean per budget.}\label{tab:recent-curves-1}
\begin{tabular}{lrrrr}\toprule
Method & $k=1$ & $k=10$ & $k=100$ & $k=1,000$ \\\midrule
Linear & $0.3990\!\pm\!0.0183$ & $0.1325\!\pm\!0.0096$ & $0.0237\!\pm\!0.0035$ & $0.0284\!\pm\!0.0013$ \\
Orthogonal & $0.4288\!\pm\!0.0518$ & $0.1326\!\pm\!0.0090$ & $0.0160\!\pm\!0.0018$ & $0.0045\!\pm\!0.0001$ \\
ASIF~\cite{norelli2023asif} & $0.4619\!\pm\!0.0070$ & $0.2034\!\pm\!0.0314$ & $0.0420\!\pm\!0.0047$ & $0.0151\!\pm\!0.0001$ \\
LocalCKA~\cite{maniparambil2024localcka} & $0.4994\!\pm\!0.0014$ & $0.2066\!\pm\!0.0132$ & $0.0181\!\pm\!0.0025$ & $0.0105\!\pm\!0.0006$ \\
SUE~\cite{yacobi2025sue} & $0.4562\!\pm\!0.0298$ & $0.3835\!\pm\!0.0483$ & $0.0634\!\pm\!0.0088$ & $0.0407\!\pm\!0.0013$ \\
STRUCTURE~\cite{groger2025structure} & $0.4924\!\pm\!0.0110$ & $0.2945\!\pm\!0.0294$ & $0.0191\!\pm\!0.0028$ & $0.0049\!\pm\!0.0002$ \\
SOTAlign~\cite{roschmann2026sotalign} & $0.4554\!\pm\!0.0261$ & $0.1411\!\pm\!0.0132$ & $0.0215\!\pm\!0.0038$ & $0.0094\!\pm\!0.0003$ \\
GeoWarP+$k$~\cite{schnaus2026rosetta} & $0.0151\!\pm\!0.0072$ & $0.0065\!\pm\!0.0017$ & $0.0049\!\pm\!0.0001$ & $0.0040\!\pm\!0.0000$ \\
Pooled features + Orthogonal & $0.4696\!\pm\!0.1470$ & $0.2285\!\pm\!0.0789$ & $0.0161\!\pm\!0.0025$ & $0.0041\!\pm\!0.0001$ \\
\textbf{\method\ + $k$} & $\mathbf{0.0060}\!\pm\!0.0025$ & $\mathbf{0.0046}\!\pm\!0.0002$ & $\mathbf{0.0043}\!\pm\!0.0000$ & $\mathbf{0.0037}\!\pm\!0.0000$ \\
\bottomrule\end{tabular}

\end{table}
\begin{table}[H]\centering\fontsize{10}{11}\selectfont\setlength{\tabcolsep}{4pt}
\caption{Few-pair alignment at six additional budgets, on the same target and three seeds as \cref{tab:recent-curves-1}. FOSCTTM mean$\pm$sample SD. Bold: lowest mean per budget.}\label{tab:recent-curves-2}
\begin{tabular}{lrrr}\toprule
Method & $k=2$ & $k=5$ & $k=20$ \\\midrule
Linear & $0.3318\!\pm\!0.0398$ & $0.2193\!\pm\!0.0130$ & $0.0858\!\pm\!0.0069$ \\
Orthogonal & $0.3636\!\pm\!0.0315$ & $0.2570\!\pm\!0.0250$ & $0.0765\!\pm\!0.0069$ \\
ASIF~\cite{norelli2023asif} & $0.4111\!\pm\!0.0177$ & $0.3139\!\pm\!0.0408$ & $0.1215\!\pm\!0.0126$ \\
LocalCKA~\cite{maniparambil2024localcka} & $0.3997\!\pm\!0.0316$ & $0.3146\!\pm\!0.0636$ & $0.1025\!\pm\!0.0068$ \\
SUE~\cite{yacobi2025sue} & $0.4610\!\pm\!0.0170$ & $0.3853\!\pm\!0.0619$ & $0.4191\!\pm\!0.0773$ \\
STRUCTURE~\cite{groger2025structure} & $0.4766\!\pm\!0.0073$ & $0.4325\!\pm\!0.0106$ & $0.1581\!\pm\!0.0188$ \\
SOTAlign~\cite{roschmann2026sotalign} & $0.4550\!\pm\!0.0211$ & $0.2617\!\pm\!0.0279$ & $0.0797\!\pm\!0.0149$ \\
GeoWarP+$k$~\cite{schnaus2026rosetta} & $0.0171\!\pm\!0.0105$ & $0.0137\!\pm\!0.0115$ & $0.0056\!\pm\!0.0005$ \\
Pooled features + Orthogonal & $0.4792\!\pm\!0.1062$ & $0.3655\!\pm\!0.1229$ & $0.1252\!\pm\!0.0173$ \\
\textbf{\method\ + $k$} & $\mathbf{0.0063}\!\pm\!0.0029$ & $\mathbf{0.0060}\!\pm\!0.0026$ & $\mathbf{0.0046}\!\pm\!0.0002$ \\
\bottomrule\end{tabular}
\par\medskip
\begin{tabular}{lrrr}\toprule
Method & $k=50$ & $k=200$ & $k=500$ \\\midrule
Linear & $0.0403\!\pm\!0.0049$ & $0.0186\!\pm\!0.0018$ & $0.0263\!\pm\!0.0007$ \\
Orthogonal & $0.0312\!\pm\!0.0029$ & $0.0098\!\pm\!0.0004$ & $0.0059\!\pm\!0.0002$ \\
ASIF~\cite{norelli2023asif} & $0.0644\!\pm\!0.0039$ & $0.0316\!\pm\!0.0007$ & $0.0210\!\pm\!0.0005$ \\
LocalCKA~\cite{maniparambil2024localcka} & $0.0357\!\pm\!0.0040$ & $0.0131\!\pm\!0.0017$ & $0.0112\!\pm\!0.0009$ \\
SUE~\cite{yacobi2025sue} & $0.1898\!\pm\!0.0479$ & $0.0490\!\pm\!0.0056$ & $0.0428\!\pm\!0.0027$ \\
STRUCTURE~\cite{groger2025structure} & $0.0451\!\pm\!0.0046$ & $0.0119\!\pm\!0.0011$ & $0.0068\!\pm\!0.0003$ \\
SOTAlign~\cite{roschmann2026sotalign} & $0.0284\!\pm\!0.0051$ & $0.0232\!\pm\!0.0028$ & $0.0172\!\pm\!0.0017$ \\
GeoWarP+$k$~\cite{schnaus2026rosetta} & $0.0054\!\pm\!0.0006$ & $0.0047\!\pm\!0.0001$ & $0.0043\!\pm\!0.0001$ \\
Pooled features + Orthogonal & $0.0375\!\pm\!0.0029$ & $0.0089\!\pm\!0.0002$ & $0.0053\!\pm\!0.0002$ \\
\textbf{\method\ + $k$} & $\mathbf{0.0045}\!\pm\!0.0002$ & $\mathbf{0.0043}\!\pm\!0.0001$ & $\mathbf{0.0039}\!\pm\!0.0001$ \\
\bottomrule\end{tabular}

\end{table}

\subsection{The Hardest Native Pairs}
\label{supp:hardpair}
\begin{figure}[H]\centering\includegraphics[width=\textwidth]{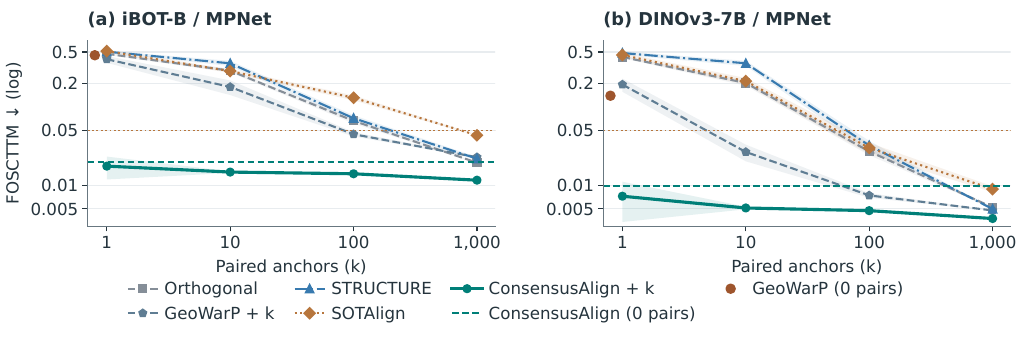}\caption{Few-pair curves on the two hardest native pairs of the grid. Each method is summarised on the seeds that all displayed methods share; bands are $\pm$ one sample SD. The dashed teal line is \method\ with no image-text pairs; the brown dot at $k{=}0$ is pairwise GeoWarP.}\label{fig:hardpair}\end{figure}

The few-pair comparison in the main text uses DINOv2-B/Qwen3-generative, a pair on which pairwise WP already succeeds. Here we repeat it on the two pairs where pairwise WP is weakest: iBOT-B/MPNet, at chance without pairs, and DINOv3-7B/MPNet. We compare GeoWarP with $k$ anchors, SOTAlign, STRUCTURE, orthogonal Procrustes and \method\ with the same $k$ anchors, on the seeds all of them share (\cref{fig:hardpair,tab:hardpair}). On iBOT-B/MPNet, pairwise WP with 1,000 pairs reaches \CAHPIbotWPThousand, still above the zero-pair pooled fit (\CAHPIbotOursZero); orthogonal fitting needs 1,000 pairs to match it (\CAHPIbotOrthoThousand). With 1,000 pairs, STRUCTURE and SOTAlign reach \CAHPIbotStructureThousand and \CAHPIbotSotalignThousand. With ten pairs, the pooled fit reaches \CAHPIbotOursTen. On DINOv3-7B/MPNet, pairwise WP overtakes the zero-pair pooled fit at about 100 pairs (\CAHPDinoWPHundred\ against \CAHPDinoOursZero), and with ten pairs the pooled fit (\CAHPDinoOursTen) matches what pairwise WP and orthogonal fitting reach with 1,000 (\CAHPDinoWPThousand\ and \CAHPDinoOrthoThousand). With 1,000 pairs, STRUCTURE and SOTAlign reach \CAHPDinoStructureThousand and \CAHPDinoSotalignThousand. The collection is worth the most where the pair alone is hardest, and paired anchors remain complementary to it. All methods in this section ran on the same local GPU. SOTAlign on iBOT-B/MPNet was also run on a CPU server: the two devices agree within \CAHPSotalignCpuGpuMaxDiff\ FOSCTTM on the \CAHPSotalignCpuGpuCells\ cells with $k\ge10$ and differ by up to \CAHPSotalignCpuGpuMaxDiffOne\ at $k=1$, where both are at chance.

\begin{table}[H]\centering\small\setlength{\tabcolsep}{3pt}\caption{COCO FOSCTTM on the hard pairs as a function of the number $k$ of paired anchors. Seed means; same seeds within each pair. Zero-pair references are given in the group headings.}\label{tab:hardpair}\begin{tabular}{lrrrr}\toprule Method & 1 & 10 & 100 & 1000\\\midrule
\multicolumn{5}{l}{\textit{iBOT-B / MPNet} (3 seeds; zero-pair WP: 0.454; pooled: \textbf{0.0198})}\\
Orthogonal & 0.474 & 0.288 & 0.0661 & 0.0193 \\
GeoWarP + $k$ & 0.404 & 0.179 & 0.0446 & 0.0227 \\
STRUCTURE & 0.504 & 0.358 & 0.0714 & 0.0218 \\
SOTAlign & 0.512 & 0.285 & 0.130 & 0.0433 \\
\method\ + $k$ & \textbf{0.0175} & \textbf{0.0147} & \textbf{0.0140} & \textbf{0.0116} \\
\midrule
\multicolumn{5}{l}{\textit{DINOv3-7B / MPNet} (3 seeds; zero-pair WP: 0.138; pooled: \textbf{0.0097})}\\
Orthogonal & 0.432 & 0.201 & 0.0270 & 0.0052 \\
GeoWarP + $k$ & 0.192 & 0.0265 & 0.0074 & 0.0048 \\
STRUCTURE & 0.484 & 0.358 & 0.0321 & 0.0050 \\
SOTAlign & 0.458 & 0.212 & 0.0299 & 0.0089 \\
\method\ + $k$ & \textbf{0.0072} & \textbf{0.0051} & \textbf{0.0047} & \textbf{0.0037} \\\bottomrule\end{tabular}\end{table}

\section{Why Pooling Works}
\label{supp:mechanism}
\subsection{Shared Geometry and the Native Objective}
Two checks use only the fitted maps and the stored embeddings. \textbf{Shared geometry.} Linear CKA on the 40,504 COCO validation pairs, between the pooled vision and text spaces ($Z_V,Z_L$, bases fitted on each seed's fitting halves) and between every native pair: the pooled spaces reach 0.733 in all five seeds, while the 21 native pairs range from 0.480 to 0.681 (\cref{tab:mech-encoders}). Projecting a native space through its factor $Q_v$ or $R_\ell$ leaves its CKA unchanged. \textbf{Native objective.} For each of the 103 pair-seed cells with a stored pairwise WP map (seed 1 lacks two), we evaluate the native WP objective, the mean matched inner product on centered and row-normalized features under a Hungarian assignment on two fixed unpaired batches of 10,000. The comparison uses $W_{\mathrm{WP}}$ and $\polar(W_{v\ell})$, where $W_{v\ell}=Q_vCR_\ell^\top$. The rank-$d$ composition is not generally semi-orthogonal in the native dimensions; its polar projection is feasible and changes FOSCTTM by at most $5e-06$. We then run five WP alternation steps (Hungarian on fresh 10,000-row batches, then a Procrustes update) from the pairwise map and from the polar-projected pooled map. \Cref{tab:mech-regimes} groups cells by the pairwise outcome.

\begin{table}[H]\centering\small\setlength{\tabcolsep}{3.5pt}
\caption{Native WP objective of the pairwise map and of the distributed pooled map on the same unpaired batches, by pairwise outcome. Obj.: native mean matched inner product (higher is preferred by the solver). The pooled objective uses $\polar(W_{v\ell})$, while its FOSCTTM column scores the distributed map. ``$>$'': cells where the pooled-side objective exceeds the pairwise objective. ``+5 steps'': after five WP alternation steps from that map. FOSCTTM on COCO validation (lower is better).}
\label{tab:mech-regimes}
\begin{tabular}{lrrrrrrrrrr}
\toprule
& & \multicolumn{3}{c}{Objective of the map} & \multicolumn{3}{c}{Objective, +5 steps} & \multicolumn{3}{c}{FOSCTTM}\\
Pairwise outcome & Cells & WP & Pooled & $>$ & from WP & from pooled & $>$ WP & WP & Pooled & Pooled+5\\
\midrule
pairwise succeeds ($\le0.05$) & 38 & 0.5662 & 0.5636 & 13/38 & 0.5663 & 0.5772 & 35/38 & 0.028 & 0.0076 & 0.0086 \\
pairwise fails ($0.05$ to $0.3$) & 48 & 0.5381 & 0.5467 & 44/48 & 0.5381 & 0.5626 & 48/48 & 0.135 & 0.0090 & 0.0112 \\
pairwise at chance ($>0.3$) & 17 & 0.5197 & 0.4856 & 2/17 & 0.5198 & 0.5215 & 5/17 & 0.405 & 0.0174 & 0.0350 \\
\bottomrule
\end{tabular}
\end{table}

\begin{table}[H]\centering\small\setlength{\tabcolsep}{4pt}
\caption{Per vision encoder: mean native cross-modal CKA over its three text pairs and seeds, and the objective comparison of \cref{tab:mech-regimes}. $\Delta$Obj.: pooled minus pairwise objective.}
\label{tab:mech-encoders}
\begin{tabular}{lrrrrrrr}
\toprule
Vision encoder & Native CKA & Cells & WP FOS & Pooled FOS & $\Delta$Obj. & Pooled $>$ WP & +5 steps $>$ WP\\
\midrule
DINOv2-B & 0.673 & 14 & 0.032 & 0.0078 & -0.0022 & 6/14 & 11/14 \\
DINOv2-G & 0.648 & 14 & 0.045 & 0.0077 & -0.0012 & 6/14 & 14/14 \\
DINOv3-7B & 0.602 & 15 & 0.160 & 0.0093 & +0.0066 & 13/15 & 15/15 \\
Franca-G (CLS) & 0.627 & 15 & 0.083 & 0.0079 & +0.0074 & 11/15 & 15/15 \\
Franca-G (mean) & 0.637 & 15 & 0.080 & 0.0083 & +0.0072 & 10/15 & 15/15 \\
iBOT-Swin & 0.623 & 15 & 0.149 & 0.0093 & +0.0066 & 13/15 & 15/15 \\
iBOT-B & 0.489 & 15 & 0.416 & 0.0187 & -0.0421 & 0/15 & 3/15 \\
\bottomrule
\end{tabular}
\end{table}

\subsection{Shared Geometry Against Collection Size}
\label{sec:ckascale}

The pooled-kernel identity of the main paper averages the member Gram matrices. We tested an interpretation under which every member carries the same shared latent and the private parts are nearly orthogonal. Under that reading the shared term is independent of $M$ and the private term shrinks as $1/M$, so the cross-modal CKA of the pooled space should (i) rise with $M$ for every seed, (ii) rise concavely, (iii) exceed the best member at $M{=}11$ and (iv) predict the grid FOSCTTM across subsets. We measure linear CKA between the pooled vision space of each subset of the collection-size study and the pooled text space of the same seed on the COCO validation pairs, which are used for measurement only.

The outcome is \CACKAOutcome. \Cref{tab:ckascale} gives the size averages. Prediction (i) holds for \CACKAMonoSeeds\ of \CACKASeeds\ seeds, (iii) holds for \CACKAAboveSeeds\ of \CACKASeeds\ seeds, and (iv) holds with Spearman \CACKARho\ over all \CACKASubsets\ subsets (\CACKARhoWithinMin\ to \CACKARhoWithinMax\ within a fixed $M$). The pooled CKA tracks the mean of its members rather than their best: over the \CACKAMultiSubsets\ subsets with more than one member it exceeds the member mean in \CACKAAboveMeanCount\ cases, by \CACKAGapMeanAvg\ on average, and lies \CACKAGapBestAvg\ below the best member. The eleven-member pool with unit weights reaches \CACKAFullPooled\ against \CACKAFullBest\ for its best member (\CACKAFullBestName) and \CACKAFullWorst\ for its worst (\CACKAFullWorstName). Pooled CKA varies non-monotonically with collection size. Members such as MAE have low cross-modal CKA, and uniform pooling gives their Gram matrices full weight. The observed gain over the member mean is consistent with combining member geometries, while the pooled CKA is associated with the alignment quality of the tested subsets.

\paragraph{Follow-Up with Agreement Weights.} The same measurement with the agreement weights of the main text (\cref{tab:ckascale-weighted}, a follow-up analysis) gives an eleven-member pooled CKA of \CACKAWFullPooled\ against \CACKAWFullBest\ for the best member, a grid FOSCTTM of \CACKAWFullGrid\ against \CACKAFullGrid\ with unit weights, and Spearman \CACKAWRho\ with the grid FOSCTTM. The weighted pooled CKA exceeds the best member on \CACKAWAboveSeeds\ of \CACKASeeds\ seeds and rises modestly beyond four members (\CACKAWSizeFour\ at $M{=}4$, \CACKAWSizeSeven\ at $M{=}7$, \CACKAWFullPooled\ at $M{=}11$, against \CACKAUSizeFour, \CACKAUSizeSeven\ and \CACKAFullPooled\ with unit weights). Growth is still not monotone for any seed, since the draw decides which members enter at $M\le2$. In this expansion, agreement weights attenuate low-agreement members and improve both pooled CKA and alignment.

\paragraph{One-Sided Pooling for the Base Collection.} The main text compares the pooled spaces with the 21 native pairs, which pools both sides at once. Pooling one side alone on \CACKAOneSeeds\ seeds gives \CACKAOnePooled\ for the two pooled spaces, \CACKAOneBestVision\ for the best vision member (\CACKAOneBestVisionName) against the pooled text space, \CACKAOneBestText\ for the pooled vision space against the best text member (\CACKAOneBestTextName), and \CACKAOneBestNative\ for the best native pair. The two pooled spaces are above the best one-sided value on \CACKAOnePooledAboveVision\ of \CACKAOneSeeds\ seeds for vision and \CACKAOnePooledAboveText\ for text.

\begin{table}[H]\centering\small\setlength{\tabcolsep}{4pt}\caption{Cross-modal linear CKA between the pooled vision space of a subset and the pooled text space, COCO validation, averaged over the subsets of the collection-size study (\CACKASubsets\ subsets, \CACKASeeds\ seeds, unit weights). Member columns measure each member against the same pooled text space. The last column is the grid-mean FOSCTTM of the same subset.}\label{tab:ckascale}\begin{tabular}{rrcccc}\toprule $M$ & subsets & pooled & mean & best & FOSCTTM\\\midrule
1 & 9 & 0.535 & 0.535 & 0.535 & 0.2564\\
2 & 9 & 0.502 & 0.490 & 0.640 & 0.1484\\
4 & 9 & 0.575 & 0.559 & 0.701 & 0.2070\\
7 & 9 & 0.569 & 0.547 & 0.706 & 0.3509\\
11 & 3 & 0.600 & 0.548 & 0.718 & 0.2030\\\bottomrule\end{tabular}\end{table}

\begin{table}[H]\centering\small\setlength{\tabcolsep}{4pt}\caption{As \cref{tab:ckascale}, with agreement-weighted pooling (labelled follow-up).}\label{tab:ckascale-weighted}\begin{tabular}{rrcccc}\toprule $M$ & subsets & pooled & mean & best & FOSCTTM\\\midrule
1 & 9 & 0.533 & 0.533 & 0.533 & 0.2734\\
2 & 9 & 0.501 & 0.490 & 0.640 & 0.1758\\
4 & 9 & 0.699 & 0.558 & 0.699 & 0.0210\\
7 & 9 & 0.720 & 0.546 & 0.704 & 0.0261\\
11 & 3 & 0.723 & 0.547 & 0.716 & 0.0107\\\bottomrule\end{tabular}\end{table}

\subsection{Scoring and WP Inputs}
\label{sec:scoring}

This test asks whether the pooled advantage depends on the scoring convention, and whether the CKA ordering holds on the inputs that WP actually matches. We rescore the cached maps of the main grid on COCO validation in three ways. \emph{S0} is the released score: inputs are centered and row-normalized, and each image ranks all captions. \emph{S1} ranks all images for each caption. \emph{S2} ranks captions for each image, with inputs centered but not row-normalized. Cosine and raw bilinear scores give the same image-query FOSCTTM, because scaling a query row does not change its ranking. The pooled maps are those of Table~1 of the main paper. The pairwise maps are the stored pairwise WP maps of \cref{supp:mechanism}, which gives 103 pair-seed cells over seed 0 to 4. We also measure linear CKA on the COCO validation pairs after the centering and row normalization that WP applies. The pooled latents use the WP fitting means, and each native pair uses its own means.

We predicted three outcomes. First, the pooled grid mean is below the pairwise grid mean under S0, S1 and S2 in every seed. Second, S1 and S2 change each pooled grid mean by at most 0.01 from S0, and S0 reproduces the stored pooled values within $10^{-4}$. Third, the pooled CKA on WP inputs exceeds the largest native CKA on WP inputs in every seed. The protocol, the predictions and the reporting rule are recorded with the paper.

All three predictions hold (\cref{tab:scoring}). S0 reproduces every stored pooled value exactly. Under every score the pooled maps have no cell above 0.05, while about two thirds of the pairwise cells fail. S1 raises the pooled grid mean by 0.0011 to 0.0013 across seeds, and S2 by 0.0007 to 0.0008. The largest change in a single pooled cell is 0.0038. On WP inputs the pooled CKA is 0.731 in all five seeds, against 0.733 on the raw latents. The native pairs range from 0.480 to 0.681 (median 0.632), as on raw features. Row normalization therefore leaves the CKA ordering of \cref{supp:mechanism} unchanged.

\begin{table}[H]\centering\small\setlength{\tabcolsep}{5pt}
\caption{COCO-validation FOSCTTM of the same cached maps under three scoring conventions, 103 pair-seed cells over seed 0 to 4. Range: lowest and highest seed mean. Fail: cells above 0.05. Lower is better.}
\label{tab:scoring}
\begin{tabular}{lcccccc}
\toprule
& \multicolumn{3}{c}{Pooled} & \multicolumn{3}{c}{Pairwise WP}\\
Score & Mean & Range & Fail & Mean & Range & Fail\\
\midrule
S0 released & 0.0100 & 0.006 to 0.015 & 0/103 & 0.1398 & 0.127 to 0.147 & 65/103\\
S1 text query & 0.0111 & 0.007 to 0.016 & 0/103 & 0.1396 & 0.127 to 0.147 & 66/103\\
S2 no row norm & 0.0107 & 0.006 to 0.016 & 0/103 & 0.1405 & 0.128 to 0.147 & 66/103\\
\bottomrule
\end{tabular}
\end{table}

\subsection{Other Bridge Solvers}
\label{sec:solver}

The construction controls of the main paper retain Wasserstein Procrustes (WP) while changing how the collection is combined. Here we instead hold the concatenation-PCA spaces fixed and change the bridge solver. The hypothesis was that the native pair is the problem and not the solver, so that solvers which fail on every native pair recover a usable bridge once they receive the two pooled spaces. Each of the six other zero-pair solvers of Table~1 of the main paper, and the GW map that initialises GW + WP, fits one bridge between the pooled spaces of the same seed, with its paper-default settings and the adaptations of Table~1 unchanged. The bridge is distributed to the 21 native pairs with the closed-form native factors of the main paper and scored exactly as Table~1, with failure above 0.05. We predicted that GW + WP and mini-vec2vec reach a five-seed grid mean below 0.05, that VecMap and MUSE reach below 0.10, and that no solver is worse on the pooled spaces than on its native pairs. The decision rule was STRONG for at least four solvers below 0.05, PARTIAL for one to three or for a two-fold improvement of every solver, and FAIL otherwise. The protocol and the rule are recorded with the paper.

The outcome is \CASolverOutcome. \Cref{tab:solver} gives the numbers. \CASolverPassWord\ of the \CASolverSolversWord\ solvers reaches a grid mean below 0.05. Grid means range from \CASolverPooledMin\ to \CASolverPooledMax, and \CASolverFailsTotal\ of \CASolverCellsTotal\ distributed cells fail. A two-fold improvement is reached by \CASolverTwoFoldWord\ of the \CASolverWithRefWord\ solvers with a native reference. The best of them is \CASolverBestName, which goes from \CASolverBestNative\ on its native pairs to \CASolverBestPooled\ and still fails \CASolverBestFails\ of \CASolverBestCells\ cells. \CASolverWorseSentence.  The pooled-space column scores each bridge inside the pooled spaces and shows the same picture, so the failure lies in the bridge and not in the distribution step.

These results distinguish the pooling construction from the bridge solver. Alternative within-modality constructions retain the gain with WP, whereas replacing WP on the fixed concatenation-PCA spaces does not recover a usable grid. WP is the one solver among those tested that converts the pooled spaces into a usable bridge, with a grid mean of \CASolverWPPooled\ and no failure against \CASolverWPNative\ and \CASolverWPNativeFails\ of \CASolverWPNativeCells\ failures on the native pairs. The claim of the main paper is restricted accordingly: the gain belongs to the collection together with a solver that can exploit it, and we do not claim that the collection rescues solvers of other families.

\begin{table}[H]\centering\footnotesize\setlength{\tabcolsep}{2.7pt}\caption{Bridge-solver swap on the pooled spaces. Native: the solver on the 21 native pairs, as in Table~1 of the main paper (mean grid FOSCTTM and failures over five seeds). Pooled bridge: the same solver fitted once between the two pooled spaces and distributed to the 21 native pairs, mean\,$\pm$\,SD over \CASolverSeeds\ seed-level grid means, with failures over all cells. Space: FOSCTTM of the bridge inside the pooled spaces, without distribution. Ratio: native mean over the pooled grid mean. The GW initialization is reported separately below.}\label{tab:solver}\begin{tabular}{lcccccc}\toprule & \multicolumn{2}{c}{native} & \multicolumn{2}{c}{pooled bridge} & & \\\cmidrule(lr){2-3}\cmidrule(lr){4-5} solver & mean & fail & mean & fail & space & ratio\\\midrule
VecMap & 0.335 & 105/105 & 0.342\,$\pm$\,0.102 & 105/105 & 0.333 & 1.0\\
MUSE & 0.431 & 105/105 & 0.475\,$\pm$\,0.249 & 105/105 & 0.478 & 0.9\\
GW + WP & 0.467 & 103/105 & 0.381\,$\pm$\,0.137 & 105/105 & 0.373 & 1.2\\
mini-vec2vec & 0.230 & 102/105 & 0.107\,$\pm$\,0.035 & 94/105 & 0.094 & 2.1\\
vec2vec & 0.478 & 105/105 & 0.436\,$\pm$\,0.073 & 105/105 & 0.429 & 1.1\\
UCCA & 0.330 & 105/105 & 0.201\,$\pm$\,0.008 & 105/105 & 0.193 & 1.6\\
\midrule
\rowcolor{oursrow}WP (ours) & 0.146 & 68/105 & 0.010\,$\pm$\,0.004 & 0/105 & \CANonlinPooledMean & 14.7\\\bottomrule\end{tabular}\end{table}

The GW initialization alone gives pooled-grid FOSCTTM 0.392\,$\pm$\,0.118, with 105/105 failed cells, and internal pooled-space FOSCTTM 0.385.

\subsection{One Propagated Bridge}
\label{sec:hub}

Pooling uses the within-modality correspondences to combine the encoders before the bridge is fitted. A simpler use of the same resource keeps one pairwise WP map and propagates it. For a hub pair $(h_v,h_\ell)$ with stored pairwise WP map $W_h$, every grid pair receives $W_{v\ell}=\polar(X_v^\top X_{h_v})\,W_h\,\polar(Y_\ell^\top Y_{h_\ell})^\top$, where $X$ and $Y$ are the centred and normalised fitting rows of each encoder. Each of the 21 grid pairs serves as the hub at each of the five seeds, and the propagated grid is scored exactly as Table~1 of the main paper. Two seed-1 hubs have no stored pairwise map, which leaves 103 hub-seed draws. We predicted that a hub chosen at random gives a grid mean above 0.05 in at least half of the draws, and that the best hub, chosen with the evaluation labels, comes within 0.005 of the pooled fit in at least three of five seeds. The protocol and both predictions are recorded with the paper.

Both predictions hold (\cref{tab:hub}). A random hub gives a grid mean above 0.05 in 64 of 103 draws, and only 25 draws give a grid with no failed cell. The propagated grid fails exactly when the hub's own pairwise fit fails, in 102 of 103 draws. The best hub comes within 0.005 of the pooled fit in three seeds. It is better at seed 1 and worse at seeds 0 and 3. Successful hubs average 0.0219, twice the pooled mean of 0.0099.

Propagation therefore carries the outcome of one pairwise search to the whole grid. It matches pooling only when a successful hub is known in advance, which requires labels or a diagnostic. Pooling needs no such selection and fails no cell.

\begin{table}[H]\centering\footnotesize\setlength{\tabcolsep}{4pt}\caption{One pairwise WP bridge propagated through the collection. Each grid pair serves once as the hub. Grid $>0.05$: hubs whose 21-pair grid mean exceeds 0.05. Clean: hubs whose grid has no failed cell. Best hub: lowest grid mean, selected with the evaluation labels. Pooled: the pooled fit of the same seed.}\label{tab:hub}
\begin{tabular}{lcccccc}\toprule
seed & hubs & grid $>0.05$ & clean & best hub & median hub & pooled\\\midrule
0 & 21 & 13 & 5 & 0.0172 & 0.0859 & 0.0055\\
1 & 19 & 11 & 3 & 0.0067 & 0.0702 & 0.0116\\
2 & 21 & 12 & 7 & 0.0125 & 0.0922 & 0.0119\\
3 & 21 & 13 & 6 & 0.0142 & 0.0724 & 0.0060\\
4 & 21 & 15 & 4 & 0.0190 & 0.1160 & 0.0147\\\bottomrule
\end{tabular}\end{table}

\subsection{Common Width Without Pooling}
\label{sec:pcawp}

Pooling also reduces every encoder to the same width $d=768$ before WP. This test asks whether the reduction alone explains the gain. For each grid pair at seed 0 with at least one native dimension above 768, each encoder is reduced to 768 dimensions by PCA on its own fitting half, without whitening. Pairwise WP then runs with the paper-default settings, and the map is returned to native coordinates. The three pairs whose two encoders are both 768-dimensional are not refitted, because PCA to 768 dimensions is a rotation for them. We predicted that the reduced fits fail on at least half as many of these pairs as pairwise WP. The protocol and the prediction are recorded with the paper.

The prediction holds (\cref{tab:pcawp}). The reduced fits fail on 11 of the 18 pairs, the same count as pairwise WP. Individual pairs move in both directions, and the mean rises from 0.113 to 0.133. With the three unrefitted pairs, the seed-0 grid has 13 of 21 failed cells and a mean of 0.144, against 13 of 21 and 0.127 for pairwise WP and none and 0.0055 for the pooled fit. A common width therefore does not remove the native failures. The gain comes from combining the encoders before the bridge is fitted.

\begin{table}[H]\centering\footnotesize\setlength{\tabcolsep}{1.5pt}\caption{Per-encoder PCA to 768 dimensions before pairwise WP, seed 0, COCO-validation FOSCTTM. Dims: native widths. PCA + WP: each encoder reduced to 768 dimensions first. Bold: above 0.05.}\label{tab:pcawp}
\begin{tabular}{lccc}\toprule
pair & dims & WP & PCA + WP\\\midrule
DINOv2-B / Qwen3-Emb & 768/4096 & 0.022 & 0.016\\
DINOv2-B / Qwen3-Gen & 768/4096 & 0.019 & 0.014\\
DINOv2-G / MPNet & 1536/768 & 0.016 & 0.011\\
DINOv2-G / Qwen3-Emb & 1536/4096 & 0.039 & 0.050\\
DINOv2-G / Qwen3-Gen & 1536/4096 & 0.028 & 0.020\\
DINOv3-7B / MPNet & 4096/768 & \textbf{0.141} & \textbf{0.199}\\
DINOv3-7B / Qwen3-Emb & 4096/4096 & \textbf{0.106} & \textbf{0.129}\\
DINOv3-7B / Qwen3-Gen & 4096/4096 & \textbf{0.205} & \textbf{0.240}\\
Franca-G (CLS) / MPNet & 1536/768 & 0.048 & \textbf{0.061}\\
Franca-G (CLS) / Qwen3-Emb & 1536/4096 & 0.046 & 0.033\\
Franca-G (CLS) / Qwen3-Gen & 1536/4096 & \textbf{0.060} & \textbf{0.111}\\
Franca-G (mean) / MPNet & 1536/768 & \textbf{0.096} & \textbf{0.167}\\
Franca-G (mean) / Qwen3-Emb & 1536/4096 & \textbf{0.101} & 0.022\\
Franca-G (mean) / Qwen3-Gen & 1536/4096 & \textbf{0.142} & \textbf{0.098}\\
iBOT-Swin / Qwen3-Emb & 768/4096 & \textbf{0.149} & \textbf{0.177}\\
iBOT-Swin / Qwen3-Gen & 768/4096 & \textbf{0.084} & \textbf{0.213}\\
iBOT-B / Qwen3-Emb & 768/4096 & \textbf{0.317} & \textbf{0.405}\\
iBOT-B / Qwen3-Gen & 768/4096 & \textbf{0.422} & \textbf{0.436}\\
\midrule
\multicolumn{2}{l}{18 pairs, failed} & 11/18 & 11/18\\
\bottomrule\end{tabular}\end{table}

\subsection{Search Budget of WP}
\label{sec:budget}

The WP solver initialises from $S$ matched $k$-means clusterings of the two spaces and refines with $R$ Hungarian steps on batches of 10,000 rows, with $S{=}30$ and $R{=}100$ by default. \Cref{sec:solver} shows that the other bridge solvers do not profit from the pooled spaces. This experiment, specified before it was run, asks whether the collection makes WP's own search easier. We refit WP with a reduced budget on the pooled spaces of the five seeds and distribute the bridge as in Table~1 of the main paper, and we refit WP with one restart on each of the 21 native pairs for \CABudgetNativeSeeds\ seeds. Everything else stays at the paper default. We predicted a five-seed grid mean below 0.02 with few failed cells for $S{=}1$, $R{=}100$ on the pooled spaces, below 0.05 for $S{=}1$, $R{=}25$, and at least 65\% failed cells for one restart on the native pairs.

\Cref{tab:budget} gives the result. On the pooled spaces, four restarts reach \CABudgetPFourGrid\ with no failed cell, against \CABudgetDefaultPooled\ for the default thirty. A single restart still yields a usable bridge on \CABudgetPOneOk\ of \CABudgetPOneCells\ cells, with a grid mean of \CABudgetPOneGrid, so the first prediction \CABudgetPredOne\ (the mean is just above 0.02, the failures are few). Shortening the refinement to 25 steps costs more than removing the restarts: \CABudgetPShortGrid\ with \CABudgetPShortFails\ failures, and the second prediction \CABudgetPredShort. On the native pairs, one restart fails on \CABudgetNOneFailPct\% of the cells (\CABudgetNOneFails\ of \CABudgetNOneCells), against \CABudgetDefaultNativeFailPct\% for the default budget and \CABudgetRestartsFourFailPct\% for the best of four restarts, and the third prediction \CABudgetPredNative. The collection therefore removes most of the initialisation search that WP needs on a native pair. On the pooled spaces the first basin that WP finds is usually the right one and the refinement does the rest, whereas on a native pair thirty restarts do not suffice on two thirds of the cells. Wall time is dominated by the refinement, so fewer restarts do not make the fit faster. This is the sense in which the collection changes the problem and WP exploits it.

Fit times are measured for seed 0 on one 20-core CPU with no other job running. They cover only the WP fit. Native times are the mean over three pairs: DINOv2-B/MPNet, DINOv2-B/Qwen3-generative and DINOv3-7B/Qwen3-Emb-8B.

\begin{table}[H]\centering\small\setlength{\tabcolsep}{2.5pt}
\caption{WP search budget. $S$: clustered initializations. $R$: refinement steps. Grid FOSCTTM is mean$\pm$SD over five seed means, three for reduced native WP. Failures exceed 0.05. Space: FOSCTTM inside the pooled spaces. Fit: minutes per bridge or per pair. Native default is GeoWarP. Restarts-4 keeps the best of four fits by objective, each with $S{=}30$, $R{=}30$ and batches of 5,000 rows.}
\label{tab:budget}
\begin{tabular}{lrrcccr}\toprule
\multicolumn{7}{l}{\textit{Pooled bridge, distributed to 21 native pairs}}\\
Budget & $S$ & $R$ & Grid & Failed & Space & Fit\\\midrule
\rowcolor{oursrow}default & 30 & 100 & 0.0099\,$\pm$\,0.0040 & 0/105 & \CANonlinPooledMean & 28 \\
reduced & 4 & 100 & 0.0110\,$\pm$\,0.0030 & 0/105 & 0.0069 & 27 \\
reduced & 1 & 100 & 0.0217\,$\pm$\,0.0129 & 3/105 & 0.0156 & 26 \\
reduced & 1 & 25 & 0.0311\,$\pm$\,0.0272 & 24/105 & 0.0240 & 8 \\
\bottomrule\end{tabular}
\par\medskip
\begin{tabular}{lrrccr}\toprule
\multicolumn{6}{l}{\textit{Native pairs, one WP fit per pair}}\\
Budget & $S$ & $R$ & Grid & Failed & Fit\\\midrule
default & 30 & 100 & 0.1457\,$\pm$\,0.0151 & 68/105 & 46 \\
Restarts-4 & $4{\times}30$ & 30 & 0.1274\,$\pm$\,0.0060 & 65/105 & 46 \\
reduced & 1 & 100 & 0.2727\,$\pm$\,0.0417 & 58/63 & 38 \\
\bottomrule\end{tabular}
\end{table}

\section{Larger Collections, New Encoders and VLM Members}
\label{supp:collections}
\begin{table}[H]
\centering\small
\caption{The full 11-visual/13-text unimodal-only pool is evaluated on the original 21 targets. A failed seed run has grid-mean FOSCTTM above 0.05, distinct from a failed target cell. Bold: best within each column.}
\label{tab:current-full-pool}
\setlength{\tabcolsep}{4pt}
\begin{tabular}{lrrr}
\toprule
Pooling & FOSCTTM & Failed runs & Failed target cells \\
\midrule
Unweighted & $0.2020\!\pm\!0.1182$ & 4/5 & 84/105 \\
Weighted & $\mathbf{0.0126}\!\pm\!0.0043$ & \textbf{0/5} & \textbf{0/105} \\
\bottomrule
\end{tabular}
\end{table}

\paragraph{Collection-Size Sweeps.} Every visual size uses matched subset, draw and seed keys for weighted and uniform runs, with 9, 9, 9, 9 and 3 runs at sizes 1, 2, 4, 7 and 11. Members of a modality with fewer than three encoders retain uniform weights; the other modality may still be weighted. The full-pool five-seed experiment is distinct from these subset draws. Quality is nonmonotonic with collection size. The 55 member-agreement points repeat eleven encoder configurations across five seeds; correlations are descriptive.

\paragraph{Attachment.} The ten added text encoders contribute no feature blocks to the original bridge. Each fits a same-input native factor to the frozen text latent. Visual attachment similarly fixes the bridge; all four unimodal-only visual configurations, including MAE mean pooling, are retained in the main table. Each visual timing is one cross-product/polar interval per seed, repeated in three text-target rows; the table reports the median of the five. The ten attached text encoders reach COCO-validation FOSCTTM 0.0095 to 0.0113 (per-encoder mean over five seeds and seven vision targets). The whole pipeline for seed 0 on one 20-core CPU takes concatenation-PCA 4.6 min, pooled WP 31.4 min, distribution of ten factors 2.5 s, against 35.9 min for one pairwise WP fit on the same machine and settings.

\subsection{One Encoder per Family}
\label{sec:family-pooling}

\begin{table}[H]
\centering\small
\setlength{\tabcolsep}{7pt}
\renewcommand{\arraystretch}{1.05}
\caption{\textbf{One encoder per family.} All collections include DINOv3-7B and MPNet, plus the four choices shown. Gen and Emb denote the two Qwen3 8B models. FOSCTTM is the mean $\pm$ sample SD of the 21-pair grid mean over five seeds. Failed cells have FOSCTTM $>0.05$, out of 105 pair-seed cells per row. The last column counts seeds whose grid mean was lower than pairwise WP.}
\label{tab:family-pooling}
\begin{tabular}{@{}llllrcc@{}}
\toprule
DINOv2 & Franca & iBOT & Qwen3 & FOSCTTM $\downarrow$ & Failed cells & Beats WP\\
\midrule
B & CLS & Swin & Gen & $0.0056\!\pm\!0.0002$ & 0/105 & 5/5\\
B & CLS & B & Gen & $0.0141\!\pm\!0.0050$ & 0/105 & 5/5\\
B & mean & Swin & Gen & $0.0061\!\pm\!0.0001$ & 0/105 & 5/5\\
B & mean & B & Gen & $0.0239\!\pm\!0.0098$ & 3/105 & 5/5\\
G & CLS & Swin & Gen & $0.0056\!\pm\!0.0002$ & 0/105 & 5/5\\
G & CLS & B & Gen & $0.0152\!\pm\!0.0053$ & 0/105 & 5/5\\
G & mean & Swin & Gen & $0.0123\!\pm\!0.0039$ & 0/105 & 5/5\\
G & mean & B & Gen & $0.0302\!\pm\!0.0328$ & 21/105 & 5/5\\
\midrule
B & CLS & Swin & Emb & $0.0188\!\pm\!0.0187$ & 9/105 & 5/5\\
B & CLS & B & Emb & $0.0216\!\pm\!0.0068$ & 1/105 & 5/5\\
B & mean & Swin & Emb & $0.0433\!\pm\!0.0326$ & 44/105 & 5/5\\
B & mean & B & Emb & $0.1400\!\pm\!0.0168$ & 105/105 & 4/5\\
G & CLS & Swin & Emb & $0.0196\!\pm\!0.0268$ & 21/105 & 5/5\\
G & CLS & B & Emb & $0.0383\!\pm\!0.0410$ & 21/105 & 5/5\\
G & mean & Swin & Emb & $0.0750\!\pm\!0.0345$ & 84/105 & 5/5\\
G & mean & B & Emb & $0.2134\!\pm\!0.1365$ & 105/105 & 1/5\\
\bottomrule
\end{tabular}
\end{table}

To assess how repeated members of an encoder family affect pooling, we formed collections with one member per family. The visual families were DINOv2-B/G, DINOv3-7B, Franca-G with CLS/mean readouts, and iBOT-Swin/B; the text families were MPNet and Qwen3, with the embedding and generative 8B models in the latter. DINOv2 and DINOv3 were treated as separate families. Each collection therefore contained four visual and two text encoders. Enumerating the choices within the DINOv2, Franca, iBOT and Qwen3 families gave 16 collections. For each collection and each of the five seeds of Table~1, we fitted one WP bridge between 768-dimensional concat-PCA spaces with unit weights. The fitting halves and WP settings matched the full collection. Encoders omitted from a collection were attached by same-modality Procrustes, so every fit was evaluated on the original 21 encoder pairs on COCO validation. The comparison used the main table's pairwise WP grid mean of the same seed.

The pooled grid mean was lower than pairwise WP in 75 of 80 fits (\cref{tab:family-pooling}). Across the 16 collections, the median of the five-seed means was 0.0206, with a range of 0.0056--0.2134. The overall mean was 0.0427, compared with 0.0099 for the full collection. All 21 cells were below the failure threshold in 57 of 80 fits, and 414 of 1,680 cells exceeded 0.05. The worst fit had a grid mean of 0.4545.

The Qwen3 choice separated the results: collections with the generative member averaged 0.0141 with 24/840 failing cells, while those with the embedding member averaged 0.0712 with 390/840 failing cells. Failures also occurred among the members used to build the pool: their mean was 0.0444 (160/640 failures), compared with 0.0416 (254/1040) for pairs involving an attached encoder. Most reduced collections retained an advantage over pairwise fitting, with reliability varying across member selections. Relative to the full pool, this ablation changes both collection size and membership.

\subsection{VLM Members}
\label{supp:vlm}
CLIP\textquotesingle s two towers can join the collection as ordinary members. \Cref{tab:current-vlm-detail} compares such collections with Text-to-Concept for six SSL vision configurations, and \cref{tab:recent-vlm-datasets} gives per-dataset accuracy on DINOv2-B, including GPUA and CLIP\textquotesingle s own zero-shot accuracy.
\begin{table}[H]
\centering\small
\caption{VLM-assisted measurements, each encoder averaged over five matched seeds. Both methods use the same CLIP text tower. Bold: best per configuration and metric; rounded ties share bold.}
\label{tab:current-vlm-detail}
\setlength{\tabcolsep}{4pt}
\begin{tabular}{llrrrr}
\toprule
Visual configuration & Method & AVG10 & IN-1k & I$\to$T & T$\to$I \\
\midrule
dinov2\_vit-b14@224\_mean & Collection + CLIP & \textbf{26.8} & \textbf{19.1} & \textbf{31.4} & 27.3 \\
dinov2\_vit-b14@224\_mean & Text-to-Concept~\cite{moayeri2023textconcept} & 23.9 & 11.6 & 16.9 & \textbf{34.7} \\
dinov2\_vit-g14@224\_mean & Collection + CLIP & \textbf{25.8} & \textbf{17.7} & \textbf{31.5} & 27.1 \\
dinov2\_vit-g14@224\_mean & Text-to-Concept~\cite{moayeri2023textconcept} & 24.6 & 12.3 & 27.5 & \textbf{36.0} \\
franca\_vit-g14@224\_laion & Collection + CLIP & \textbf{27.7} & \textbf{30.4} & \textbf{36.5} & 31.2 \\
franca\_vit-g14@224\_laion & Text-to-Concept~\cite{moayeri2023textconcept} & 23.9 & 10.4 & 12.8 & \textbf{40.6} \\
franca\_vit-g14@224\_laion\_mean & Collection + CLIP & \textbf{25.6} & \textbf{22.0} & \textbf{30.1} & 25.5 \\
franca\_vit-g14@224\_laion\_mean & Text-to-Concept~\cite{moayeri2023textconcept} & 24.6 & 17.4 & 22.6 & \textbf{36.7} \\
ibot\_swint\_14@224\_mean & Collection + CLIP & \textbf{24.9} & \textbf{18.9} & \textbf{24.1} & 20.5 \\
ibot\_swint\_14@224\_mean & Text-to-Concept~\cite{moayeri2023textconcept} & 20.1 & 6.5 & 6.2 & \textbf{23.3} \\
ibot\_vit-b16@224\_mean & Collection + CLIP & \textbf{18.6} & \textbf{9.7} & \textbf{12.5} & 11.7 \\
ibot\_vit-b16@224\_mean & Text-to-Concept~\cite{moayeri2023textconcept} & 14.1 & 4.0 & 3.8 & \textbf{13.4} \\
\bottomrule
\end{tabular}
\end{table}

\begin{table}[H]\centering\small
\caption{Per-dataset top-1 accuracy for the fixed DINOv2-B SSL target. Text-to-Concept and collection+CLIP average five matched seeds and use only the SSL image tower at inference. GPUA and its native-CLIP reference come from the same one-seed classification run with CLIP visual features; GPUA's fitting pool is transductive. Bold: best per dataset within each inference block. }\label{tab:recent-vlm-datasets}
\setlength{\tabcolsep}{7pt}
\begin{tabular}{lrrrr}\toprule
& \multicolumn{2}{c}{SSL image tower only} & \multicolumn{2}{c}{CLIP image tower used}\\
Dataset & T2C~\cite{moayeri2023textconcept} & Collection + CLIP & CLIP native~\cite{radford2021clip} & GPUA~\cite{yu2026gpua}\\\midrule
ImageNet-1k & 11.6 & \textbf{19.1} & 75.1 & \textbf{79.1} \\
Food101 & \textbf{20.3} & 18.3 & 93.4 & \textbf{94.2} \\
Flowers102 & 4.9 & \textbf{8.0} & 75.3 & \textbf{81.1} \\
DTD & 13.4 & \textbf{19.8} & 53.6 & \textbf{61.4} \\
EuroSAT & 19.8 & \textbf{28.4} & 55.4 & \textbf{86.8} \\
OxfordIIITPet & \textbf{24.3} & 15.2 & 93.6 & \textbf{95.4} \\
Caltech101 & 51.6 & \textbf{60.2} & \textbf{91.8} & 68.0 \\
FGVCAircraft & 2.4 & \textbf{2.7} & 31.2 & \textbf{35.6} \\
STL10 & 87.3 & \textbf{91.4} & 99.4 & \textbf{99.7} \\
GTSRB & 3.4 & \textbf{5.2} & 44.7 & \textbf{54.5} \\
\bottomrule\end{tabular}\end{table}

On Flickr30k, collection+CLIP reaches image-to-text R@1 of \CAVLMCollIT\% against \CAVLMTTCIT\% for Text-to-Concept, while text-to-image R@1 is \CAVLMCollTI\% against \CAVLMTTCTI\%. Both averages use the same six SSL vision configurations and five seeds. This is a directional retrieval trade-off. The current comparison changes the alignment construction and does not isolate rank constraints or gallery hubness as its cause.

\section{New Corpora and Modalities}
\label{supp:corpora}
The matched transfer comparison fixes COCO validation and 105 target-seed keys for every arm. Own fitting and pairwise fitting use only the named corpus; reuse additionally retains the COCO fitting resources in its frozen source PCA and bridge. The held-out target column of Table~3(b) of the main paper scores reuse on target pairs that enter no fit (\cref{sec:r4holdout}). A separate cross-corpus fit uses COCO images and SPC texts; its COCO-validation FOSCTTM is 0.0065 versus 0.2348 for WP (21 pairs, three seeds).

\subsection{Frozen COCO Maps on CC3M}
\label{sec:cc3m}

All COCO results of the main paper select settings and report scores on the same 40,504 validation pairs. This test scores the frozen COCO maps on a corpus that played no part in any choice. The pooled and pairwise WP maps of the main grid are applied without refitting to 85,782 CC3M image-caption pairs~\cite{sharma2018cc3m}, one caption per image. Each encoder is centred with its COCO fitting mean, as at deployment. The cells are the seven vision configurations with MPNet and Qwen3-Emb, since the generative Qwen3 features have no CC3M extraction, over five seeds. Two seed-1 cells have no stored pairwise map, which leaves 68 matched cells. We predicted that the pooled mean is lower than the pairwise mean in every seed, and that the pooled failure fraction is at most half of the pairwise fraction. The protocol, both predictions and the reporting rule are recorded with the paper.

The first prediction holds and the second does not (\cref{tab:cc3m}). Pooled maps are better than pairwise maps in all 68 cells and in every seed, with means of 0.067 against 0.222. Pooled maps exceed 0.05 in 57 of 68 cells, against 68 of 68 for pairwise WP. To separate the map from the corpus shift, we add a reference after the test: the orthogonal map fitted on the true partners of the COCO fitting half, frozen and scored on the same CC3M pairs. It reaches 0.049 and itself exceeds 0.05 in 19 of 68 cells. CC3M alt-text captions are short and noisy, so the 0.05 threshold calibrated on COCO is strict here. Pooled maps stay 0.018 above this paired reference on average and remove 90\% of the gap between pairwise WP and the reference. The ranking of encoder pairs is also preserved: iBOT-B is the hardest vision encoder for all three maps, and Qwen3-Emb is easier than MPNet in every row.

\begin{table}[H]\centering\footnotesize\setlength{\tabcolsep}{6pt}\caption{Frozen COCO maps scored on 85,782 CC3M pairs. FOSCTTM is the mean over the $n$ matched seeds. Paired ref.: the orthogonal map fitted on the true partners of the COCO fitting half, added after the test. Fail: cells above 0.05.}\label{tab:cc3m}
\begin{tabular}{lcccccc}\toprule
& & \multicolumn{3}{c}{CC3M FOSCTTM} & \multicolumn{2}{c}{fail}\\\cmidrule(lr){3-5}\cmidrule(lr){6-7}
pair & $n$ & paired ref. & pairwise & pooled & pairwise & pooled\\\midrule
DINOv2-B / MPNet & 5 & 0.045 & 0.114 & 0.062 & 5/5 & 5/5\\
DINOv2-B / Qwen3-Emb & 4 & 0.033 & 0.098 & 0.049 & 4/4 & 2/4\\
DINOv2-G / MPNet & 4 & 0.046 & 0.147 & 0.064 & 4/4 & 4/4\\
DINOv2-G / Qwen3-Emb & 5 & 0.033 & 0.124 & 0.051 & 5/5 & 3/5\\
DINOv3-7B / MPNet & 5 & 0.047 & 0.234 & 0.069 & 5/5 & 5/5\\
DINOv3-7B / Qwen3-Emb & 5 & 0.033 & 0.204 & 0.055 & 5/5 & 5/5\\
Franca-G (CLS) / MPNet & 5 & 0.043 & 0.139 & 0.059 & 5/5 & 5/5\\
Franca-G (CLS) / Qwen3-Emb & 5 & 0.029 & 0.129 & 0.045 & 5/5 & 0/5\\
Franca-G (mean) / MPNet & 5 & 0.050 & 0.228 & 0.068 & 5/5 & 5/5\\
Franca-G (mean) / Qwen3-Emb & 5 & 0.036 & 0.170 & 0.054 & 5/5 & 3/5\\
iBOT-Swin / MPNet & 5 & 0.063 & 0.314 & 0.080 & 5/5 & 5/5\\
iBOT-Swin / Qwen3-Emb & 5 & 0.046 & 0.249 & 0.064 & 5/5 & 5/5\\
iBOT-B / MPNet & 5 & 0.099 & 0.491 & 0.114 & 5/5 & 5/5\\
iBOT-B / Qwen3-Emb & 5 & 0.076 & 0.421 & 0.097 & 5/5 & 5/5\\
\midrule
all & 68 & 0.049 & 0.222 & 0.067 & 68/68 & 57/68\\
\bottomrule\end{tabular}\end{table}

\subsection{Held-Out Target Pairs}
\label{sec:r4holdout}

This test scores corpus reuse on target pairs that enter no fit. For each of SPC, DCI and DOCCI and each of seed 0 to 4, 20\% of the target items are drawn as held-out test pairs. The remaining 80\% are split into disjoint image and text halves, which are used without pairs. We compare five maps on all 21 grid pairs. \emph{COCO maps} are the pooled maps of the main grid with their COCO fitting means. \emph{+ target means} keeps these maps and centres each encoder with its target-half mean. \emph{Reuse} keeps the COCO bridge $C$ and its PCA bases and refits the native factors on the target halves. \emph{Target fit} runs the full pooled pipeline on the target halves. \emph{Paired ref.} is the orthogonal map fitted on the true captions of the target image half, so it uses pairs. We also score reuse on the image half with its true captions, which is the protocol of the target column before this test.

We predicted three outcomes. First, reuse has a lower held-out mean than target fit on at least two of the three corpora. Second, the held-out mean of reuse is within 0.01 of its mean on the fitting half on every corpus. Third, reuse is lower than + target means on every corpus. The protocol, the predictions and the reporting rule are recorded with the paper.

All three predictions hold (\cref{tab:r4holdout}). Reuse is lower than target fit on DCI and DOCCI, where target fit exceeds 0.05 in all 105 cells. On SPC, target fit is lower by 0.0015. Scores on held-out pairs exceed scores on the fitting half by at most 0.0022. Refitting the native factors improves on the frozen COCO maps on every corpus, and it improves on recentring alone. Reuse stays above the paired reference, which uses true target captions.

\begin{table}[H]\centering\fontsize{10}{11}\selectfont\setlength{\tabcolsep}{4pt}
\caption{Target-corpus FOSCTTM over five seeds and 21 grid pairs. Each cell gives the mean, with the number of failed cells ($>0.05$) out of 105 below it. The final column scores reuse on the image fitting half with its true captions; all other maps are scored on held-out pairs.}
\label{tab:r4holdout}
\begin{tabular}{lrcccccc}\toprule
Corpus & \shortstack{Held-out\\pairs} & \shortstack{COCO\\maps} & \shortstack{+ target\\means} & Reuse & \shortstack{Target\\fit} & \shortstack{Paired\\reference} & \shortstack{Reuse,\\fit half}\\\midrule
SPC & 3,912 & \shortstack{0.0172\\(0/105)} & \shortstack{0.0165\\(0/105)} & \shortstack{0.0164\\(0/105)} & \shortstack{0.0149\\(0/105)} & \shortstack{0.0070\\(0/105)} & \shortstack{0.0158\\(0/105)} \\
DCI & 1,561 & \shortstack{0.0525\\(57/105)} & \shortstack{0.0436\\(37/105)} & \shortstack{0.0413\\(34/105)} & \shortstack{0.2805\\(105/105)} & \shortstack{0.0132\\(0/105)} & \shortstack{0.0393\\(30/105)} \\
DOCCI & 2,969 & \shortstack{0.0442\\(37/105)} & \shortstack{0.0387\\(31/105)} & \shortstack{0.0335\\(23/105)} & \shortstack{0.4180\\(105/105)} & \shortstack{0.0076\\(0/105)} & \shortstack{0.0313\\(20/105)} \\
\bottomrule\end{tabular}\end{table}

\subsection{Union of the Three Small Corpora}
\label{sec:union}
Table~3b of the main paper reports that a fresh bridge succeeds on SPC but fails on DCI and DOCCI, while the COCO bridge with refitted native factors succeeds on all three in the matched COCO-validation evaluation. Pooling these corpora tests whether their combined sample size supports a fresh cross-modal search. The three corpora are pooled into one corpus of \CAUnionN\ images with their captions, embedded with the seven vision and three text encoders of the main grid, and each seed splits the union into an image half and a caption half of \CAUnionHalfN\ disjoint items, as the COCO protocol does. The method is that of the main paper, concat-PCA pooling with $d=768$ and unit weights, one WP bridge with the paper default and closed-form distribution. The references are pairwise WP on every pair at seed 0 and the orthogonal ceiling on the hidden partners of the image half. Evaluation uses the COCO validation pairs under the paper protocol and, transductively, the image half against its hidden captions. The pooled bridge runs on the three initial seeds and on the two seeds added afterwards to match the five of the COCO grid. Nothing is tuned. SPC images are Visual Genome images, some of which are also COCO train images, but the captions differ and no COCO fit enters this test. One pooled fit takes \CAUnionFitMinutesOurs\ minutes and one pairwise fit \CAUnionFitMinutesPair\ minutes on the shared CPU.

\paragraph{Predictions and Outcome.} (1) At seed 0 the pooled bridge has a COCO grid mean below $0.05$ with at most 3 of the 21 cells above $0.05$: the mean is \CAUnionOursSeedZero\ with \CAUnionOursFailsSeedZero\ cells above. (2) The same holds in at least two of the three initial seeds: it holds in all three, and in the two seeds added afterwards to match the five of the COCO grid, with grid means from \CAUnionOursLow\ to \CAUnionOursHigh\ across the five, at most \CAUnionOursFailsMax\ cells above $0.05$ in any seed and the largest cell at \CAUnionOursCellHigh. (3) Pairwise WP at seed 0 fails on more cells than the pooled bridge: it fails on \CAUnionPairFails\ of 21 with a grid mean of \CAUnionPairMean, and the pooled bridge is below it on \CAUnionOursBelowPair\ of 21. All three predictions hold, the PASS outcome, so the sentence on small corpora in Sec.~4.4 of the main paper reports the union. The ceiling has a grid mean of \CAUnionCeilMean, so the pooled bridge trails paired fitting by the same margin as on COCO. On the union itself the pooled bridge reaches \CAUnionTransOursSeedZero\ at seed 0 and \CAUnionTransOursLow\ to \CAUnionTransOursHigh\ across seeds, pairwise WP \CAUnionTransPairMean\ and the ceiling \CAUnionTransCeilMean\ (\cref{tab:union-pairs}).

\begin{table}[H]
\centering\small
\caption{Bridges fitted on the union of SPC, DCI and DOCCI, halves of \CAUnionHalfN\ items. FOSCTTM on COCO validation, where 0.5 is chance, for the ceiling and pairwise WP at seed 0 and the pooled bridge at five seeds, and the transductive FOSCTTM of the image half against its hidden captions at seed 0. Bold: lowest value in the row within each block.}
\label{tab:union-pairs}
\setlength{\tabcolsep}{3.5pt}
\begin{tabular}{llrrrrrrrrrr}
\toprule
& & \multicolumn{7}{c}{COCO validation} & \multicolumn{3}{c}{Union, transductive} \\
\cmidrule(lr){3-9}\cmidrule(lr){10-12}
Vision & Text & Ceiling & Pairwise & S0 & S1 & S2 & S3 & S4 & Ceiling & Pairwise & Pooled \\
\midrule
DINOv2-B & MPNet & \textbf{0.003} & 0.042 & 0.027 & 0.018 & 0.026 & 0.005 & 0.038 & \textbf{0.002} & 0.017 & 0.012 \\
DINOv2-B & Qwen3-Emb-8B & \textbf{0.003} & 0.180 & 0.022 & 0.015 & 0.021 & 0.004 & 0.033 & \textbf{0.001} & 0.114 & 0.008 \\
DINOv2-B & Qwen3-8B gen. & \textbf{0.004} & 0.367 & 0.023 & 0.016 & 0.021 & 0.006 & 0.031 & \textbf{0.004} & 0.389 & 0.025 \\
DINOv2-G & MPNet & \textbf{0.003} & 0.100 & 0.026 & 0.017 & 0.024 & 0.005 & 0.035 & \textbf{0.002} & 0.111 & 0.012 \\
DINOv2-G & Qwen3-Emb-8B & \textbf{0.002} & 0.306 & 0.022 & 0.014 & 0.019 & 0.004 & 0.031 & \textbf{0.000} & 0.289 & 0.009 \\
DINOv2-G & Qwen3-8B gen. & \textbf{0.003} & 0.483 & 0.022 & 0.016 & 0.020 & 0.005 & 0.030 & \textbf{0.003} & 0.494 & 0.029 \\
DINOv3-7B & MPNet & \textbf{0.004} & 0.311 & 0.028 & 0.019 & 0.025 & 0.006 & 0.037 & \textbf{0.002} & 0.334 & 0.013 \\
DINOv3-7B & Qwen3-Emb-8B & \textbf{0.002} & 0.318 & 0.023 & 0.016 & 0.020 & 0.005 & 0.032 & \textbf{0.000} & 0.338 & 0.009 \\
DINOv3-7B & Qwen3-8B gen. & \textbf{0.004} & 0.499 & 0.027 & 0.019 & 0.023 & 0.008 & 0.035 & \textbf{0.004} & 0.522 & 0.040 \\
Franca-G CLS & MPNet & \textbf{0.003} & 0.382 & 0.033 & 0.023 & 0.031 & 0.004 & 0.045 & \textbf{0.002} & 0.317 & 0.015 \\
Franca-G CLS & Qwen3-Emb-8B & \textbf{0.002} & 0.084 & 0.029 & 0.020 & 0.025 & 0.004 & 0.042 & \textbf{0.001} & 0.104 & 0.011 \\
Franca-G CLS & Qwen3-8B gen. & \textbf{0.003} & 0.559 & 0.031 & 0.020 & 0.026 & 0.005 & 0.041 & \textbf{0.003} & 0.485 & 0.023 \\
Franca-G mean & MPNet & \textbf{0.004} & 0.141 & 0.026 & 0.018 & 0.024 & 0.005 & 0.036 & \textbf{0.002} & 0.127 & 0.013 \\
Franca-G mean & Qwen3-Emb-8B & \textbf{0.003} & 0.319 & 0.021 & 0.015 & 0.020 & 0.004 & 0.032 & \textbf{0.001} & 0.292 & 0.009 \\
Franca-G mean & Qwen3-8B gen. & \textbf{0.004} & 0.481 & 0.023 & 0.017 & 0.021 & 0.006 & 0.031 & \textbf{0.004} & 0.496 & 0.030 \\
iBOT-Swin & MPNet & \textbf{0.005} & 0.049 & 0.031 & 0.022 & 0.030 & 0.007 & 0.041 & \textbf{0.002} & 0.030 & 0.015 \\
iBOT-Swin & Qwen3-Emb-8B & \textbf{0.004} & 0.051 & 0.026 & 0.019 & 0.024 & 0.006 & 0.037 & \textbf{0.001} & 0.022 & 0.011 \\
iBOT-Swin & Qwen3-8B gen. & \textbf{0.005} & 0.461 & 0.026 & 0.018 & 0.024 & 0.007 & 0.035 & \textbf{0.004} & 0.441 & 0.028 \\
iBOT-B & MPNet & \textbf{0.014} & 0.563 & 0.044 & 0.033 & 0.041 & 0.016 & 0.054 & \textbf{0.007} & 0.558 & 0.027 \\
iBOT-B & Qwen3-Emb-8B & \textbf{0.010} & 0.345 & 0.036 & 0.027 & 0.033 & 0.013 & 0.047 & \textbf{0.003} & 0.339 & 0.019 \\
iBOT-B & Qwen3-8B gen. & \textbf{0.014} & 0.559 & 0.041 & 0.032 & 0.036 & 0.020 & 0.050 & \textbf{0.012} & 0.532 & 0.059 \\
\bottomrule
\end{tabular}
\end{table}

\subsection{ROCOv2 Radiology}
\label{supp:roco}
\begin{table}[H]
\centering\small
\caption{ROCOv2 radiology images and report sentences: 18 encoder pairs, 3 seeds. FOSCTTM on held-out true pairs (0.5 is chance); WP objective is mean matched cosine on unpaired batches (higher is preferred by the solver); agreement is the encoder-disjoint assignment agreement of the main paper on 5,000-row batches, where independent random assignment gives $1/5000=0.0002$. The objective prefers the worst map; the agreement score places all three at chance.}
\label{tab:current-roco}
\setlength{\tabcolsep}{4pt}
\begin{tabular}{lrrr}
\toprule
Map & FOSCTTM & WP objective & Agreement \\
\midrule
Pooled bridge (ours) & 0.513 & 0.406 & 0.0004 \\
Pairwise WP & 0.466 & 0.456 & 0.0003 \\
Orthogonal map, all true pairs & 0.146 & 0.361 & 0.0004 \\
\bottomrule
\end{tabular}
\end{table}

\subsection{Molecule-Text}
\label{sec:mol}
ROCOv2 in Sec.~4.4 of the main paper is a vision-language pair. This test asks whether the collection repairs the unpaired search on a pair that is not, with a corpus of the size of the COCO halves, and whether the agreement score flags the outcome. The molecule encoders are five SMILES transformers trained without text, ChemBERTa-10M-MTR, ChemBERTa-77M-MTR, ChemBERTa-77M-MLM, ChemBERTa-zinc-base and MoLFormer-XL. The text encoders are MPNet, Qwen3-Embedding-0.6B, BGE-large and E5-large-v2, so the grid has \CAMolPairs\ pairs. The unpaired pool holds \CAMolPoolN\ PubChem entries outside ChEBI-20 validation and test. Each seed splits it into a molecule half of \CAMolHalfN\ SMILES strings and a text half of \CAMolHalfN\ descriptions, so no fitting molecule has its description in the text half. The method is that of the main paper: concat-PCA pooling with $d=768$, one WP bridge with the paper default and closed-form distribution. The references are pairwise WP on every pair at seed 0, the orthogonal ceiling fitted on the \CAMolTrainN\ ChEBI-20 training pairs, and the agreement score of Sec.~3.5 of the main paper on a 5,000-item batch with the twelve pairs that share no encoder. Evaluation uses the \CAMolTestN\ ChEBI-20 test pairs, with FOSCTTM molecule-to-text and R@1 in both directions. The pooled bridge and the ceiling run on the three initial seeds and on the two seeds added afterwards to match the five of the COCO grid. Nothing is tuned on the test pairs. One pooled fit takes \CAMolFitMinutesOurs\ minutes and one pairwise fit \CAMolFitMinutesPair\ minutes on the shared CPU.

\paragraph{Predictions and Outcome.} (1) Geometry exists: the ceiling reaches FOSCTTM below $0.10$ on at least 10 of the 20 pairs. It does on all 20, between \CAMolCeilLow\ and \CAMolCeilHigh, and below $0.05$ on \CAMolCeilBelowFive. (2) The pooled bridge has a lower FOSCTTM than pairwise WP on at least 15 of the 20 pairs at seed 0: \CAMolPTwo\ of 20. (3) The pooled bridge reaches FOSCTTM below $0.05$ on at least 10 of the 20 pairs in at least two of the three initial seeds: at most \CAMolPThreeMax\ of 20 in every seed, including the two seeds added afterwards to match the five of the COCO grid. (4) On every pair the agreement score orders the three candidates as their test FOSCTTM does, or all three sit at the random level of $3/5000$: \CAMolPFourOrdered\ pairs are ordered, \CAMolPFourRandom\ sit at the random level and \CAMolPFourViol\ do neither. Under the stated criterion, predictions 1 and 2 hold and 3 fails, which is the PARTIAL outcome. The margin to the ceiling is the whole range (\cref{tab:mol-pairs,tab:mol-seeds}): the pooled bridge has a grid mean of \CAMolOursSeedZero\ at seed 0 and \CAMolOursLow\ to \CAMolOursHigh\ across seeds, pairwise WP \CAMolPairMean, the ceiling \CAMolCeilMean. Both pair-free maps are at chance, with cells from \CAMolCellLow\ to \CAMolCellHigh, so the 15 of 20 of prediction 2 compares two chance-level maps and we do not read it as a gain. Molecule-to-text R@1 is \CAMolCeilRone\% for the ceiling and \CAMolOursRone\% for the pooled bridge.

\paragraph{Agreement Score on the Molecule Grid.} The agreement of all \CAMolCertCands\ pair-free candidates lies between $0$ and \CAMolAgreeMaxOurs, where random assignment gives $0.0002$, at most \CAMolAgreeMaxRows\ rows of 5,000. The ceiling sits at the same level, at most \CAMolAgreeMaxSup, because the reference assignment is the consensus of twelve pairwise fits that all failed. The score therefore states that no map in the grid is supported, which is correct for the pair-free maps and uninformative for the ceiling. The \CAMolPFourViol\ ordering violations of prediction 4 are differences of a few rows between candidates that all sit at the random level. The WP objective prefers the chance-level maps, \CAMolObjPair\ for pairwise WP and \CAMolObjOurs\ for the pooled bridge against \CAMolObjSup\ for the ceiling, as on ROCOv2.

\paragraph{Objective at Paired-Reference Maps.} Prediction 1 tested the existence of an orthogonal map. A follow-up diagnostic, specified with its predictions before it was run, asks whether the WP objective prefers that map (\cref{tab:mol-diag}). On the 20 native pairs at seed 0, five WP steps from the ceiling raise the objective on \CAMolDiagObjUp\ of 20 pairs, from \CAMolDiagSupObj\ to \CAMolDiagRefObj\ on average and towards the \CAMolObjPair\ of the chance-level pairwise map, while the test FOSCTTM moves from \CAMolDiagSupFOS\ to \CAMolDiagRefFOS\ and worsens on every pair. The pooled space does not change this. The ceiling fitted in the pooled spaces reaches \CAMolPooledSupFOS, better than any native ceiling, but its objective is \CAMolPooledSupObj\ against \CAMolPooledBridgeObj\ for the fitted bridge, and five steps from it end at \CAMolPooledRefFOS. The agreement weights of the main paper lie between \CAMolWeightLow\ and \CAMolWeightHigh\ and leave the drift unchanged (\CAMolWeightedRefFOS). On COCO the same test gives the opposite picture: the pooled ceiling has FOSCTTM \CAMolCocoSupFOS\ and objective \CAMolCocoSupObj, the fitted bridge \CAMolCocoBridgeFOS\ and \CAMolCocoBridgeObj, and five steps from the ceiling stay at \CAMolCocoRefFOS. The COCO controls support an initialization benefit; the molecule-text controls expose an objective mismatch that persists in the tested pooled spaces. The diagnostic predicted a drift beyond $0.30$ within five steps. The drift reached \CAMolDiagRefFOS\ natively and \CAMolPooledRefFOS\ pooled, in the predicted direction on every pair but short of that number.

\begin{table}[H]
\centering\small
\caption{WP updates from paired-reference maps. Seed 0. The ceiling is the orthogonal map fitted with true pairs, the fitted map is pairwise WP (native row) or the pooled bridge (pooled rows), and the last columns give the ceiling after five WP steps on unpaired batches of 5,000. Objective: matched cosine, higher is preferred by the solver. On molecule-text the steps raise the objective and degrade the map in every space, on COCO the FOSCTTM change remains small.}
\label{tab:mol-diag}
\setlength{\tabcolsep}{4pt}
\begin{tabular}{lrrrrrr}
\toprule
& \multicolumn{2}{c}{Ceiling} & \multicolumn{2}{c}{Fitted map} & \multicolumn{2}{c}{Ceiling after 5 steps} \\
\cmidrule(lr){2-3}\cmidrule(lr){4-5}\cmidrule(lr){6-7}
Space & FOSCTTM & Objective & FOSCTTM & Objective & FOSCTTM & Objective \\
\midrule
Molecule-text, native (mean of 20) & 0.056 & 0.271 & 0.510 & 0.522 & 0.188 & 0.471 \\
Molecule-text, pooled & 0.020 & 0.254 & 0.492 & 0.505 & 0.107 & 0.414 \\
COCO, pooled & 0.001 & 0.606 & 0.003 & 0.605 & 0.003 & 0.596 \\
\bottomrule
\end{tabular}
\end{table}

For agreement-weighted pooled features, five WP updates from the paired reference change FOSCTTM from 0.020 to 0.107, while the objective rises from 0.256 to 0.411 .

\begin{table}[H]
\centering\small
\caption{Seed-level grid means over the 20 molecule-text pairs: FOSCTTM of the ceiling and the pooled bridge, the range of the pooled cells, and molecule-to-text R@1 in percent.}
\label{tab:mol-seeds}
\setlength{\tabcolsep}{1.8pt}
\begin{tabular}{lrrrrr}
\toprule
Seed & Ceiling & Pooled & Range & R@1 ceiling & R@1 pooled \\
\midrule
S0 & 0.056 & 0.484 & 0.445 to 0.534 & 6.2 & 0.02 \\
S1 & 0.056 & 0.573 & 0.526 to 0.605 & 6.2 & 0.00 \\
S2 & 0.056 & 0.533 & 0.488 to 0.570 & 6.2 & 0.01 \\
S3 & 0.056 & 0.535 & 0.492 to 0.571 & 6.2 & 0.02 \\
S4 & 0.056 & 0.496 & 0.447 to 0.534 & 6.2 & 0.05 \\
\bottomrule
\end{tabular}
\end{table}

\begin{table}[H]
\centering\small
\caption{The 20 molecule-text pairs. FOSCTTM on the 3,300 ChEBI-20 test pairs, where 0.5 is chance, for the ceiling and pairwise WP at seed 0 and the pooled bridge at five seeds, and the agreement score of the three candidates at seed 0 on a 5,000-item batch, where random assignment gives 0.0002. Bold: lowest FOSCTTM in the row.}
\label{tab:mol-pairs}
\setlength{\tabcolsep}{3.5pt}
\begin{tabular}{llrrrrrrrrrr}
\toprule
& & & & \multicolumn{5}{c}{Pooled bridge} & \multicolumn{3}{c}{Agreement} \\
\cmidrule(lr){5-9}\cmidrule(lr){10-12}
Molecule & Text & Ceiling & Pairwise & S0 & S1 & S2 & S3 & S4 & Pooled & Pairwise & Ceiling \\
\midrule
ChemBERTa-10M-MTR & MPNet & \textbf{0.054} & 0.504 & 0.481 & 0.580 & 0.541 & 0.535 & 0.501 & 0.0014 & 0.0000 & 0.0000 \\
ChemBERTa-10M-MTR & Qwen3-Emb-0.6B & \textbf{0.044} & 0.534 & 0.489 & 0.587 & 0.531 & 0.539 & 0.500 & 0.0010 & 0.0000 & 0.0002 \\
ChemBERTa-10M-MTR & BGE-large & \textbf{0.037} & 0.521 & 0.445 & 0.546 & 0.488 & 0.492 & 0.453 & 0.0000 & 0.0000 & 0.0000 \\
ChemBERTa-10M-MTR & E5-large-v2 & \textbf{0.040} & 0.560 & 0.472 & 0.564 & 0.509 & 0.508 & 0.488 & 0.0004 & 0.0002 & 0.0002 \\
ChemBERTa-77M-MTR & MPNet & \textbf{0.068} & 0.510 & 0.485 & 0.559 & 0.549 & 0.533 & 0.492 & 0.0010 & 0.0002 & 0.0000 \\
ChemBERTa-77M-MTR & Qwen3-Emb-0.6B & \textbf{0.054} & 0.449 & 0.495 & 0.564 & 0.552 & 0.546 & 0.500 & 0.0010 & 0.0002 & 0.0000 \\
ChemBERTa-77M-MTR & BGE-large & \textbf{0.046} & 0.568 & 0.446 & 0.526 & 0.508 & 0.494 & 0.447 & 0.0006 & 0.0010 & 0.0000 \\
ChemBERTa-77M-MTR & E5-large-v2 & \textbf{0.049} & 0.537 & 0.483 & 0.546 & 0.525 & 0.510 & 0.480 & 0.0006 & 0.0002 & 0.0002 \\
ChemBERTa-77M-MLM & MPNet & \textbf{0.055} & 0.495 & 0.493 & 0.582 & 0.547 & 0.545 & 0.523 & 0.0010 & 0.0004 & 0.0000 \\
ChemBERTa-77M-MLM & Qwen3-Emb-0.6B & \textbf{0.041} & 0.526 & 0.497 & 0.602 & 0.542 & 0.559 & 0.516 & 0.0006 & 0.0002 & 0.0000 \\
ChemBERTa-77M-MLM & BGE-large & \textbf{0.034} & 0.476 & 0.455 & 0.571 & 0.504 & 0.518 & 0.479 & 0.0004 & 0.0006 & 0.0000 \\
ChemBERTa-77M-MLM & E5-large-v2 & \textbf{0.041} & 0.440 & 0.479 & 0.583 & 0.513 & 0.532 & 0.508 & 0.0000 & 0.0000 & 0.0000 \\
ChemBERTa-zinc & MPNet & \textbf{0.076} & 0.476 & 0.509 & 0.585 & 0.545 & 0.553 & 0.526 & 0.0002 & 0.0000 & 0.0004 \\
ChemBERTa-zinc & Qwen3-Emb-0.6B & \textbf{0.058} & 0.563 & 0.534 & 0.605 & 0.555 & 0.571 & 0.534 & 0.0006 & 0.0004 & 0.0002 \\
ChemBERTa-zinc & BGE-large & \textbf{0.050} & 0.490 & 0.481 & 0.570 & 0.511 & 0.522 & 0.482 & 0.0000 & 0.0000 & 0.0006 \\
ChemBERTa-zinc & E5-large-v2 & \textbf{0.056} & 0.568 & 0.508 & 0.584 & 0.528 & 0.541 & 0.518 & 0.0002 & 0.0002 & 0.0000 \\
MoLFormer-XL & MPNet & \textbf{0.096} & 0.515 & 0.494 & 0.583 & 0.570 & 0.558 & 0.506 & 0.0006 & 0.0000 & 0.0008 \\
MoLFormer-XL & Qwen3-Emb-0.6B & \textbf{0.076} & 0.481 & 0.495 & 0.578 & 0.557 & 0.559 & 0.502 & 0.0002 & 0.0000 & 0.0006 \\
MoLFormer-XL & BGE-large & \textbf{0.070} & 0.468 & 0.470 & 0.570 & 0.538 & 0.532 & 0.470 & 0.0016 & 0.0002 & 0.0002 \\
MoLFormer-XL & E5-large-v2 & \textbf{0.079} & 0.508 & 0.477 & 0.575 & 0.553 & 0.546 & 0.491 & 0.0000 & 0.0000 & 0.0002 \\
\bottomrule
\end{tabular}
\end{table}

\subsection{English-Chinese Sentences}
\label{sec:xling}
The molecule-text test above found a pair whose geometry the pair-free objective does not prefer. This test asks whether the collection repairs the unpaired search on a pair that is not vision-language and that the objective does prefer: monolingual English and monolingual Chinese sentence encoders, the sentence-level form of unsupervised word translation~\cite{conneau2018wordtranslation}. A screen on 20,000 pairs (MPNet against BGE-base-zh) placed the pair in the COCO regime: the ceiling on 3,000 held-out pairs is $0.0014$, five WP steps from it reach $0.0019$ while the objective rises from $0.326$ to $0.421$, and a random map has objective $0.124$ and FOSCTTM $0.481$. The data are the English-Chinese sentence pairs of the TED talks corpus~\cite{reimers2020multilingual}, English of 20 to 400 characters and Chinese of 5 to 300 characters with unique English sentences, shuffled once with seed 0: \CAXlingTestN\ evaluation pairs, \CAXlingTrainN\ training pairs for the ceiling and an unpaired pool of \CAXlingPoolN\ pairs that each run seed splits into an English half and a Chinese half of \CAXlingHalfN\ sentences, so no fitting sentence has its translation in the other half. The English encoders are MPNet, BGE-large-en~\cite{xiao2024cpack}, E5-large-v2~\cite{wang2022e5}, GTE-large~\cite{li2023gte} and MiniLM-L12~\cite{wang2020minilm}. The Chinese encoders are BGE-base-zh and BGE-large-zh~\cite{xiao2024cpack}, text2vec-base-chinese~\cite{xu2023text2vec}, M3E-base~\cite{wang2023m3e} and GTE-base-zh~\cite{li2023gte}. None is multilingual, so no pair shares a space before fitting, and the grid has \CAXlingPairs\ pairs. The method is that of the main paper: concat-PCA pooling with $d=768$ and unit weights, one WP bridge with the paper default and closed-form distribution. The references are pairwise WP on every pair at seed 0, the orthogonal ceiling fitted on the \CAXlingTrainN\ training pairs with the means of the fitting halves, and the agreement score of Sec.~3.5 of the main paper on a 5,000-item batch with the 16 pairs that share no encoder. Evaluation uses the \CAXlingTestN\ held-out pairs, with FOSCTTM English-to-Chinese and R@1 in both directions. The pooled bridge and the ceiling run on the three initial seeds and on the two seeds added afterwards to match the five of the COCO grid. Nothing is tuned on the held-out pairs. One pooled fit takes \CAXlingFitMinutesOurs\ minutes and one pairwise fit \CAXlingFitMinutesPair\ minutes on the shared CPU.

\paragraph{Predictions and Outcome.} (1) Geometry exists: the ceiling reaches FOSCTTM below $0.05$ on at least 20 of the 25 pairs. It does on all 25, between \CAXlingCeilLow\ and \CAXlingCeilHigh. (2) The pooled bridge has a lower FOSCTTM than pairwise WP on at least 19 of the 25 pairs at seed 0: \CAXlingPTwo\ of 25. (3) The pooled bridge reaches FOSCTTM below $0.05$ on at least 20 of the 25 pairs in at least two of the three initial seeds: at least \CAXlingPThreeMin\ of 25 in every seed, including the two seeds added afterwards to match the five of the COCO grid. (4) On every pair the agreement score orders the three candidates as their test FOSCTTM does, or all three sit at the random level of $3/5000$: \CAXlingPFourOrdered\ pairs are ordered and \CAXlingPFourViol\ are not. Predictions 1 to 3 hold, the PASS outcome, so the pair joins Sec.~4.4 of the main paper as a second positive modality pair. Pairwise WP fails on \CAXlingPairFails\ of 25 pairs, with FOSCTTM from \CAXlingPairFailLow\ to \CAXlingPairFailHigh, and succeeds on \CAXlingPairPassed, at most \CAXlingPairPassHigh\ (\cref{tab:xling-pairs}). The pooled bridge has a grid mean of \CAXlingOursSeedZero\ at seed 0 and \CAXlingOursLow\ to \CAXlingOursHigh\ across seeds, with the largest cell at \CAXlingOursCellHigh, against a ceiling of \CAXlingCeilMean\ (\cref{tab:xling-seeds}). English-to-Chinese R@1 over the 3,000 held-out sentences is \CAXlingOursRone\% for the pooled bridge, \CAXlingCeilRone\% for the ceiling and \CAXlingPairRone\% for pairwise WP, and Chinese-to-English R@1 is \CAXlingOursRoneZtoE\% and \CAXlingCeilRoneZtoE\%.

\paragraph{Agreement Score on the Cross-Lingual Grid.} The score separates the pairwise failures from every success: the \CAXlingPairFails\ failed pairwise maps have agreement at most \CAXlingAgreeFailMax, that is \CAXlingAgreeFailRows\ rows of 5,000, and the \CAXlingPairPassed\ successful ones at least \CAXlingAgreePassMin. The pooled bridge lies between \CAXlingAgreeOursLow\ and \CAXlingAgreeOursHigh\ and the ceiling between \CAXlingAgreeSupLow\ and \CAXlingAgreeSupHigh. The \CAXlingPFourViol\ ordering violations of prediction 4 are all of one kind: the pooled bridge scores above the ceiling by at most \CAXlingViolGap\ while its test FOSCTTM is within $0.001$ of the ceiling's, so the score swaps two successful maps that the test barely separates and never ranks a failure above a success. The WP objective is \CAXlingObjSup\ at the ceiling, \CAXlingObjOurs\ at the pooled bridge and \CAXlingObjPair\ at the pairwise maps, so here the objective prefers the paired-reference map, as on COCO and unlike molecule-text.

\begin{table}[H]
\centering\small
\caption{Seed-level grid means over the 25 English-Chinese pairs: FOSCTTM of the ceiling and the pooled bridge, the range of the pooled cells, and English-to-Chinese R@1 in percent.}
\label{tab:xling-seeds}
\setlength{\tabcolsep}{1.8pt}
\begin{tabular}{lrrrrr}
\toprule
Seed & Ceiling & Pooled & Range & R@1 ceil. & R@1 pooled \\
\midrule
S0 & 0.0015 & 0.0018 & 0.0013 to 0.0028 & 92.9 & 89.5 \\
S1 & 0.0014 & 0.0018 & 0.0013 to 0.0030 & 92.9 & 89.5 \\
S2 & 0.0014 & 0.0019 & 0.0013 to 0.0030 & 92.9 & 89.2 \\
S3 & 0.0014 & 0.0018 & 0.0014 to 0.0029 & 92.9 & 89.1 \\
S4 & 0.0014 & 0.0017 & 0.0012 to 0.0030 & 92.9 & 89.2 \\
\bottomrule
\end{tabular}
\end{table}

\begin{table}[H]
\centering\small
\caption{The 25 English-Chinese pairs. FOSCTTM on the 3,000 held-out TED pairs, where 0.5 is chance, for the ceiling and pairwise WP at seed 0 and the pooled bridge at five seeds, and the agreement score of the three candidates at seed 0 on a 5,000-item batch, where random assignment gives 0.0006. Bold: lowest FOSCTTM in the row.}
\label{tab:xling-pairs}
\setlength{\tabcolsep}{3.5pt}
\begin{tabular}{llrrrrrrrrrr}
\toprule
& & & & \multicolumn{5}{c}{Pooled bridge} & \multicolumn{3}{c}{Agreement} \\
\cmidrule(lr){5-9}\cmidrule(lr){10-12}
English & Chinese & Ceiling & Pairwise & S0 & S1 & S2 & S3 & S4 & Pooled & Pairwise & Ceiling \\
\midrule
MPNet & BGE-base-zh & \textbf{0.0019} & 0.4988 & 0.0024 & 0.0023 & 0.0023 & 0.0025 & 0.0021 & 0.1832 & 0.0000 & 0.1820 \\
MPNet & BGE-large-zh & \textbf{0.0017} & 0.5066 & 0.0021 & 0.0021 & 0.0021 & 0.0022 & 0.0021 & 0.1678 & 0.0002 & 0.1596 \\
MPNet & text2vec-zh & \textbf{0.0024} & 0.5156 & 0.0028 & 0.0030 & 0.0030 & 0.0029 & 0.0030 & 0.1430 & 0.0002 & 0.1432 \\
MPNet & M3E-base & \textbf{0.0017} & 0.0027 & 0.0023 & 0.0024 & 0.0022 & 0.0025 & 0.0023 & 0.1432 & 0.1152 & 0.1304 \\
MPNet & GTE-base-zh & \textbf{0.0019} & 0.5049 & 0.0024 & 0.0024 & 0.0024 & 0.0024 & 0.0022 & 0.1616 & 0.0000 & 0.1534 \\
BGE-large-en & BGE-base-zh & \textbf{0.0012} & 0.0016 & 0.0014 & 0.0014 & 0.0013 & 0.0015 & 0.0013 & 0.1844 & 0.1664 & 0.1826 \\
BGE-large-en & BGE-large-zh & \textbf{0.0012} & 0.0015 & 0.0014 & 0.0014 & 0.0014 & 0.0015 & 0.0013 & 0.1692 & 0.1470 & 0.1522 \\
BGE-large-en & text2vec-zh & \textbf{0.0016} & 0.4951 & 0.0018 & 0.0018 & 0.0018 & 0.0018 & 0.0017 & 0.1588 & 0.0012 & 0.1490 \\
BGE-large-en & M3E-base & \textbf{0.0011} & 0.2832 & 0.0013 & 0.0013 & 0.0014 & 0.0014 & 0.0012 & 0.1088 & 0.0024 & 0.0962 \\
BGE-large-en & GTE-base-zh & \textbf{0.0012} & 0.0018 & 0.0016 & 0.0015 & 0.0015 & 0.0015 & 0.0014 & 0.1474 & 0.1264 & 0.1376 \\
E5-large-v2 & BGE-base-zh & \textbf{0.0012} & 0.4572 & 0.0014 & 0.0014 & 0.0015 & 0.0014 & 0.0013 & 0.1758 & 0.0006 & 0.1746 \\
E5-large-v2 & BGE-large-zh & \textbf{0.0012} & 0.2292 & 0.0015 & 0.0015 & 0.0014 & 0.0016 & 0.0014 & 0.1672 & 0.0028 & 0.1638 \\
E5-large-v2 & text2vec-zh & \textbf{0.0013} & 0.5241 & 0.0017 & 0.0019 & 0.0017 & 0.0017 & 0.0017 & 0.1480 & 0.0006 & 0.1362 \\
E5-large-v2 & M3E-base & \textbf{0.0009} & 0.5473 & 0.0013 & 0.0013 & 0.0014 & 0.0014 & 0.0012 & 0.1328 & 0.0002 & 0.1306 \\
E5-large-v2 & GTE-base-zh & \textbf{0.0013} & 0.0022 & 0.0017 & 0.0017 & 0.0017 & 0.0016 & 0.0015 & 0.1704 & 0.1358 & 0.1640 \\
GTE-large & BGE-base-zh & \textbf{0.0011} & 0.0017 & 0.0014 & 0.0013 & 0.0015 & 0.0015 & 0.0013 & 0.1586 & 0.1292 & 0.1600 \\
GTE-large & BGE-large-zh & \textbf{0.0012} & 0.0016 & 0.0014 & 0.0014 & 0.0014 & 0.0015 & 0.0013 & 0.1566 & 0.1420 & 0.1516 \\
GTE-large & text2vec-zh & \textbf{0.0015} & 0.4138 & 0.0018 & 0.0018 & 0.0020 & 0.0019 & 0.0017 & 0.1422 & 0.0004 & 0.1402 \\
GTE-large & M3E-base & \textbf{0.0010} & 0.0018 & 0.0013 & 0.0013 & 0.0015 & 0.0014 & 0.0013 & 0.0932 & 0.0772 & 0.0862 \\
GTE-large & GTE-base-zh & \textbf{0.0012} & 0.0015 & 0.0015 & 0.0015 & 0.0017 & 0.0016 & 0.0015 & 0.1264 & 0.1094 & 0.1172 \\
MiniLM-L12 & BGE-base-zh & \textbf{0.0014} & 0.4821 & 0.0018 & 0.0016 & 0.0019 & 0.0018 & 0.0018 & 0.1714 & 0.0000 & 0.1644 \\
MiniLM-L12 & BGE-large-zh & \textbf{0.0016} & 0.5552 & 0.0018 & 0.0017 & 0.0019 & 0.0018 & 0.0018 & 0.1604 & 0.0000 & 0.1500 \\
MiniLM-L12 & text2vec-zh & \textbf{0.0022} & 0.3485 & 0.0027 & 0.0025 & 0.0029 & 0.0026 & 0.0025 & 0.1342 & 0.0014 & 0.1258 \\
MiniLM-L12 & M3E-base & \textbf{0.0017} & 0.0026 & 0.0021 & 0.0020 & 0.0023 & 0.0021 & 0.0021 & 0.1386 & 0.1076 & 0.1300 \\
MiniLM-L12 & GTE-base-zh & \textbf{0.0015} & 0.5350 & 0.0020 & 0.0018 & 0.0021 & 0.0019 & 0.0019 & 0.1416 & 0.0004 & 0.1326 \\
\bottomrule
\end{tabular}
\end{table}

\section{Pair-Free Diagnosis}
\label{supp:diagnosis}
Each COCO target is scored against the 12 pairwise WP reference maps that share neither of its encoders (\cref{supp:implementation}). The candidates are one cheap pairwise fit per pair and seed (105 maps), the full pairwise WP fit and three restarts of each of 14 pairs at seed 0 (56 maps), 161 maps in total, of which 124 fail. \Cref{tab:reviewed-loo} reports how well agreement and the native WP objective detect these failures, and \cref{tab:calibration} applies one threshold fixed on COCO to the other corpora.
\begin{table}[H]
\centering
\small
\setlength{\tabcolsep}{4pt}
\begin{tabular*}{\textwidth}{@{\extracolsep{\fill}}lrrrrrr@{}}
\toprule
Candidate cohort & Maps & Seeds & Fail/success & Agreement AUROC & Objective AUROC & Spearman $\rho$ \\
\midrule
Independent candidates & 161 & 5 & 124/37 & 0.9742 & 0.8729 & -0.9361 \\
\bottomrule
\end{tabular*}
\caption{\textbf{Agreement as a pair-free failure detector on COCO.} Failure means FOSCTTM $>0.05$; AUROC detects failure with negative agreement or negative native WP objective. Each reference fuses 12 pairwise fits sharing neither encoder ID with the tested edge. Among the six independent-candidate $(\mathrm{seed},v,\ell)$ groups containing both outcomes, mean within-group AUROC is \CAReviewWithinPairAgreementAUROC{} for agreement and \CAReviewWithinPairObjectiveAUROC{} for objective.}
\label{tab:reviewed-loo}
\end{table}

\paragraph{One Threshold Across Corpora.}
The main paper ranks maps by agreement within a corpus. Here we fix one absolute threshold and apply it elsewhere. On the 161 independent COCO candidates, the threshold that maximizes Youden's $J$ for detecting FOSCTTM $>0.05$ is an agreement of $0.0207$, with precision 0.991 and recall 0.871 (\cref{tab:calibration}). The same threshold makes no error on the English-Chinese sentence pairs and flags every near-chance map on ROCOv2 and the molecules. On the SPC+DCI+DOCCI union it flags 13 of 20 successful pooled maps and 12 of 21 paired references: the agreement of a correct map on that smaller corpus falls below the COCO level, so the absolute score is corpus dependent while its ranking within a corpus is not (\cref{tab:union-pairs}).

\begin{table}[H]
\centering\small
\caption{A single agreement threshold ($0.0207$, fixed on COCO) applied to every corpus. A map counts as failed at FOSCTTM $>0.05$ and as flagged when its agreement falls below the threshold. Missed: failed and not flagged. False alarm: flagged and not failed.}
\label{tab:calibration}
\setlength{\tabcolsep}{4pt}
\begin{tabular}{llrrrrr}
\toprule
Corpus & Maps & $n$ & Failed & Flagged & Missed & False alarm\\
\midrule
COCO (calibration) & independent candidates & 161 & 124 & 109 & 16 & 1 \\
\midrule
SPC+DCI+DOCCI union & \method & 21 & 1 & 13 & 1 & 13 \\
SPC+DCI+DOCCI union & pairwise WP & 21 & 18 & 21 & 0 & 3 \\
SPC+DCI+DOCCI union & paired reference & 21 & 0 & 12 & 0 & 12 \\
\midrule
English-Chinese TED & \method & 25 & 0 & 0 & 0 & 0 \\
English-Chinese TED & pairwise WP & 25 & 15 & 15 & 0 & 0 \\
English-Chinese TED & paired reference & 25 & 0 & 0 & 0 & 0 \\
\midrule
ROCOv2 & \method & 54 & 54 & 54 & 0 & 0 \\
ROCOv2 & pairwise WP & 54 & 54 & 54 & 0 & 0 \\
ROCOv2 & paired reference & 54 & 54 & 54 & 0 & 0 \\
\midrule
ChEBI-20 molecules & \method & 20 & 20 & 20 & 0 & 0 \\
ChEBI-20 molecules & pairwise WP & 20 & 20 & 20 & 0 & 0 \\
ChEBI-20 molecules & paired reference & 20 & 11 & 20 & 0 & 9 \\
\bottomrule
\end{tabular}
\end{table}

\subsection{Family-Excluded References}
\label{sec:family-references}

References that exclude the target encoder configurations can still contain models from the same families. We assessed whether the diagnostic retained its accuracy after excluding those families. The test kept the same 161 COCO candidates, including 124 failures and 37 successes, the cached pairwise WP maps and the 5,000-item evaluation batches. The visual families were DINOv2-B/G, DINOv3-7B, the two Franca-G read-outs, and iBOT-Swin/B. The text families were MPNet and Qwen3, with both 8B variants in the latter. A family-excluded reference shared neither the target's visual family nor its text family.

Family exclusion left 5, 6, 10 or 12 reference pairs per target. To control for this change in count, we drew ten random subsets of the original 12 configuration-disjoint references, matching the family-excluded count for each target. The draws used a generator seeded with $s+11$. The reference maps were unchanged. We predicted that the family-excluded AUROC would be at least 0.90 and would be within 0.03 of the mean AUROC of the matched-count controls.

Neither prediction held (\cref{tab:family-references}). The original configuration-disjoint score was reproduced, while family exclusion reduced AUROC below every matched-count draw and below the native WP objective. Fewer references alone did not account for the reduction. On this cohort, diagnostic performance depended on reference-family composition as well as reference count. The AUROC of 0.9742 is specific to the configuration-disjoint reference ensemble.

\begin{table}[H]
\centering\small
\setlength{\tabcolsep}{6pt}
\renewcommand{\arraystretch}{1.05}
\caption{\textbf{Family exclusion in the diagnostic.} Failure detection on the same 161 COCO candidates. AUROC uses negative agreement or negative WP objective. The random-control row is the mean over ten matched-count draws, ranging from 0.8294 to 0.9544. The native WP objective uses no external reference maps.}
\label{tab:family-references}
\begin{tabular}{lcr}
\toprule
Reference set or score & Reference pairs & AUROC\\
\midrule
Configuration-disjoint & 12 & 0.9742\\
Family-excluded & 5, 6, 10 or 12 & 0.7981\\
Matched-count random & Matched per target & 0.9030\\
Native WP objective & None & 0.8729\\
\bottomrule
\end{tabular}
\end{table}

\section{Nonlinear Native Maps}
\label{supp:nonlinear}
\subsection{Nonlinear Distribution Factors}
\begin{table}[H]
\centering
\caption{\textbf{Distribution-factor comparison.} Same five seeds, same 21 pairs and the same cached bridge $C$ per seed as Table~1 of the main paper. Only the factor that maps a native encoder to its pool changes. Fit time is for all ten factors of one seed on one GPU. Retrieval in the pooled spaces themselves, with every encoder at inference and the same bridges, reaches \CANonlinPooledMean$\pm$\CANonlinPooledSD. Bold: best per column, rounded ties included.}
\label{tab:nonlin}
\small
\setlength{\tabcolsep}{4pt}
\begin{tabular}{lrcrrrrr}
\toprule
& & Failed & & IN-1k & \multicolumn{2}{c}{Flickr30k R@1 $\uparrow$} & Fit\\
Factor & FOSCTTM $\downarrow$ & cells $\downarrow$ & AVG10 $\uparrow$ & $\uparrow$ & I$\to$T & T$\to$I & s\\
\midrule
Orthogonal factor, $Q_v=\operatorname{polar}(X_v^\top Z_V)$ (paper) & $0.0098\!\pm\!0.0039$ & \textbf{0/105} & 19.8 & \textbf{12.3} & 22.2 & 18.9 & 18 \\
Least-squares factor (ridge, $\lambda$ on a held-out tenth) & \textbf{$0.0079\!\pm\!0.0034$} & \textbf{0/105} & \textbf{20.0} & 12.2 & \textbf{23.1} & \textbf{20.7} & 26 \\
Skip-connected MLP factor (one hidden layer of 2048) & \textbf{$0.0079\!\pm\!0.0034$} & \textbf{0/105} & 19.9 & 12.2 & \textbf{23.1} & 20.6 & 35 \\
\bottomrule
\end{tabular}
\end{table}

\label{sec:nonlin}
The main text reports a gap between retrieval in the pooled spaces and retrieval through the distributed native maps. Because the native-to-pool correspondence is known, the factor $Q_v$ can be any regressor from $X_v$ to $Z_V$ fitted on the fitting half, with no image-text pairs. We tested the hypothesis that the rank-$d$ orthogonal restriction causes the gap and the prediction that a skip-connected MLP factor reaches a grid mean of at most $0.0065$ with no failed cell and AVG10 of at least $20.5$. The least-squares factor and the MLP use the same held-out tenth of the fitting half for $\lambda$ and for early stopping, and inference stays at two encoders, $\cos(f_v(x)\,C,\,g_\ell(y))$.

\Cref{tab:nonlin} shows the outcome. The orthogonal arm uses the bridges and polar factors of Table~1, evaluated here with factored cosine. Least squares lowers the grid mean from \CANonlinPolarMean\ to \CANonlinRidgeMean, improves \CANonlinRidgeBetterCells\ of \CANonlinCells\ cells, keeps every cell below the failure threshold and raises Flickr R@1 in both directions. The MLP matches least squares in every cell to within \CANonlinRidgeMlpMaxAbsDiff\ FOSCTTM. Early stopping ends after about nine epochs, and the tested nonlinearity provides no further improvement. The prediction for the MLP was therefore not met, and the result is partial under the stated criterion (mean at most $0.0080$ and no new failed cell). Replacing the orthogonal factors with least squares recovers \CANonlinRidgeGapRecovered\% of the pooled-native gap while retaining a closed-form fit. A residual gap remains under both tested regressors.

\subsection{Distilled Nonlinear Heads}
\label{sec:head}
The distributed map of Table~1 of the main paper is linear. This test asks whether the collection supports a better pair-free map. The bridge fixes the cross-modal correspondence once: the fitting halves are the same items for every encoder, so one entropic plan between the pooled image coordinates mapped by $C$ and the pooled text coordinates is a property of the items, not of an encoder pair. We compute this plan once per seed in the pooled spaces of the main paper (cost $1-\cos$, $\varepsilon=0.02$ times the median cost, log-domain Sinkhorn to a marginal error of $10^{-4}$, \CAHeadSinkhornItsLow\ to \CAHeadSinkhornItsHigh\ iterations) and distill it into one head per native encoder, $f_v=\mathrm{Linear}(d_v,1024)\circ\mathrm{LayerNorm}$ and $g_\ell$ likewise, with the bidirectional soft-target contrastive loss of AnchorDOT (chunks of 2{,}000, 200 epochs, AdamW, learning rate $10^{-3}$, weight decay $10^{-2}$, 200 warm-up steps, learned scale from $1/0.07$ clamped at 100). No pair, no label and no selection on validation enter, all constants are those of a four-pair pilot and are fixed in the record, and the similarity is $\cos(f_v(x),g_\ell(y))$. One head takes \CAHeadDistillSeconds\ s on one GPU. The baseline of every cell is the linear map of the same seed and bridge.

\paragraph{Predictions and Outcome.} (1) The head has a lower FOSCTTM than the linear map in at least 95 of the 105 cells and its grid mean is at most $0.7$ times the linear map's: \CAHeadBetterCells\ of 105 and a ratio of \CAHeadRatio\ (\CAHeadMean\ against \CAHeadLinearMean). (2) Flickr30k I$\to$T R@1, averaged over seeds, is higher on at least 9 of the 12 downstream pairs: \CAHeadItoTPairs\ of 12, by \CAHeadItoTGain\ points on average. (3) AVG10 stays within $2.0$ points: \CAHeadAVGDelta\ points over the twelve pairs, with \CAHeadAVGOutside\ pairs outside this band in opposite directions (\cref{tab:head-pairs}). (4) No head cell fails: \CAHeadFails, worst \CAHeadWorst. All four hold, so the decision rule of the record places the head in Table~1 of the main paper. T$\to$I R@1 rises on \CAHeadTtoIPairs\ of 12 pairs, by \CAHeadTtoIGain\ points on average, and ImageNet-1k rises to \CAHeadIN\%. The head is a retrieval refinement of the collection, not a replacement for pairs: its zero-shot accuracy stays at the linear map's level.

\paragraph{Seed Effect of the Head.} \Cref{tab:head-seeds} lists the seed-level grid means. On the two seeds where the bridge is best the head reduces FOSCTTM to \CAHeadSeedRatioBest\ of the linear map's, on the other three to \CAHeadSeedRatioRest. The \CAHeadBetterCells\ cells are the complement of nine cells on the weaker seeds, all with a Qwen3 text target, where the head is above the linear map by at most \CAHeadNotBetterGap. The plan is computed from the bridge, so a worse bridge gives a worse plan, which the head then reproduces. Pairs, not the head, remove this seed dependence (Table~1 of the main paper).

\paragraph{Pilot Check.} The pilot used one plan per pair instead of one per seed. The record asks us to report if the per-seed plan is worse than the per-pair plan by more than $0.001$ on any of the four pilot pairs: it is worse on \CAHeadPilotWorse\ of them, and better on all four (\cref{tab:head-pilot}).

\begin{table}[H]
\centering\small
\caption{Seed-level grid means over the 21 pairs for the linear map and the distilled head, the ratio, the cells where the head is lower, and the Sinkhorn iterations of the seed's plan.}
\label{tab:head-seeds}
\setlength{\tabcolsep}{4pt}
\begin{tabular}{lrrrrr}
\toprule
Seed & Linear map & Head & Ratio & Head lower & Sinkhorn it. \\
\midrule
S0 & 0.0055 & 0.0020 & 0.37 & 21/21 & 558 \\
S1 & 0.0116 & 0.0091 & 0.79 & 17/21 & 399 \\
S2 & 0.0119 & 0.0093 & 0.78 & 18/21 & 442 \\
S3 & 0.0060 & 0.0025 & 0.41 & 21/21 & 366 \\
S4 & 0.0147 & 0.0110 & 0.75 & 19/21 & 410 \\
\bottomrule
\end{tabular}
\end{table}

\begin{table}[H]
\centering\small
\caption{The four pilot pairs at seed 0: the linear map, the pilot head distilled from a per-pair plan, and the confirmatory head distilled from the per-seed plan. FOSCTTM on COCO validation.}
\label{tab:head-pilot}
\setlength{\tabcolsep}{3pt}
\begin{tabular}{lrrr}
\toprule
Pair & Linear & Pair plan & Seed plan \\
\midrule
DINOv2-B / MPNet & 0.0042 & 0.0023 & 0.0018 \\
iBOT-B / MPNet & 0.0138 & 0.0094 & 0.0034 \\
DINOv3-7B / MPNet & 0.0048 & 0.0026 & 0.0016 \\
DINOv2-B / Qwen3-Gen & 0.0045 & 0.0025 & 0.0019 \\
\bottomrule
\end{tabular}
\end{table}

\begin{table}[H]
\centering\small
\caption{Linear map (L) and distilled head (H) on the twelve downstream pairs, each entry the mean over five seeds. FOSCTTM on COCO validation, Flickr30k R@1 and zero-shot accuracies in percent. Bold: the better of the two, rounded ties included.}
\label{tab:head-pairs}
\setlength{\tabcolsep}{3.5pt}
\begin{tabular}{lrrrrrrrrrr}
\toprule
& \multicolumn{2}{c}{FOSCTTM $\downarrow$} & \multicolumn{2}{c}{I$\to$T $\uparrow$} & \multicolumn{2}{c}{T$\to$I $\uparrow$} & \multicolumn{2}{c}{AVG10 $\uparrow$} & \multicolumn{2}{c}{IN-1k $\uparrow$} \\
\cmidrule(lr){2-3}\cmidrule(lr){4-5}\cmidrule(lr){6-7}\cmidrule(lr){8-9}\cmidrule(lr){10-11}
Pair & L & H & L & H & L & H & L & H & L & H \\
\midrule
DINOv2-B / MPNet & 0.0086 & \textbf{0.0064} & 22.8 & \textbf{31.9} & 20.0 & \textbf{21.7} & \textbf{20.0} & 19.4 & 10.8 & \textbf{11.9} \\
DINOv2-B / Qwen3-Emb & 0.0072 & \textbf{0.0063} & 26.3 & \textbf{34.7} & 22.4 & \textbf{24.6} & \textbf{22.4} & 21.7 & 13.5 & \textbf{15.2} \\
DINOv2-G / MPNet & 0.0084 & \textbf{0.0062} & 23.4 & \textbf{31.9} & 20.1 & \textbf{23.2} & 19.5 & \textbf{19.7} & 10.0 & \textbf{11.9} \\
DINOv2-G / Qwen3-Emb & 0.0071 & \textbf{0.0061} & 27.4 & \textbf{37.4} & 22.7 & \textbf{25.9} & \textbf{22.1} & 21.8 & 12.7 & \textbf{15.6} \\
Franca-G (CLS) / MPNet & 0.0094 & \textbf{0.0065} & 27.6 & \textbf{33.6} & 23.0 & \textbf{23.8} & \textbf{19.8} & 19.3 & \textbf{15.5} & 13.2 \\
Franca-G (CLS) / Qwen3-Emb & 0.0073 & \textbf{0.0065} & 30.6 & \textbf{36.9} & \textbf{26.9} & 26.4 & \textbf{22.5} & 20.4 & \textbf{19.9} & 16.1 \\
Franca-G (mean) / MPNet & 0.0090 & \textbf{0.0063} & 22.9 & \textbf{31.9} & 19.3 & \textbf{22.6} & 19.9 & \textbf{20.2} & 11.9 & \textbf{13.4} \\
Franca-G (mean) / Qwen3-Emb & 0.0076 & \textbf{0.0065} & 25.8 & \textbf{36.5} & 22.5 & \textbf{25.6} & \textbf{21.9} & \textbf{21.9} & 15.1 & \textbf{17.6} \\
iBOT-Swin / MPNet & 0.0105 & \textbf{0.0070} & 16.8 & \textbf{23.8} & 14.4 & \textbf{15.6} & 17.5 & \textbf{17.9} & 9.6 & \textbf{10.2} \\
iBOT-Swin / Qwen3-Emb & 0.0086 & \textbf{0.0070} & 19.7 & \textbf{26.8} & 16.8 & \textbf{17.7} & 19.0 & \textbf{19.4} & 12.4 & \textbf{13.6} \\
iBOT-B / MPNet & 0.0199 & \textbf{0.0078} & 9.3 & \textbf{19.8} & 8.6 & \textbf{12.7} & 14.9 & \textbf{16.6} & 5.9 & \textbf{8.5} \\
iBOT-B / Qwen3-Emb & 0.0172 & \textbf{0.0077} & 11.1 & \textbf{22.2} & 10.1 & \textbf{14.5} & 16.3 & \textbf{18.9} & 7.4 & \textbf{11.3} \\
\bottomrule
\end{tabular}
\end{table}

\section{Limitations}
\label{sec:limitations}
We examine the effects of collection composition, corpus size, semantic coverage and objective mismatch.

\paragraph{Collection Size and Composition.}
The pooled kernel combines the members' geometries, as Eq.~4 of the main paper states. Reliability depends on the choice and number of members (\cref{sec:family-pooling,tab:minimal}). Collections with fewer than three members use unit weights. In the subset study, pooled CKA tracks the member mean rather than the best member (\cref{sec:ckascale}). In the eleven-encoder collection, the two MAE configurations lower the uniform pool until it fails on \CAFullUnweightedFails\ of five seed runs. Agreement weights restore every run, but the weighted pool reaches \CAFullWeighted\ against \CAPooledMean\ for the seven-encoder base. In this expansion, weighting improves tolerance of heterogeneous additions, while the smaller base collection remains more accurate. The two-encoder draws below have mixed alignment outcomes.

\paragraph{How Small the Collection Can Be.}
This experiment pools two vision encoders with one or two text encoders, fits one bridge, and attaches the other grid encoders through their same-modality factors (\cref{tab:minimal}). With two text encoders, the in-pool pairs succeed in 4 of 6 configuration-seed cases, including both seeds of the two draws without iBOT-B, and the attached encoders reach the same error as the pool members. With one text encoder, only 1 of 6 succeed. In every successful case the pooled error is below the pairwise WP mean of the same pairs (5 of 5). Two encoders per modality can therefore already replace the pairwise search, while the draw that pairs iBOT-B with Franca-G fails at this size.

\begin{table}[H]
\centering\small
\caption{Minimal collections on the COCO grid: two vision encoders pooled with one or two text encoders, seeds 0 and 1. In-pool: mean FOSCTTM over the pooled pairs. Attached: mean over the remaining grid encoders, linked to the pool by one same-modality fit each. Failed: cells above $0.05$ among all 21 grid pairs. WP: pairwise WP mean on the in-pool pairs at the same seed. Bold: in-pool mean below $0.05$.}
\label{tab:minimal}
\setlength{\tabcolsep}{5pt}
\begin{tabular}{llcrrrr}
\toprule
Vision pool & Text pool & Seed & In-pool & Attached & Failed & WP\\
\midrule
DINOv2-B + DINOv3-7B & MPNet & 0 & 0.1565 & 0.1520 & 21/21 & 0.0912 \\
DINOv2-B + DINOv3-7B & MPNet & 1 & 0.1211 & 0.1186 & 21/21 & 0.0460 \\
DINOv2-B + DINOv3-7B & Qwen3-gen + MPNet & 0 & $\mathbf{0.0128}$ & 0.0133 & 0/21 & 0.0846 \\
DINOv2-B + DINOv3-7B & Qwen3-gen + MPNet & 1 & $\mathbf{0.0162}$ & 0.0172 & 0/21 & 0.1000 \\
\midrule
DINOv2-G + Franca-G (mean) & Qwen3-Emb & 0 & 0.0892 & 0.0940 & 21/21 & 0.1349 \\
DINOv2-G + Franca-G (mean) & Qwen3-Emb & 1 & $\mathbf{0.0368}$ & 0.0398 & 3/21 & 0.0616 \\
DINOv2-G + Franca-G (mean) & Qwen3-gen + Qwen3-Emb & 0 & $\mathbf{0.0093}$ & 0.0118 & 0/21 & 0.0954 \\
DINOv2-G + Franca-G (mean) & Qwen3-gen + Qwen3-Emb & 1 & $\mathbf{0.0199}$ & 0.0239 & 0/21 & 0.0470 \\
\midrule
iBOT-B + Franca-G (mean) & Qwen3-gen & 0 & 0.3047 & 0.3016 & 21/21 & 0.2584 \\
iBOT-B + Franca-G (mean) & Qwen3-gen & 1 & 0.2174 & 0.1979 & 21/21 & 0.2286 \\
iBOT-B + Franca-G (mean) & Qwen3-gen + Qwen3-Emb & 0 & 0.3062 & 0.2934 & 21/21 & 0.2590 \\
iBOT-B + Franca-G (mean) & Qwen3-gen + Qwen3-Emb & 1 & 0.2503 & 0.2389 & 21/21 & 0.2191 \\
\bottomrule
\end{tabular}
\end{table}

\paragraph{Corpus Size and Bridge Reuse.}
Fresh pooled fits on DCI and DOCCI, with halves of 4k and 7k items, remain above the failure threshold, whereas SPC with 10k halves and the union with \CAUnionHalfN\ halves fall below it (Tables~3b and 4 of the main paper). These comparisons vary corpus composition as well as sample size. Reusing a COCO bridge improves alignment on the smaller corpora because only the same-modality factors are refitted there. ROCOv2 provides a large-pool case with poor cross-modal alignment.

\paragraph{The Two Halves Must Cover Similar Semantics.}
The pooled bridge assumes that the image half and the text half describe the same range of content. We tested this on COCO by removing from the text half every caption of an image that mentions a dropped category group, with the image half unchanged, on seeds 0 to 2. Removing the animal group (about 22\% of the texts) raises the pooled grid mean, which is $0.006$ to $0.012$ at full coverage, to $0.044$ to $0.060$, with 3 to 19 of 21 cells above the failure threshold. On the three pairs fitted pairwise under the same drop, pairwise WP rises further, to $0.08$ to $0.27$, so moderate mismatch degrades the pooled bridge less than a pairwise fit. Removing animals, vehicles and food (about 48\%) leaves every cell of both arms above the threshold, and one pooled seed lands in an anti-aligned basin with FOSCTTM above $0.5$. The halves should therefore be drawn from the same distribution of content, as the disjoint split of one corpus provides.

\paragraph{Search Failure and Objective Mismatch.}
The collection changes the search inputs while retaining the WP objective. On ROCOv2 and molecule-text, pooled and pairwise WP fits remain near chance. Paired orthogonal references reach \CAROCOFOSSup\ and \CAMolCeilMean, respectively, while unpaired fits remain near chance and have higher average WP objectives (Table~4 of the main paper). On molecule-text, five WP updates from the paired references increase the objective and worsen FOSCTTM in both native and pooled spaces (\cref{tab:mol-diag}). These controls identify a limitation beyond initialization. The agreement score remains low for both the failed unpaired maps and the paired references because their reference ensemble is built from unsuccessful pairwise fits. The low agreement reflects weak support from this reference ensemble.

\paragraph{Seed Variance of the Pooled Fit.}
A pooled fit is one WP search, and its result varies across seeds: the grid mean of the base method has a sample SD of $0.0040$ on COCO, the union bridge ranges from \CAUnionOursLow\ to \CAUnionOursHigh\ across seeds, and the agreement score of the union is lower than that of TED, where the reference fits are more reliable. Ten known pairs reduce the observed variability on the tested hard pairs (\cref{supp:hardpair}); the zero-pair fit retains seed dependence.

\paragraph{Paired References for Linear Maps.}
The native maps factor through the pooled coordinates. Orthogonal fits with true pairs provide empirical references for these frozen features. Over five seeds, the COCO paired reference reaches mean FOSCTTM \CACeilGridMean, zero-shot accuracy \CACeilAVG\% and Flickr30k R@1 of \CACeilIToT\% and \CACeilTToI\%, against \CAPooledMean, \CAPooledAVG\% and \CAPooledIToT\% for the pooled bridge. On the union the paired reference is \CAUnionCeilMean\ against \CAUnionOursMean\ for the pooled bridge, and the downstream accuracy of self-supervised encoders through the pooled interface, \CAPooledAVG\% AVG10, stays well below CLIP's own zero-shot accuracy and trails paired fitting on fine-grained classes. The nonlinear head of \cref{sec:nonlin} narrows this gap without pairs but does not close it. COCO validation served during development; the held-out target, TED, ROCOv2 and molecule-text protocols provide additional evaluations beyond COCO development.

\FloatBarrier

\end{document}